\documentclass[lettersize,journal]{IEEEtran}
\usepackage{amsmath,amsfonts}
\usepackage{algorithmic}
\usepackage{algorithm}
\usepackage{array}
\usepackage[caption=false,font=footnotesize,labelfont=sf,textfont=sf]{subfig}
\usepackage{textcomp}
\usepackage{stfloats}
\usepackage{url}
\usepackage{verbatim}
\usepackage{graphicx}
\usepackage{cite}
\usepackage{booktabs}
\usepackage{bm}
\usepackage{amssymb}
\usepackage[group-separator={,}]{siunitx}

\begin{document}

\title{RFMask: A Simple Baseline for Human Silhouette Segmentation with Radio Signals}


\author{Zhi Wu, Dongheng Zhang, Chunyang Xie, Cong Yu, Jinbo Chen, Yang Hu 
        \\ and Yan Chen,~\IEEEmembership{Senior Member,~IEEE}
\IEEEcompsocitemizethanks{\IEEEcompsocthanksitem Zhi Wu, Dongheng Zhang, Jinbo Chen, and Yan Chen are with the school of Cyber Science and Technology, University of Science and Technology of China, Hefei 230026, China (E-mail: wzwyyx@mail.ustc.edu.cn, dongheng@ustc.edu.cn, 
jinbochen@mail.ustc.edu.cn, eecyan@ustc.edu.cn).

Yang Hu is with the Department of Electronic Engineering and Information Science, University of Science and Technology of China, Hefei 230026, China (E-mail: eeyhu@ustc.edu.cn)

Chunyang Xie and Cong Yu are with University of Electronic Science and Technology
of China. (E-mail: \{chunyangxie, congyu\}@std.uestc.edu.cn.)\protect\\
}}



\maketitle

\begin{abstract}
Human silhouette segmentation, which is originally defined in computer vision, has achieved promising results for understanding human activities. However, the physical limitation makes existing systems based on optical cameras suffer from severe performance degradation under low illumination, smoke, and/or opaque obstruction conditions.
To overcome such limitations, in this paper, we propose to utilize the radio signals, which can traverse obstacles and are unaffected by the lighting conditions to achieve silhouette segmentation. 
The proposed RFMask framework is composed of three modules.
It first transforms RF signals captured by millimeter wave radar on two planes into spatial domain and suppress interference with the signal processing module.
Then, it locates human reflections on RF frames and extract features from surrounding signals with human detection module. 
Finally, the extracted features from RF frames are aggregated with an attention based mask generation module. 
To verify our proposed framework, we collect a dataset containing \textbf{804,760} radio frames and \textbf{402,380} camera frames with human activities under various scenes. 
Experimental results show that the proposed framework can achieve impressive human silhouette segmentation even under the challenging scenarios (such as low light and occlusion scenarios) where traditional optical-camera-based methods fail. 
To the best of our knowledge, this is the first investigation towards segmenting human silhouette based on millimeter wave signals. We hope that our work can serve as a baseline and inspire further research that perform vision tasks with radio signals. 
The dataset and codes will be made in public.
\end{abstract}


\begin{figure}[t]
    \centering
    \vspace{-0.25cm}
    \subfloat[]{
        \begin{minipage}[b]{0.21\columnwidth}
            \includegraphics[width=1.1\textwidth]{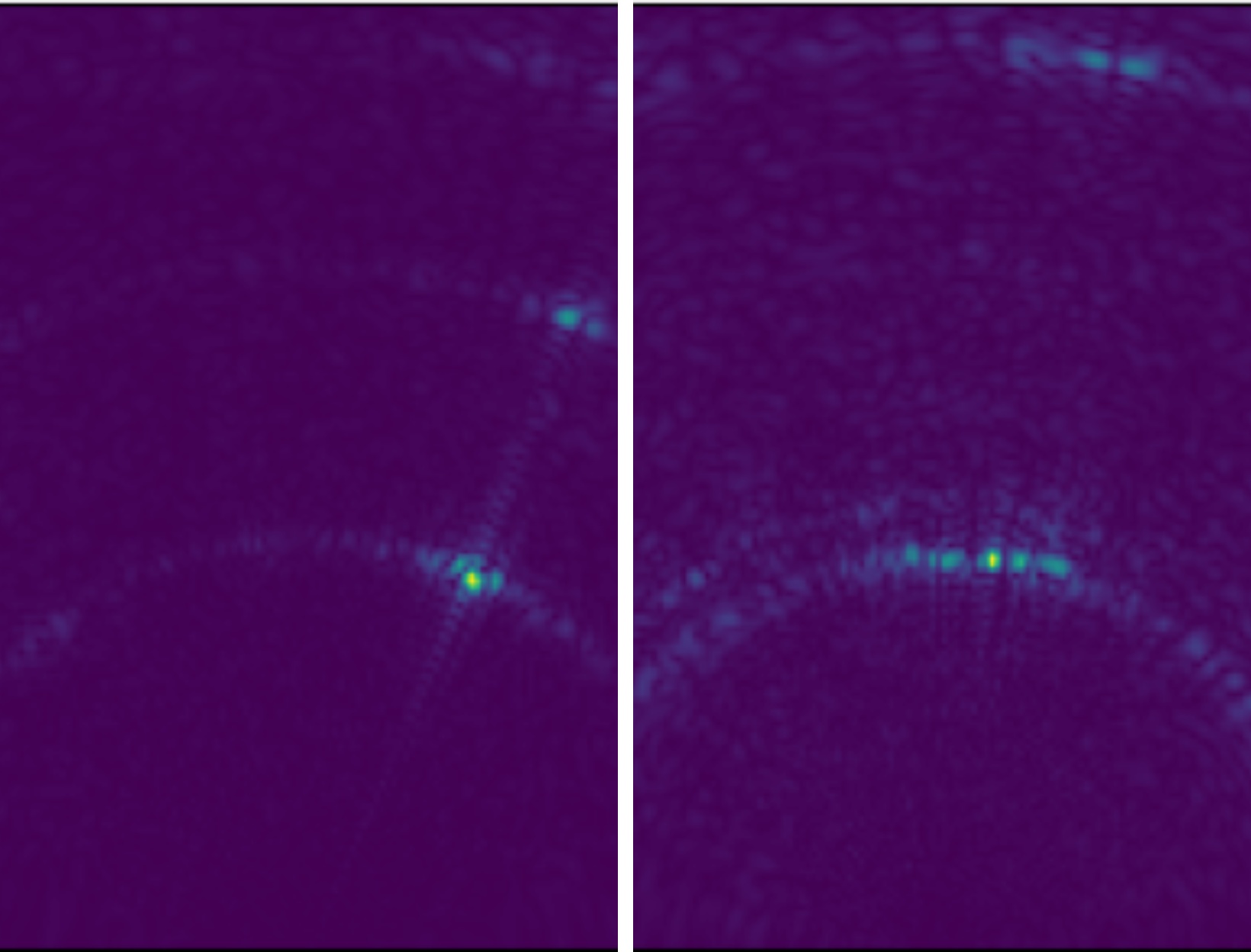} \\
            \vspace{-0.4cm}
            \includegraphics[width=1.1\textwidth]{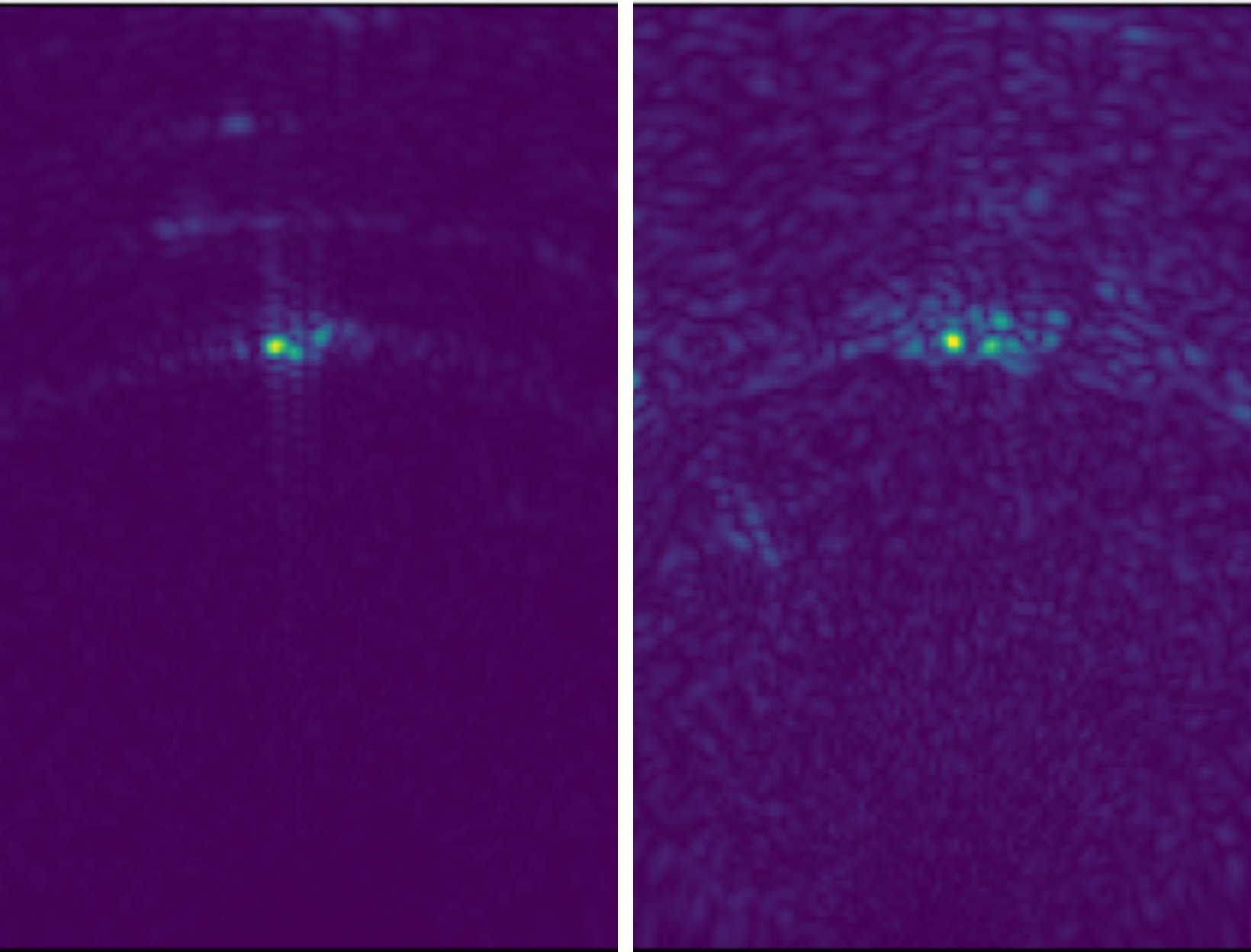} \\
            \vspace{-0.4cm}
            \includegraphics[width=1.1\textwidth]{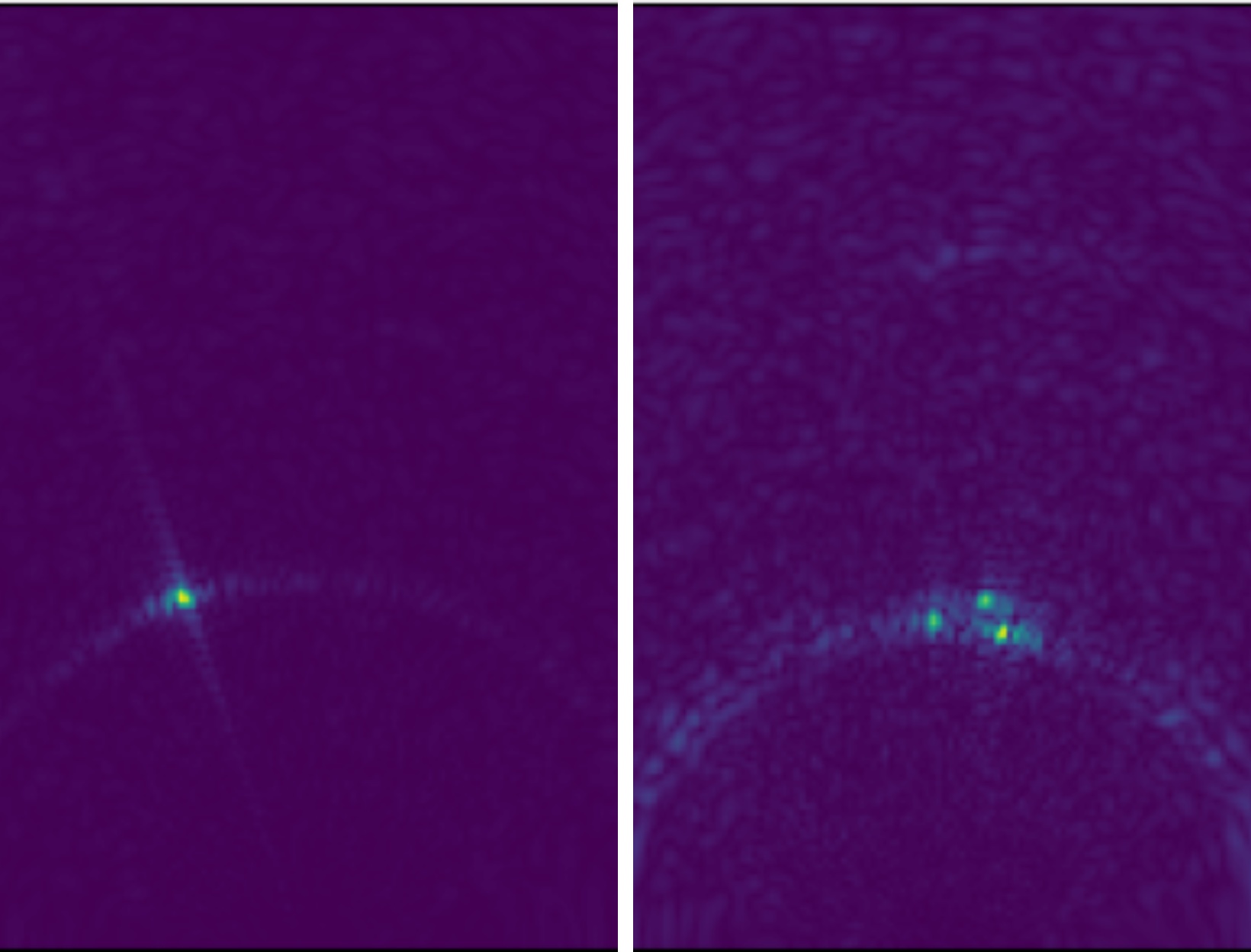}
            \label{problem:radar_frame}
        \vspace{-0.3cm}
        \end{minipage}
        \label{aoa_tof}
    }
    \subfloat[]{
        \begin{minipage}[b]{0.21\columnwidth}
            \includegraphics[width=1.1\textwidth]{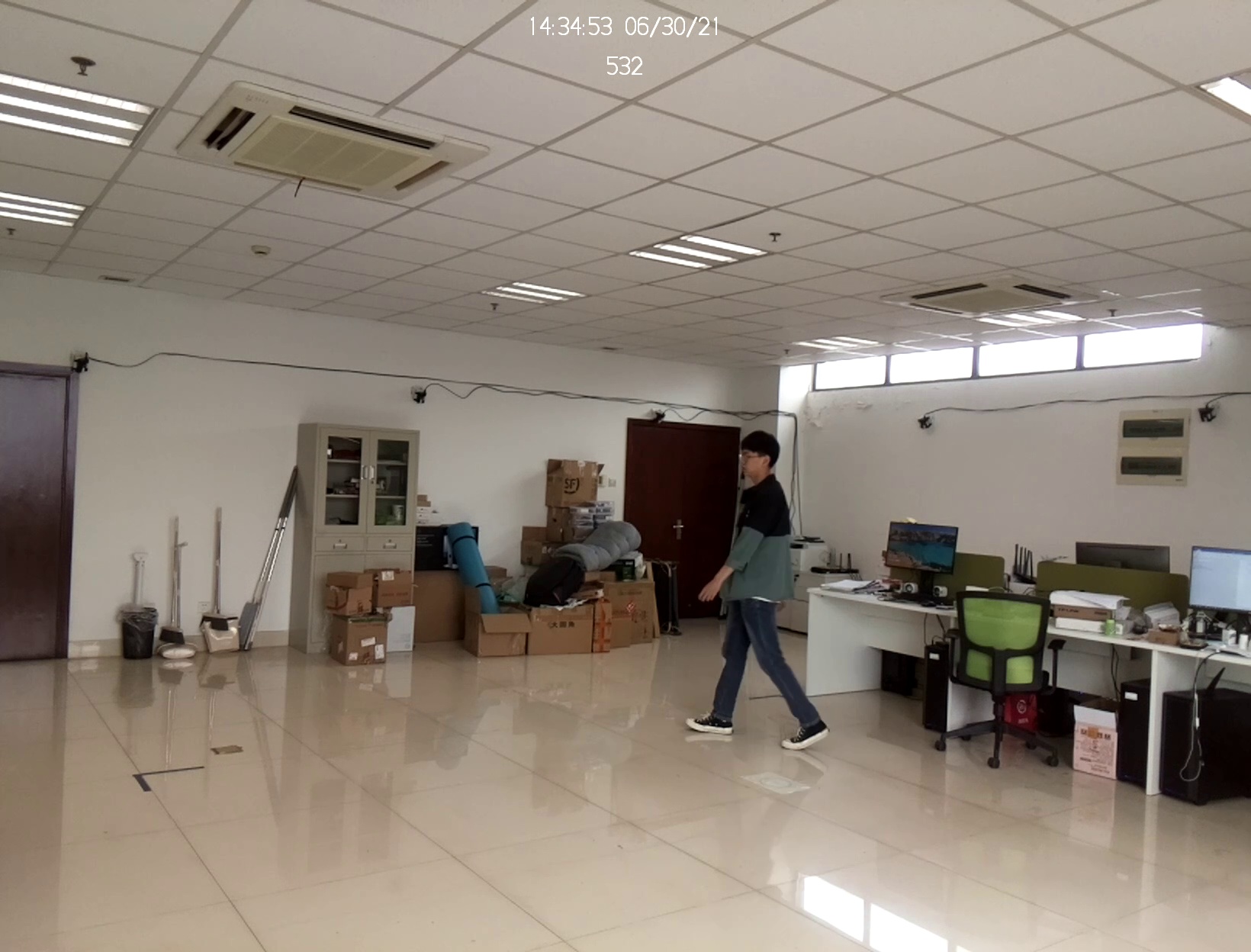} \\
            \vspace{-0.4cm}
            \includegraphics[width=1.1\textwidth]{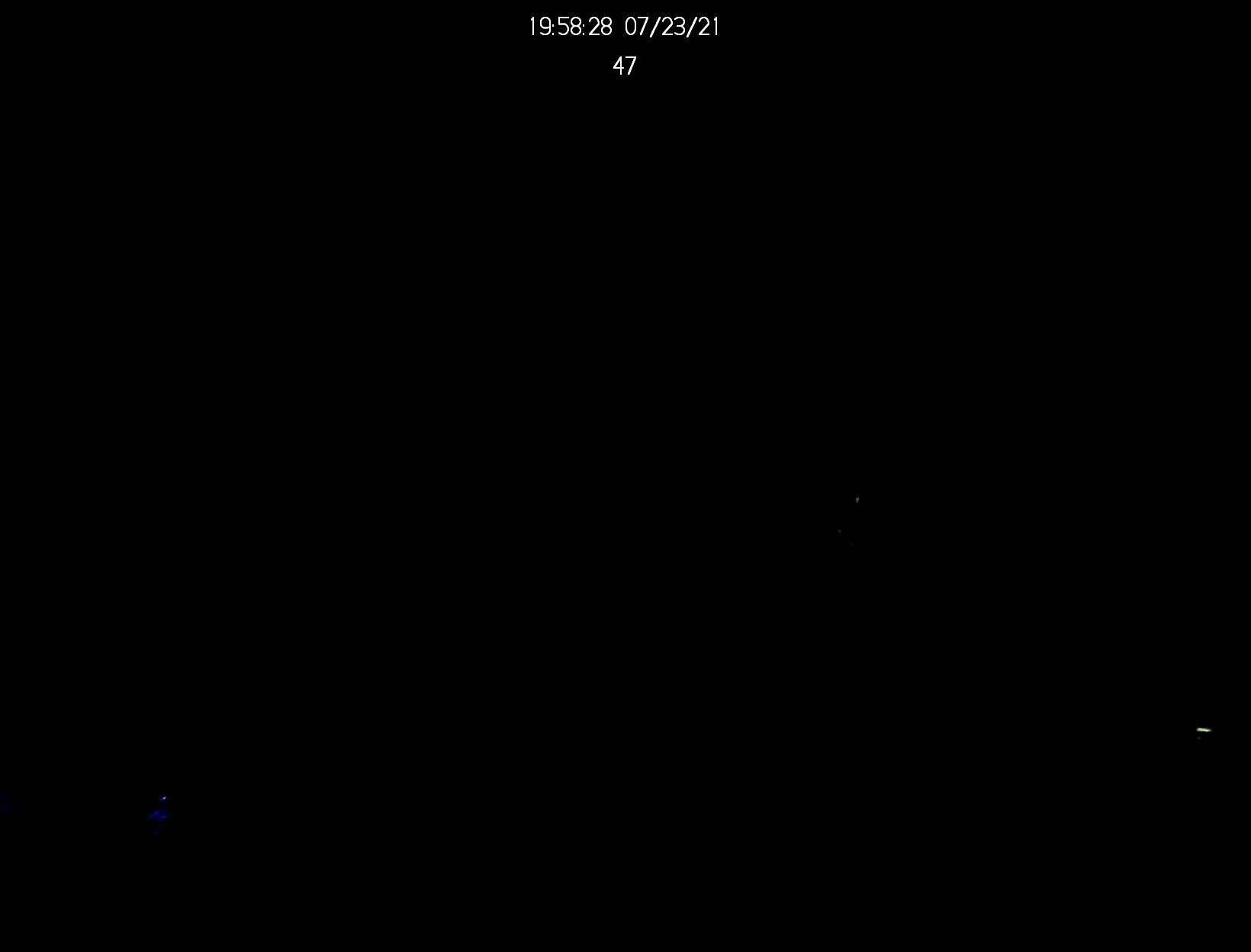} \\
            \vspace{-0.4cm}
            \includegraphics[width=1.1\textwidth]{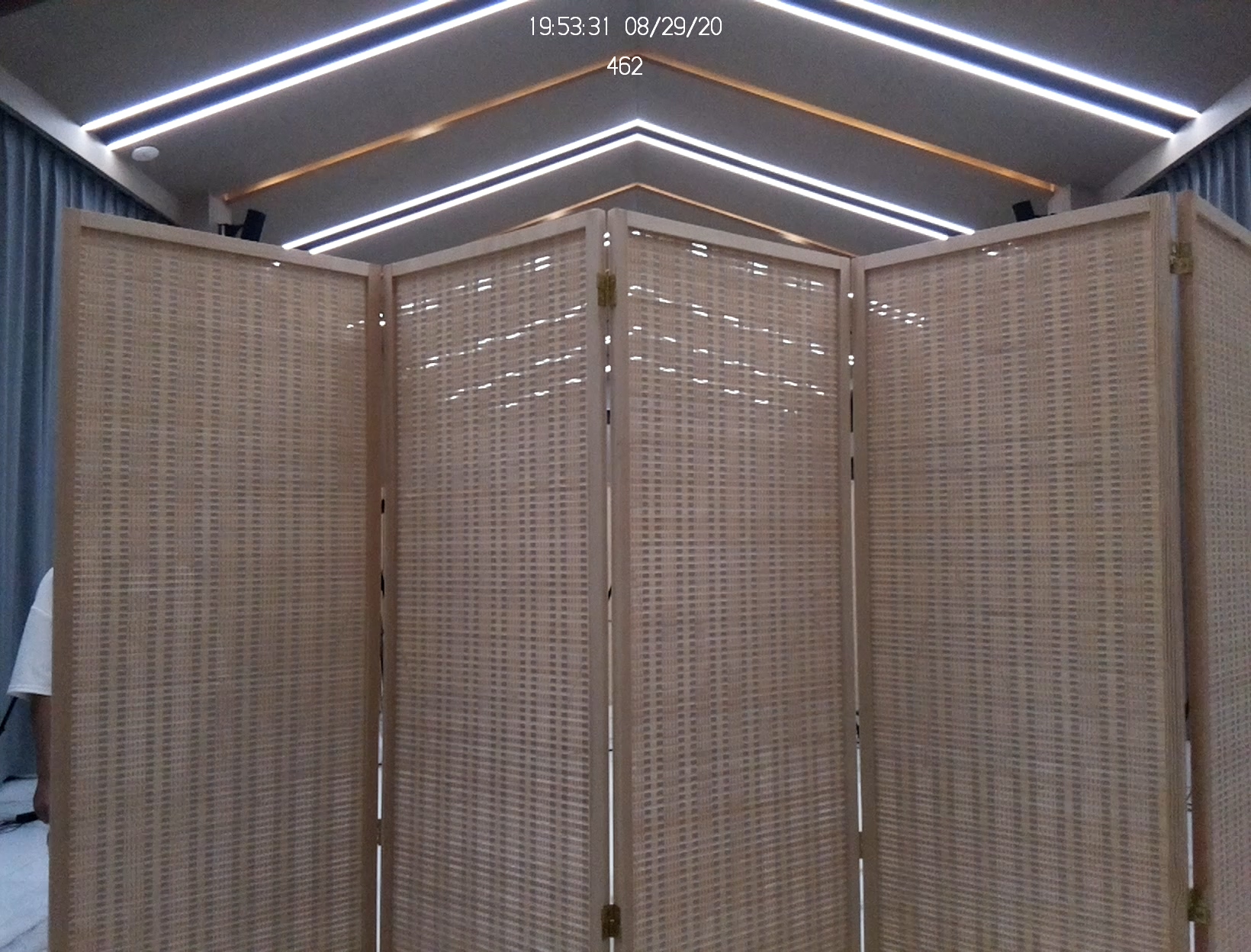}
    		\label{problem:camera}
    	\vspace{-0.3cm}
        \end{minipage}
        \label{rgb_images}
    }
    \subfloat[]{
        \begin{minipage}[b]{0.21\columnwidth}
            \includegraphics[width=1.1\textwidth]{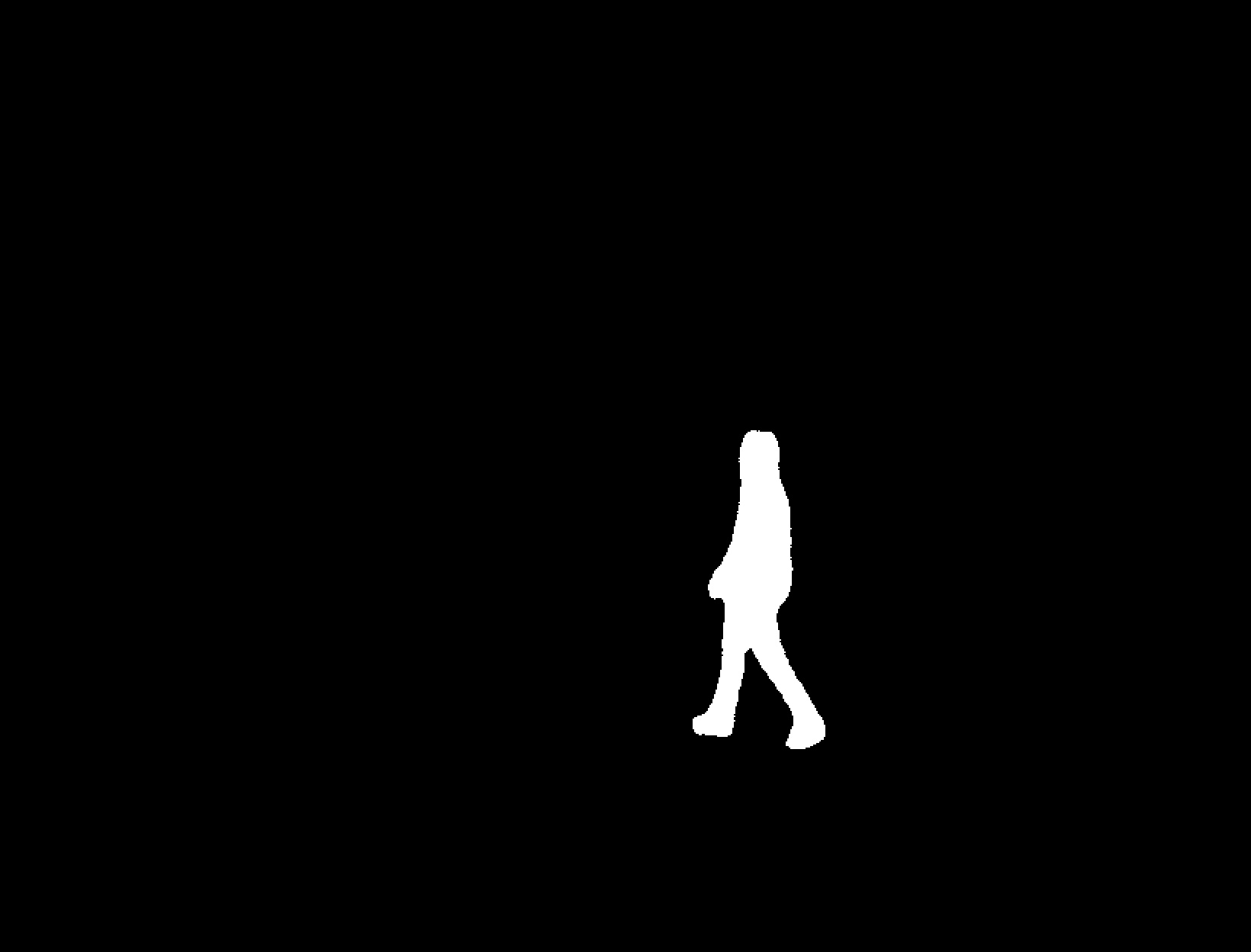} \\
            \vspace{-0.4cm}
            \includegraphics[width=1.1\textwidth]{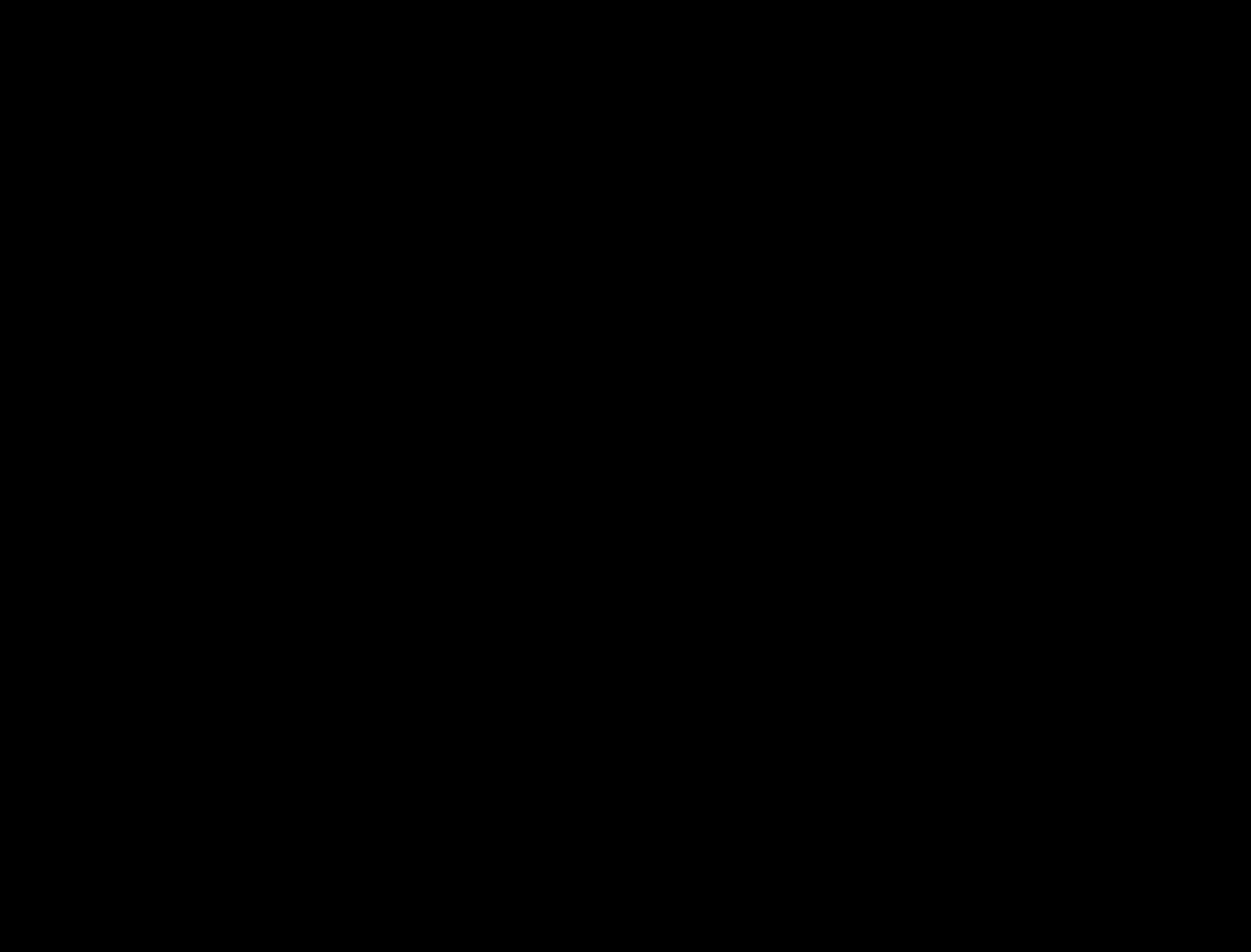} \\
            \vspace{-0.4cm}
            \includegraphics[width=1.1\textwidth]{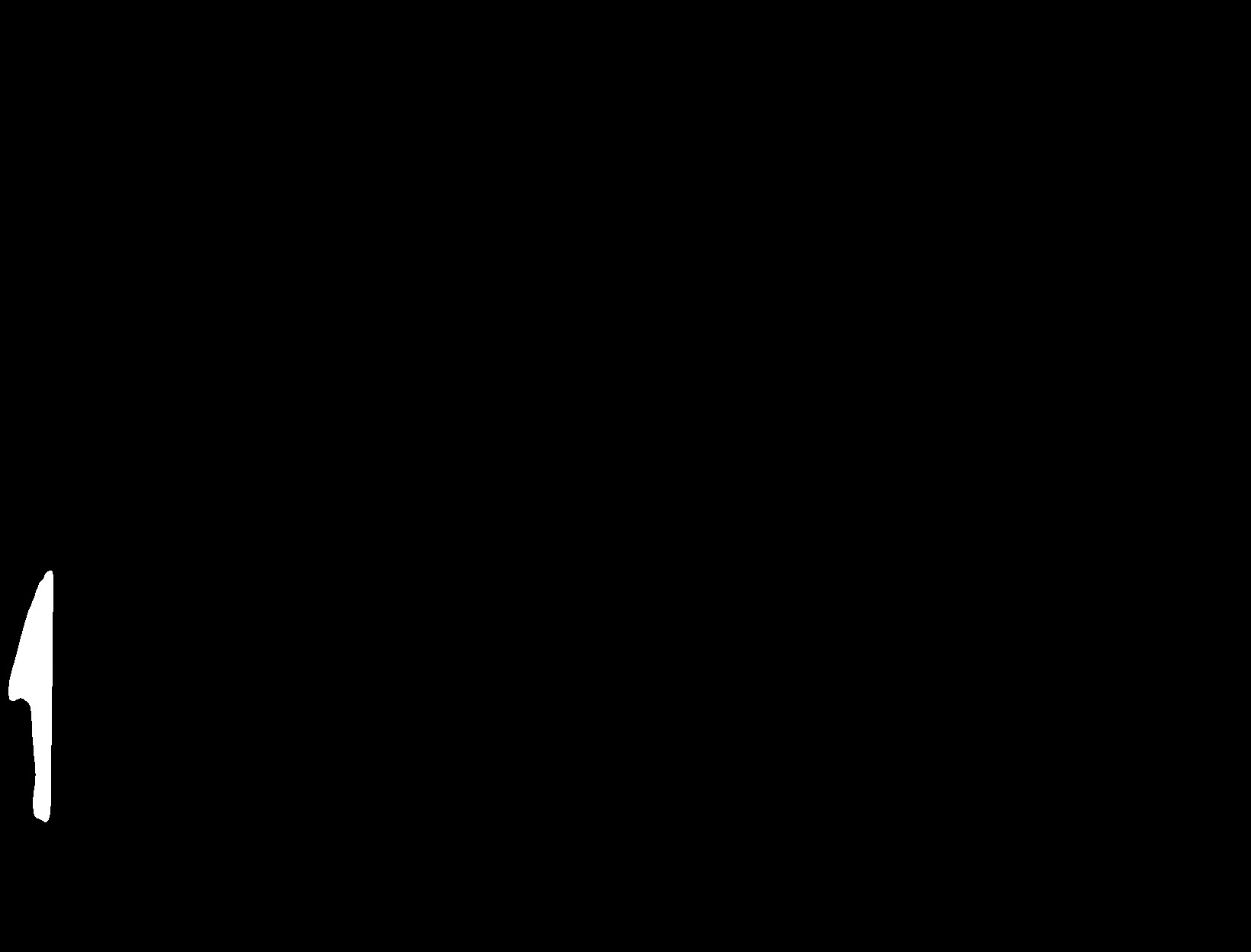}
    		\label{problem:gt}
    	\vspace{-0.3cm}
        \end{minipage}
        \label{maskrcnn_out}
    }
    \subfloat[]{
        \begin{minipage}[b]{0.21\columnwidth}
            \includegraphics[width=1.1\textwidth]{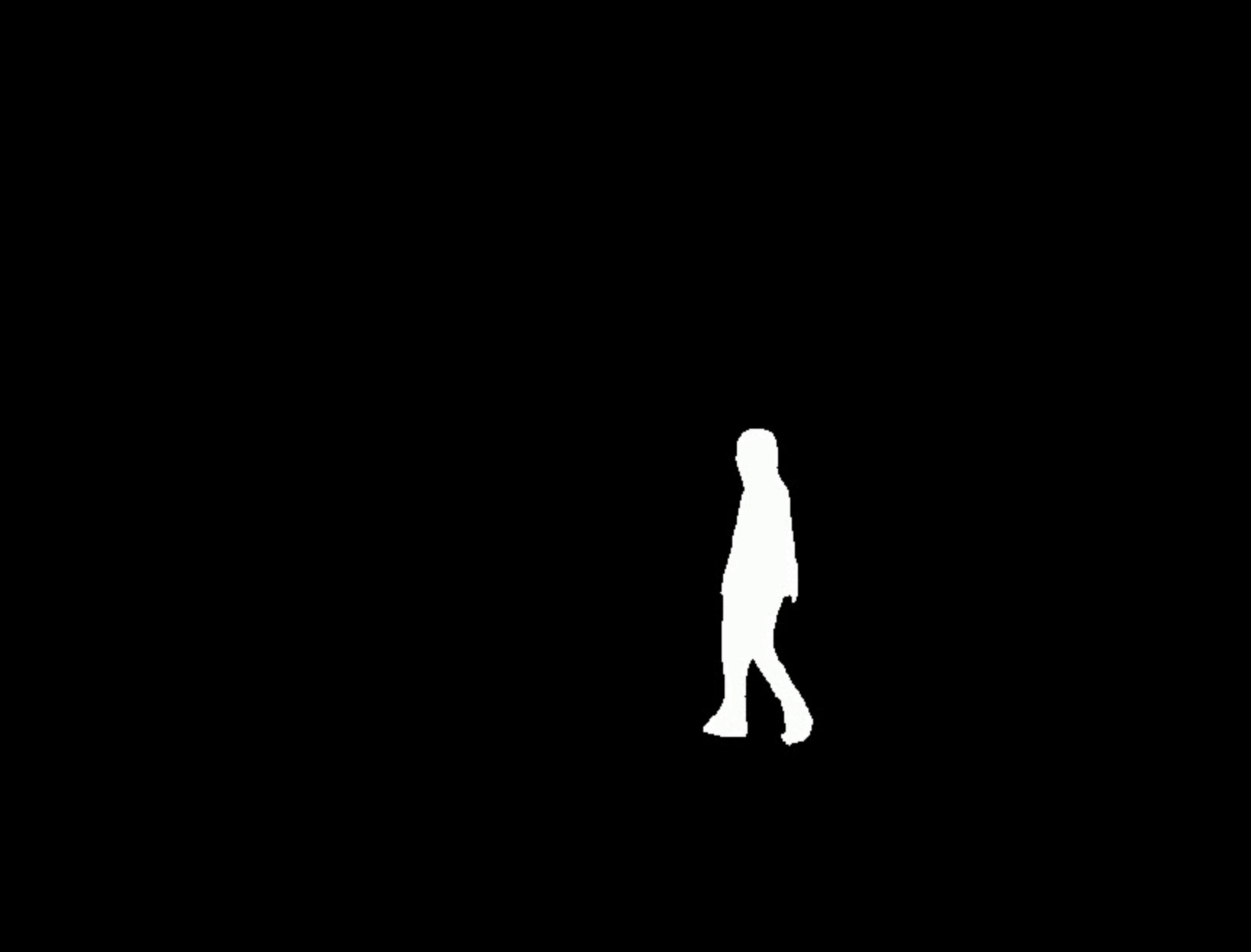} \\
            \vspace{-0.4cm}
            \includegraphics[width=1.1\textwidth]{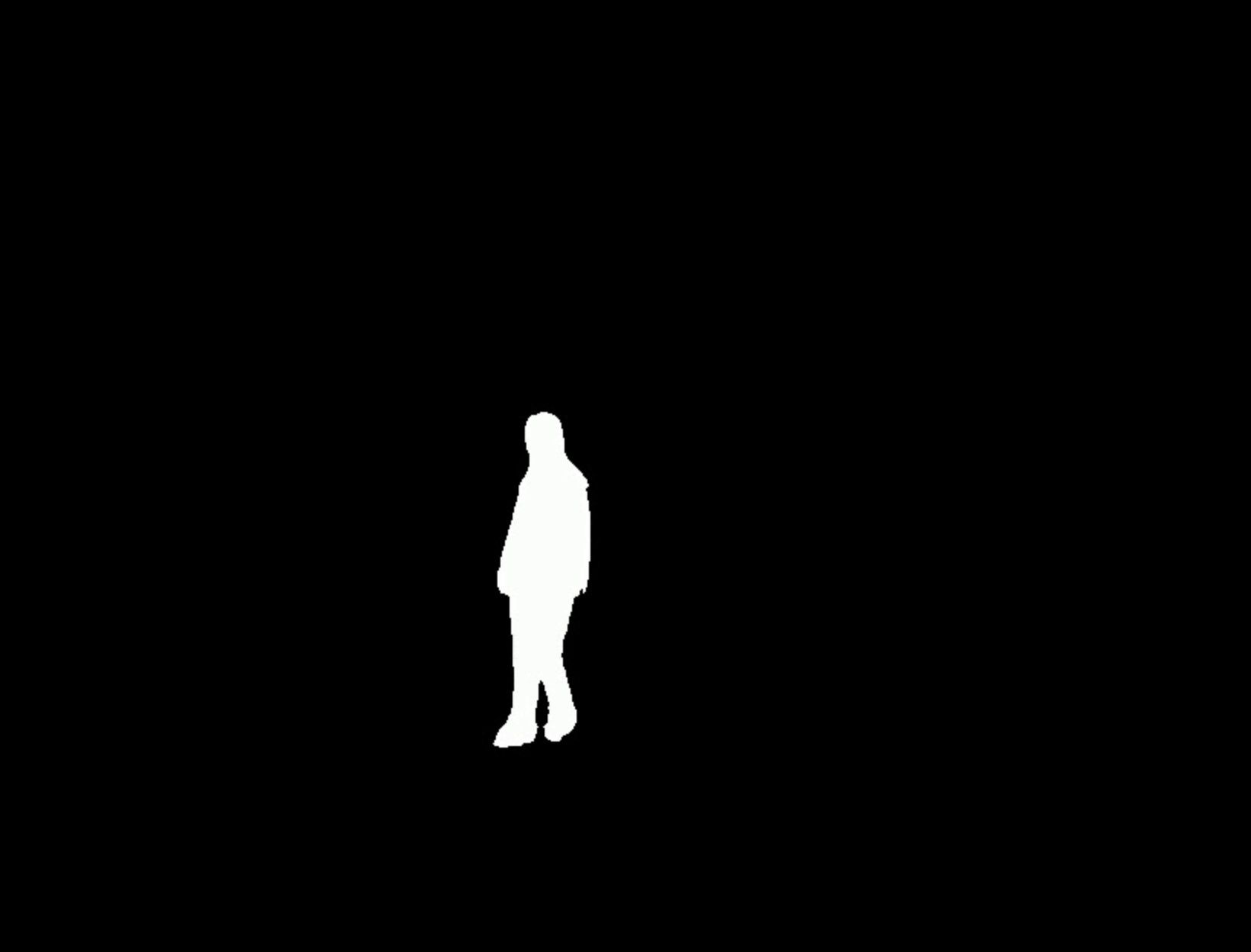} \\
            \vspace{-0.4cm}
            \includegraphics[width=1.1\textwidth]{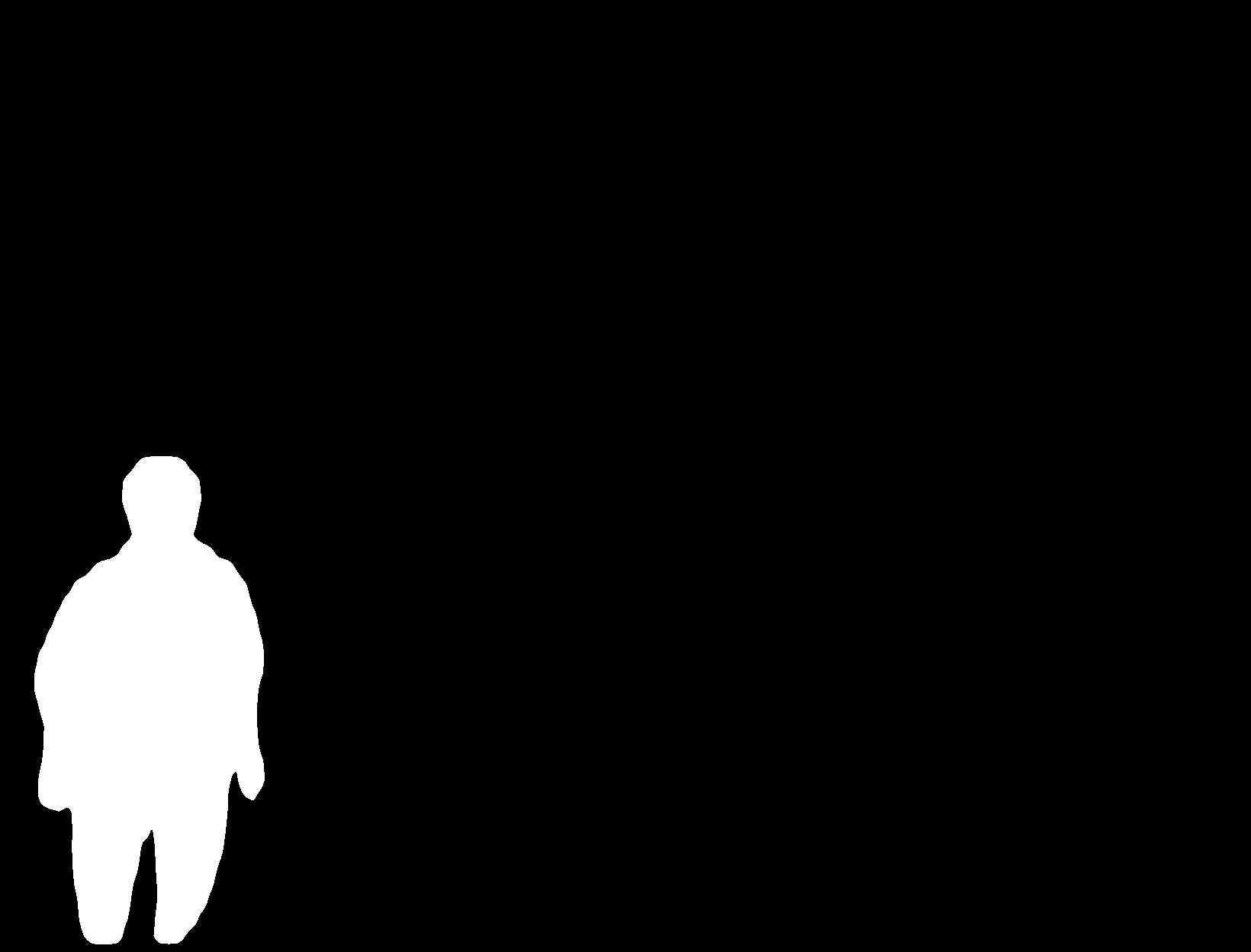}
    		\label{problem:pred}
    	\vspace{-0.3cm}
        \end{minipage}
        \label{rfmask_out}
    }
	
    \hspace{-1cm}
    \caption{RFMask segments the human silhouette purely based on millimeter wave radar even under the occlusion and/or low illumination conditions: (a) radar frames; (b) camera frames; (c) the segmentation results of vision based methods; (d) the segmentation results of the proposed RFMask. The first row corresponds to the scene with sufficient illumination without occlusion, where we can extract human silhouette based on camera frames or millimeter wave radar. The second and third row correspond to the scenes under low illumination and occlusion, where we can still extract human silhouette from millimeter wave radar while it fails for camera frames}
    \label{head_pic}
    \vspace{-3mm}
\end{figure}

\section{Introduction}
\label{introduction}


\IEEEPARstart{S}{emantic} segmentation is a fundamental task in computer vision, which enables various applications including surveillance, activity recognition, autonomous driving, etc \cite{lazarow2020learning, mohan2021efficientps}. 
The problem can be formulated as pixel-level classification with semantic labels (e.g., human, car, bicycle, background). 
Human silhouette segmentation acts as one of the most important part of semantic segmentation, which enables many applications including intrusion detection, human re-identification, smart systems, virtual reality, and crowd counting. 
With the rapid developing of deep neural networks, promising results have been achieved in the past decade \cite{kirillov2020pointrend, cheng2020panoptic}.
However, a fundamental limitation of these vision based methods lies in the fact that optical cameras could not capture fine-grained object information under occlusion or low illumination conditions. 
To resolve this challenge, existing methods focus on figuring out a novel way of fusing multi-level features and/or combining global information to improve the performance under occlusion, as well as augmenting the quality of images with adversarial strategy to improve the performance under dim environment.
These methods, however, still fail when encountering heavily occlusion (completely unseen) and/or dark environment (complete darkness). Such failure is due to the fact that visual perception is fundamentally an ill-posed problem under occlusion or low illumination conditions.   

In this paper, we propose a completely different RF-based approach to deal with occlusion and low illumination. 
Specifically, radio frequency (RF) signals are unaffected by lighting conditions which could traverse occlusions and reflected by human body.   
Recent progress in wireless sensing utilizes RF signals to achieve human detection, vital signs estimation and trajectory tracking 
\cite{gong2016adaptive, gong2017robust, yu2021wifi}. 
However, existing systems mainly focus on extracting coarse-grained information from RF signals including location, velocity, etc. 
Compared with existing works, generating vision-like silhouette results from RF signals is a completely different task. 
The core challenge lies in the discrepancy between vision and RF modality. 
Specifically, the challenge comes from three aspects. (1) Low Resolution: RF signals have much lower spatial resolution compared with optical cameras, which makes it extremely difficult to distinguish signals from different human body parts. (2) Specular Reflection: RF signals would encounter specular reflections on human body, which makes the received signal only contains partial information of the whole body. (3) View Transformation: the camera imaging plane is perpendicular to the RF signal plane, which makes it difficult to generate vision-like results from RF signals without elaborated design.  
Therefore, using wireless signals to generate detailed and accurate segmentation results similar to what has been achieved in computer vision, remains a challenge.

\begin{figure*}[ht]
    \centering
	\includegraphics[width=1\textwidth]{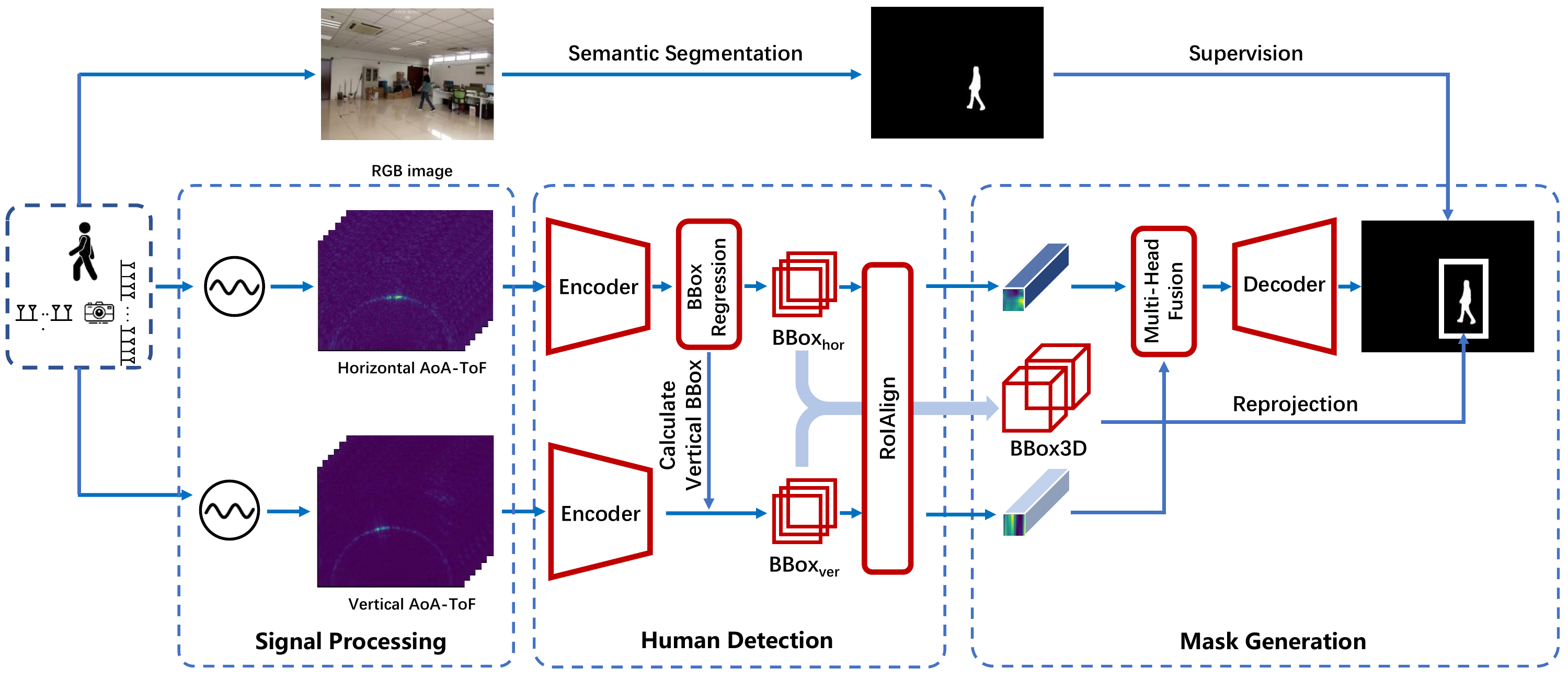}
    \caption{The architecture of RFMask. It consists of three components: signal processing, human detection, and mask generation. The whole model is trained in an end-to-end manner.}
    \label{framework}
    \vspace{-3mm}
\end{figure*}

To achieve human silhouette segmentation from RF signals, we propose RFMask, a RF-based framework that parses RF signals to extract accurate human silhouette, which is robust even under occlusion or low illumination conditions.  
RFMask utilize two FMCW radars to transmit low power wireless signals and listen the reflections from human and environment. The whole process can be divided into three steps corresponding to three elaborately designed modules: signal processing module, human detection module, and mask generation module. First, RF signals captured by millimeter wave radars are transformed into AoA-ToF(angle of arrival, time of flight) heatmaps in spatial domain by signal processing module; Next, human detection module is performed to locate human reflections on RF heatmaps and extract features from surrounding signals; Finally, the extracted features are aggregated by multiple multi-head attention layers and then decoded into resulting space, namely mask generation module.

Figure \ref{rfmask_out} demonstrates an example output of RFMask. 
Our signal processing module decodes spatial information from raw signals which only contains time variations of reflections. 
The following human detection module aims at dealing with sparsity characteristics of RF signals, which filters out noise and enables the following mask generation module to focus on surrounding signals that contain human action information. The well-designed mask generation module not only takes into account the spatial relationship of two signal planes, but also adapt to the fragmentation characteristic of RF signal by utilizing multi-head attention to calculate attention weights between frames automatically.



Different from computer vision based algorithms, the training process of RFMask is difficult due to the lack of labeled data.
The incomprehensible characteristic of RF signals also makes it impossible to annotate manually. To overcome this challenges, inspired by \cite{zhao2018through}, we adopt cross-modal supervision. Specifically, one optical camera is attached to the FMCW radars when capturing data. Segmentation results extracted from RGB images are utilized as supervision. When the training procedure of RFMask has been finished, it can be tested using RF signals only. Even though RFMask has not seen occlusion and low light examples during training, it still performs well in these challenging scenarios.
Except for cross-modal supervision, we also account for the spatial relationship of two radars, the multi-path effect of RF signals, and the discontinuity characteristic of human reflections.

Our proposed RFMask is trained and evaluated on our own collected dataset. The dataset contains hundreds of thousands of RF frames (804,760 radio frames) as well as corresponding human keypoints and segmentation annotations. The data is captured under the scenarios of different number of persons performing diverse activities: stand, walk, sit, squat. To demonstrate the performance of RFMask, we split our dataset into single-person (walking with only one person in the scene), multi-person (walking with multiple persons in the scene), and action subsets (acting various actions in the scene). The experiment results show that RFMask achieves mask IoU of 0.706, 0.711, 0.705 on single-person, multi-person, and action subset, respectively. The qualitative results also demonstrates the performance under challenging scenarios.

Our main contributions can be summarized as follows.
\begin{enumerate}
	\item To the best of our knowledge, our proposed RFMask framework is the first attempt to generate human silhouette results from millimeter wave radio signals.
	\item We propose a systematic framework which is composed of a signal processing module, a human detection module, and a mask generation module to achieve human silhouette segmentation from RF signals. 
	\item We create a multi-modal dataset that contains thousands of radio frames and corresponding optical camera images of human activity. Meanwhile, our dataset provides various forms of ground-truth including 2D/3D human skeletons and human silhouette results, etc. Our dataset and codes will be released in public. 
\end{enumerate}

The rest of this paper is organized as follows. In Section \ref{related_work}, we will review the recent development of image segmentation and wireless sensing. In Section \ref{rfmask}, we will describe the details of our proposed RFMask, including data processing pipeline, model structure, and loss function. In Section \ref{experiment}, the performance of RFMask is evaluated, both quantitative and qualitative results are demonstrated, we also conduct ablation experiments to show the effectiveness of our key components. Finally, in Section \ref{conclusion}, we briefly summarize our entire work.

\section{Related Work}
\label{related_work}
\subsection{Image Segmentation}
Image segmentation is always being a fundamental task in computer vision since the early days of the field. It can be formulated as a problem of pixel-level classification with semantic labels(semantic segmentation), 
or partitioning of individual objects (instance segmentation), or both(panoptic segmentation).
In the past decade, promising results have been achieved with the emergence of deep learning. 

Specifically, Long et al. proposed Fully Convolutional Networks (FCN) 
which utilizes convolutional layers only and outputs segmentation map with the same size as the input image \cite{long2015fully}.  
The skip connections between the final and previous layers make the feature fusion process more efficient, and enable FCN to output accurate and detailed segmentation results.
Based on FCN, Liu et al. proposed ParseNet \cite{liu2015parsenet} by adding a global context feature to FCN to augment the features at each location that enables the model to account for global information in an efficient manner. Badrinarayanan et al. proposed SegNet \cite{badrinarayanan2017segnet}, a symmetrical fully convolutional encoder-decoder architecture to achieve semantic segmentation. The main contribution of SegNet is that the upsample layers of the decoder uses pooling indices computed in the max-pooling step of the counterpart layer in encoder to perform nonlinear interpolation.  Inspired by FCN, He et al. combined Faster R-CNN with FCN, where an additional mask branch was added to Faster R-CNN to generate segmentation mask for each instance \cite{He_2017_ICCV}. Except for constructing basic model structures, many researchers are seeking novel ways of fusing features of different levels and/or stages. 
Lin et al. proposed Feature Pyramid Network (FPN) \cite{Lin_2017_CVPR}, which utilizes inherent multi-scale features to build feature pyramids for the purpose of merging different level features. Similarly, Pyramid Scene Parsing Network \cite{zhao2017pyramid} was proposed by Zhao et al. to fuse features with different scales. They pooled extracted features into four different scales which were than concatenated with original feature maps to capture local and global information. Inspired by FPN, Liu et al. proposed Path Aggregation Network (PANet), where a new bottom-up pathway is introduced in FPN backbone to improve the propagation of lower-level features and lateral connections are added between the corresponding stage of top-down and bottom-up pathway. Attention mechanisms have also been explored. For instance, in Pyramid Attention Network, proposed by Li et al. \cite{li2018pyramid}, attention mechanisms and spatial pyramids were combined to extract precise dense features for pixel labeling. Similar studies can also be found in \cite{zhang2020resnest, choi2020cars}.

One of the core limitation of existing methods lies in the fact that they all take optical images as input, which fails to handle occlusion and/or low illumination due to the fundamental limit of visible light. 
On the contrary, our proposed framework takes RF signal as input, which has completely different characteristics and can be utilized to break the limitation of existing methods.

\subsection{Wireless Sensing}
With the rapid development of sensing technology over the past two decades, quite a lot attention has been drawn on wireless sensing \cite{zhang2020mtrack, chen2020speednet}. To efficiently achieve human activity recognition, various sensing technologies have been developed to sense human activities. Some researchers utilize Wi-Fi devices to achieve sensing tasks by extracting Channel State Information (CSI) which could indicate movements in the environment. Kosba presented RASID \cite{kosba2012rasid}, a WLAN-based device-free passive localization system, which achieves statistical anomaly detection while adapting to environment changes and provides accurate and robust detection results. Wu et al. proposed DeMan \cite{2015Non}, a WiFi-based human detection framework, which extracts maximum eigenvalues of the covariance matrix from CSI 
information to enhance detection performance. Wang et al. presented CARM \cite{2017Device}, a CSI-based human activity recognition and monitoring system, to deal with human activity recognition task. They built a CSI-speed model to describe how the frequencies of CSI power variations are related to human movement speed and a CSI-activity model to describe the relationship between the movement speeds of different human body parts and a specific human activity.

Compared with WiFi, radar systems could achieve better spatial resolution with larger bandwidth and more antennas, which enables fine-grained sensing.  
Zhao adopted a modified adversarial training strategy to train a deep model with convolutional and recurrent neural networks to extract sleep-specific subject invariant features from wireless signals, which could achieve sleep stage prediction \cite{zhao2017learning}. Since labeling the RF dataset manually is extremely difficult, Zhao et al. proposed a cross-modal supervision method, RFPose, to extract fine-grained information to predict human keypoints under the occlusion\cite{Zhao_2018_CVPR}. Specifically, they adopted a dual-branch encoder followed by a decoder to extract pose information from RF stream and output human pose keypoint heatmaps. Subsequently, Zhao et al. further extended RFPose to predict human 3D keypoints in \cite{rfpose}, by discretizing the space of interests into 3D voxels, and performing voxel-level classification to achieve 3D keypoints prediction. In \cite{hsu2019enabling}, Hsu et al. introduced Marko, a system that automatically collect behavior-related data to analyze how users interact with the environment. In \cite{li2022unsupervised}, Li et al. introduce an unsupervised learning framework along with corresponding RF-specific data augmentation techniques, which learns a high-quality representations and greatly facilitates multiple downstream RF-based sensing tasks.

Compared with existing systems, our work seeks to extract fine-grained information, i.e., human silhouette from wireless signals. 
We use two FMCW radars for transmitting and receiving RF signals to physically deal with occlusion and dim environment.
To eliminate the misalignment between vision and RF modality, we propose a systematic framework including signal processing module,  human detection module, and mask generation module, which enables human silhouette segmentation from RF signals.

\begin{figure*}[ht]
    \centering
    \subfloat[]{
        \begin{minipage}{0.156\linewidth}
            \centering
            \includegraphics[width=1\textwidth]{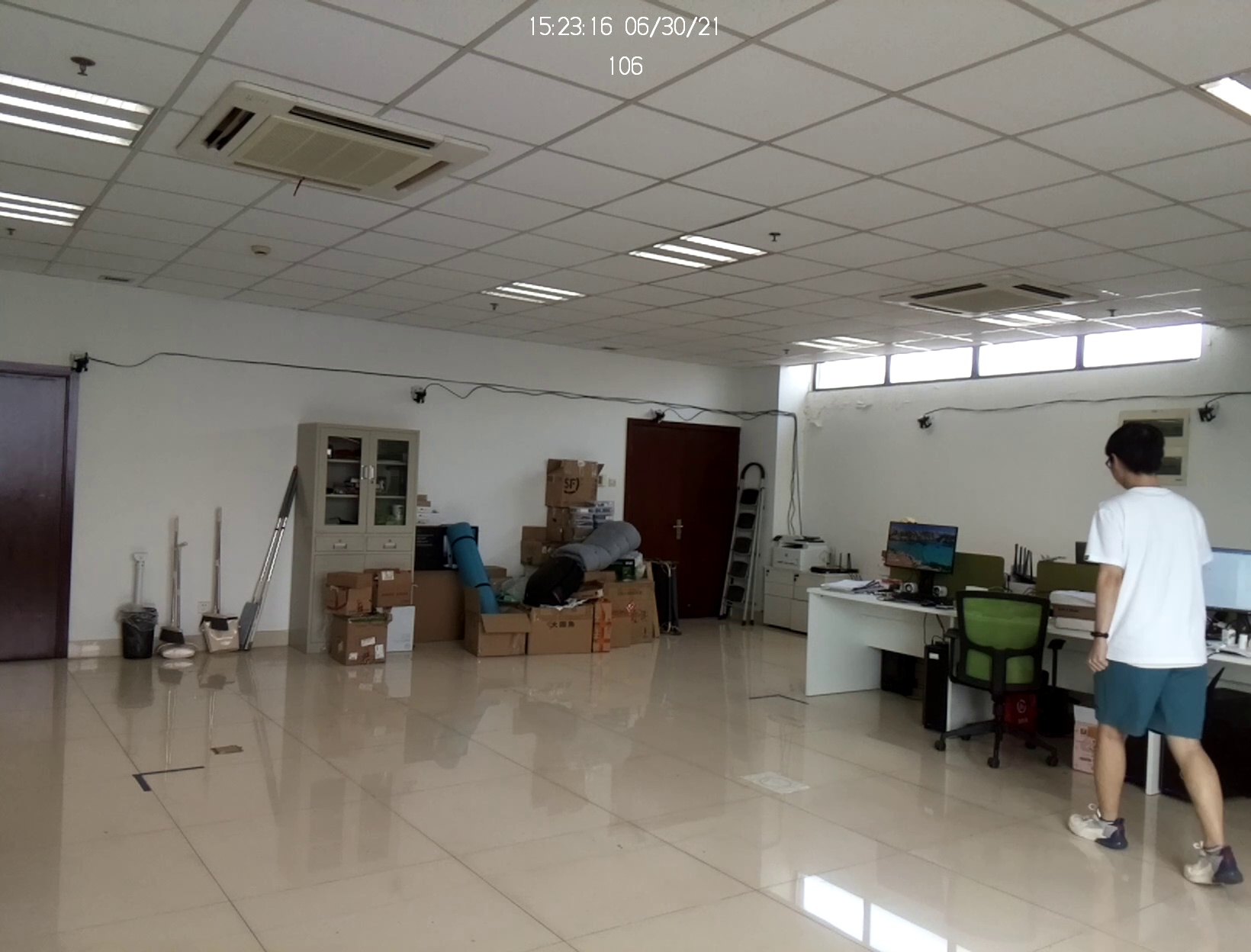}\\
            \vspace{0.03cm}
            \includegraphics[width=1\textwidth]{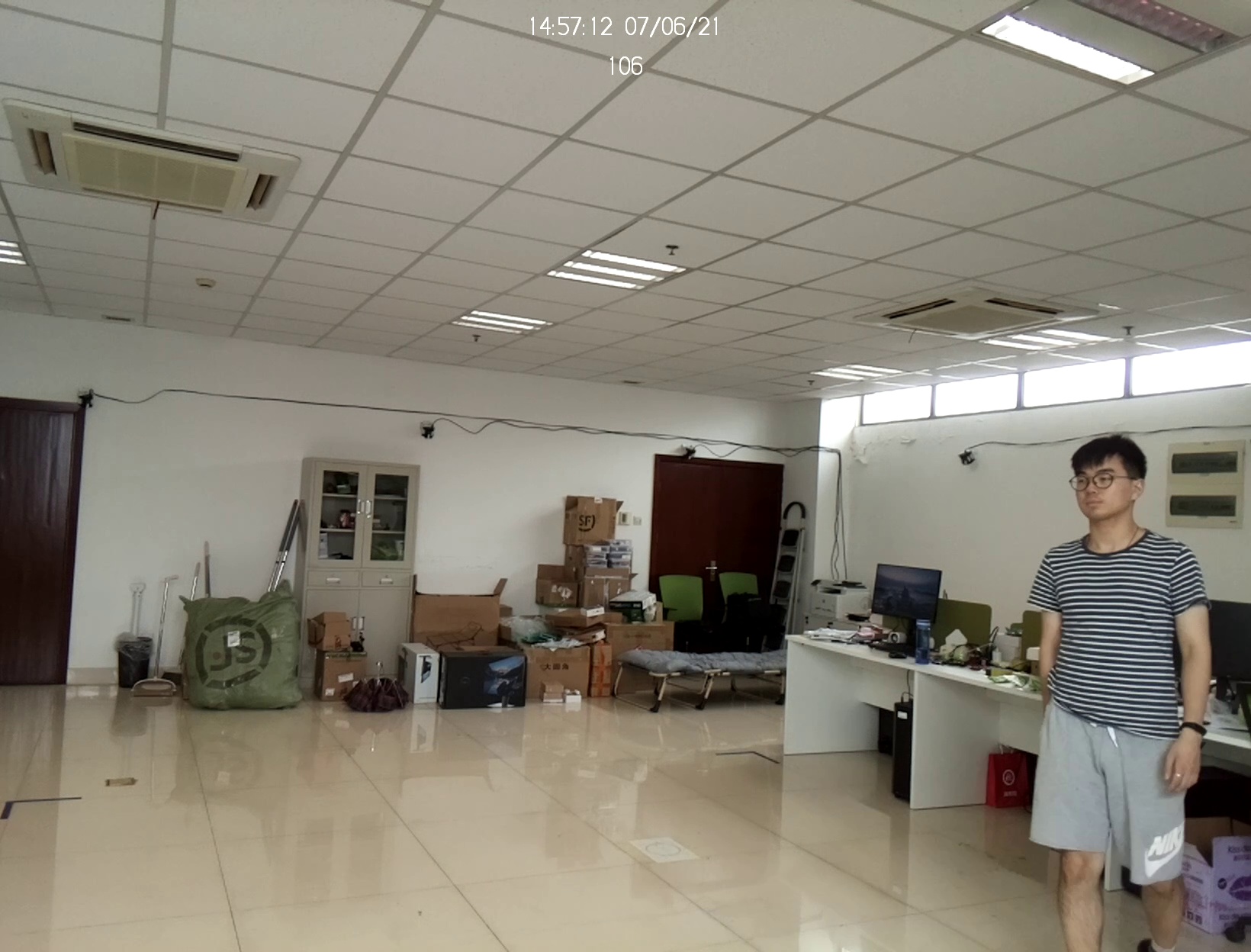}\\
            \vspace{0.03cm}
            \includegraphics[width=1\textwidth]{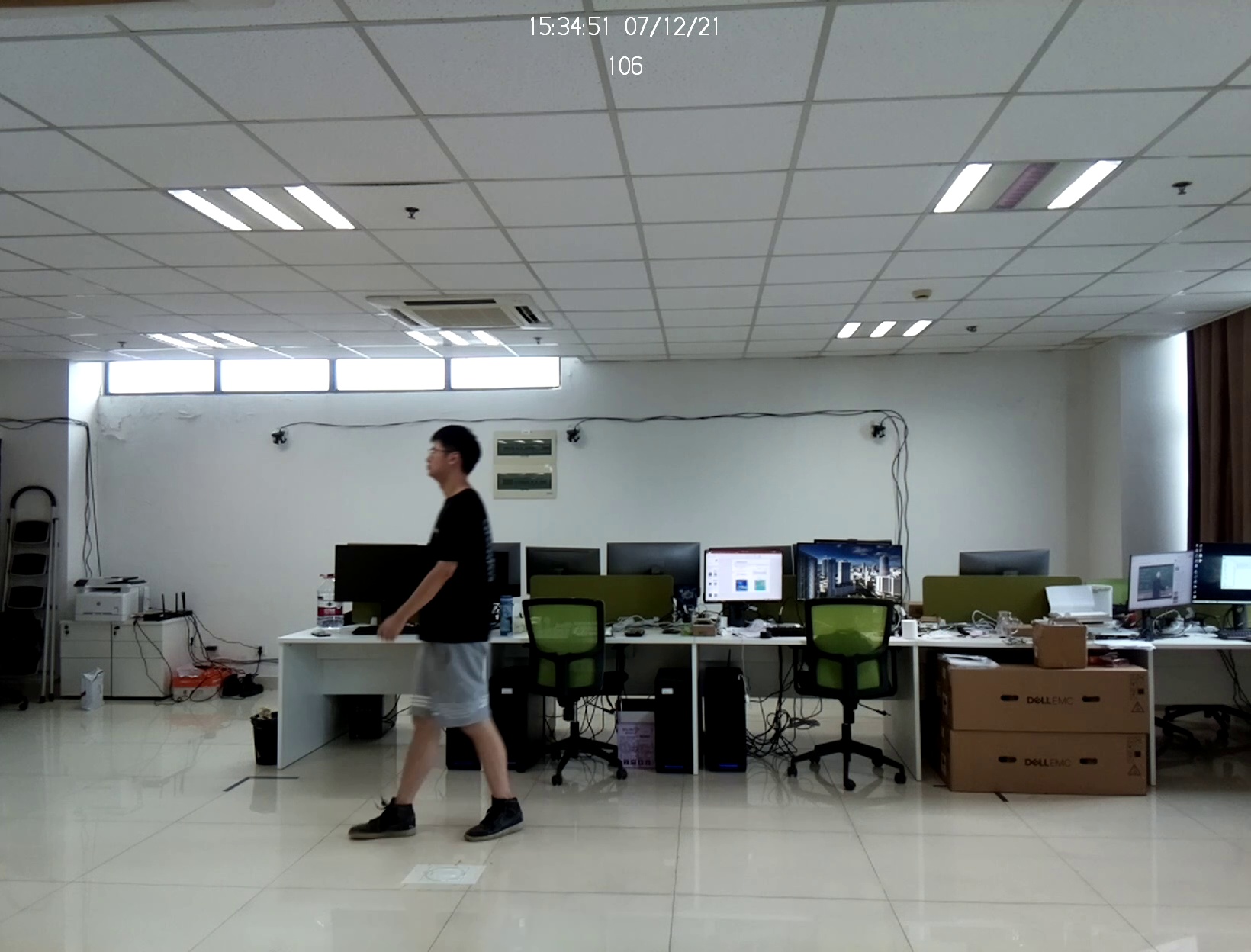}
            \label{sing_walk}
            \vspace{-0.3cm}
        \end{minipage}
        \begin{minipage}{0.156\linewidth}
            \centering
            \includegraphics[width=1\textwidth]{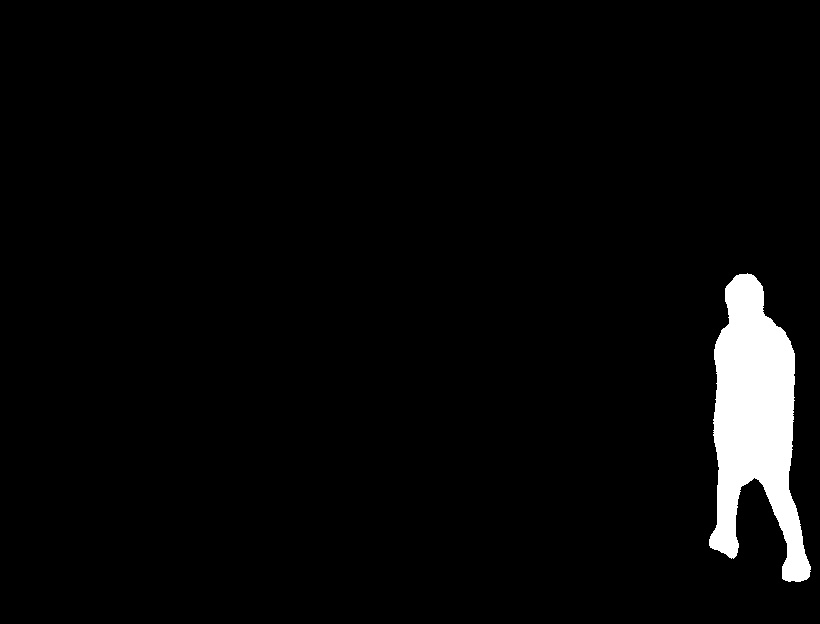}\\
            \vspace{0.03cm}
            \includegraphics[width=1\textwidth]{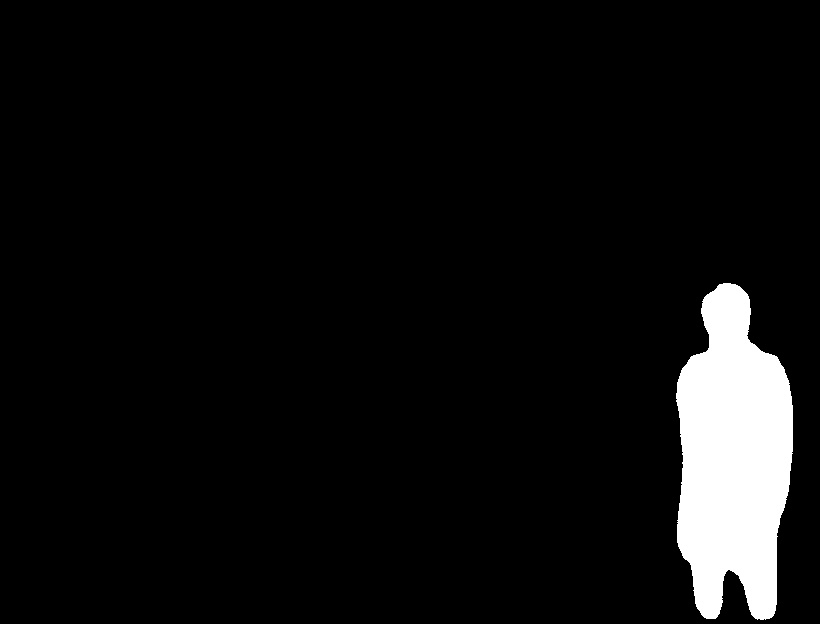}\\
            \vspace{0.03cm}
            \includegraphics[width=1\textwidth]{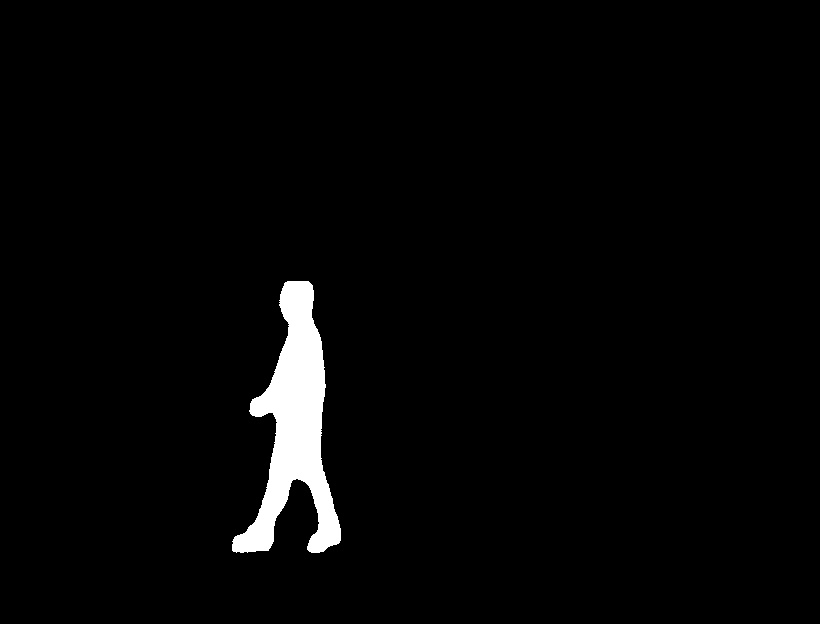}
            \label{sing_walk}
            \vspace{-0.3cm}
        \end{minipage}
    }
    \subfloat[]{
        \begin{minipage}{0.156\linewidth}
            \centering
            \includegraphics[width=1\textwidth]{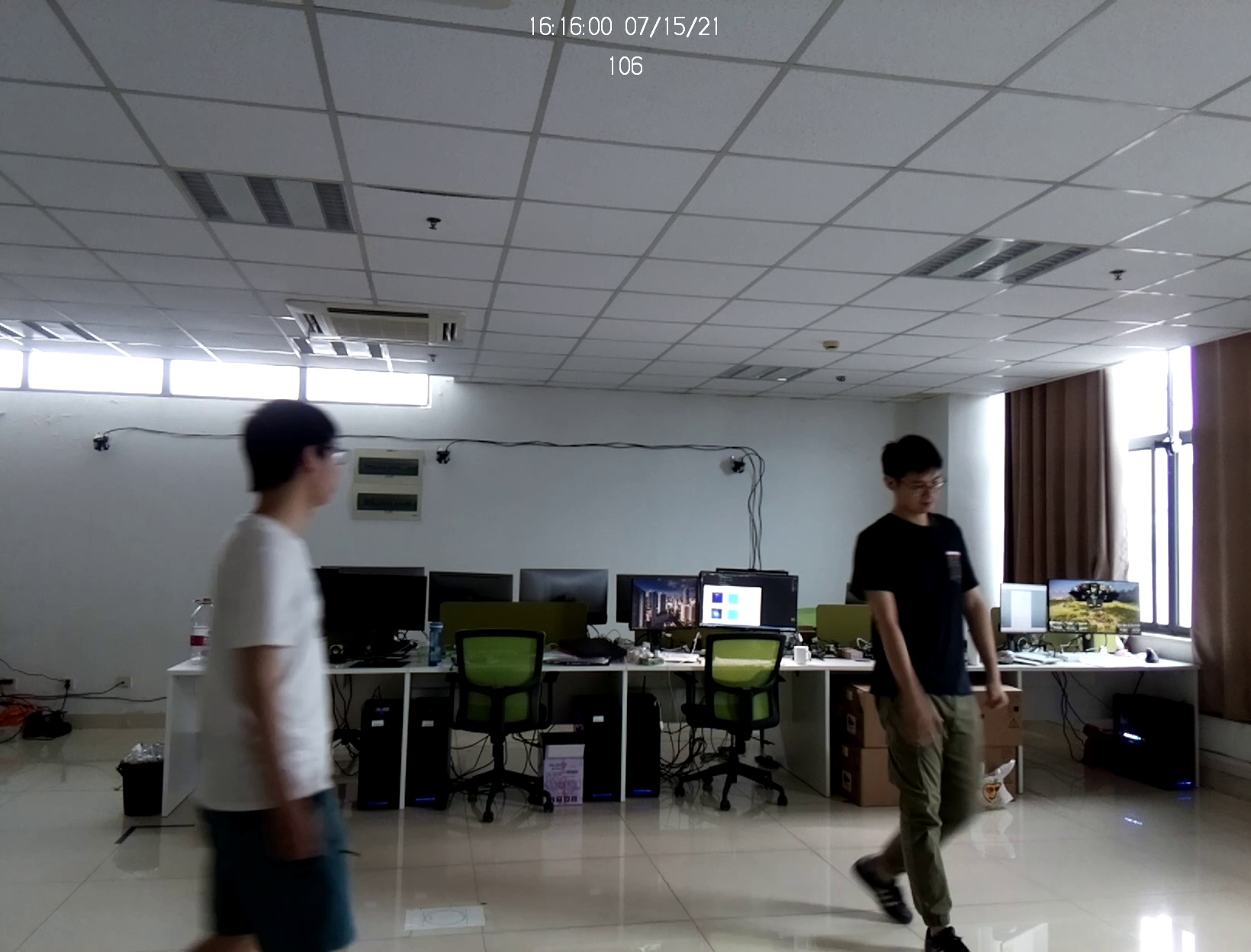}\\
            \vspace{0.03cm}
            \includegraphics[width=1\textwidth]{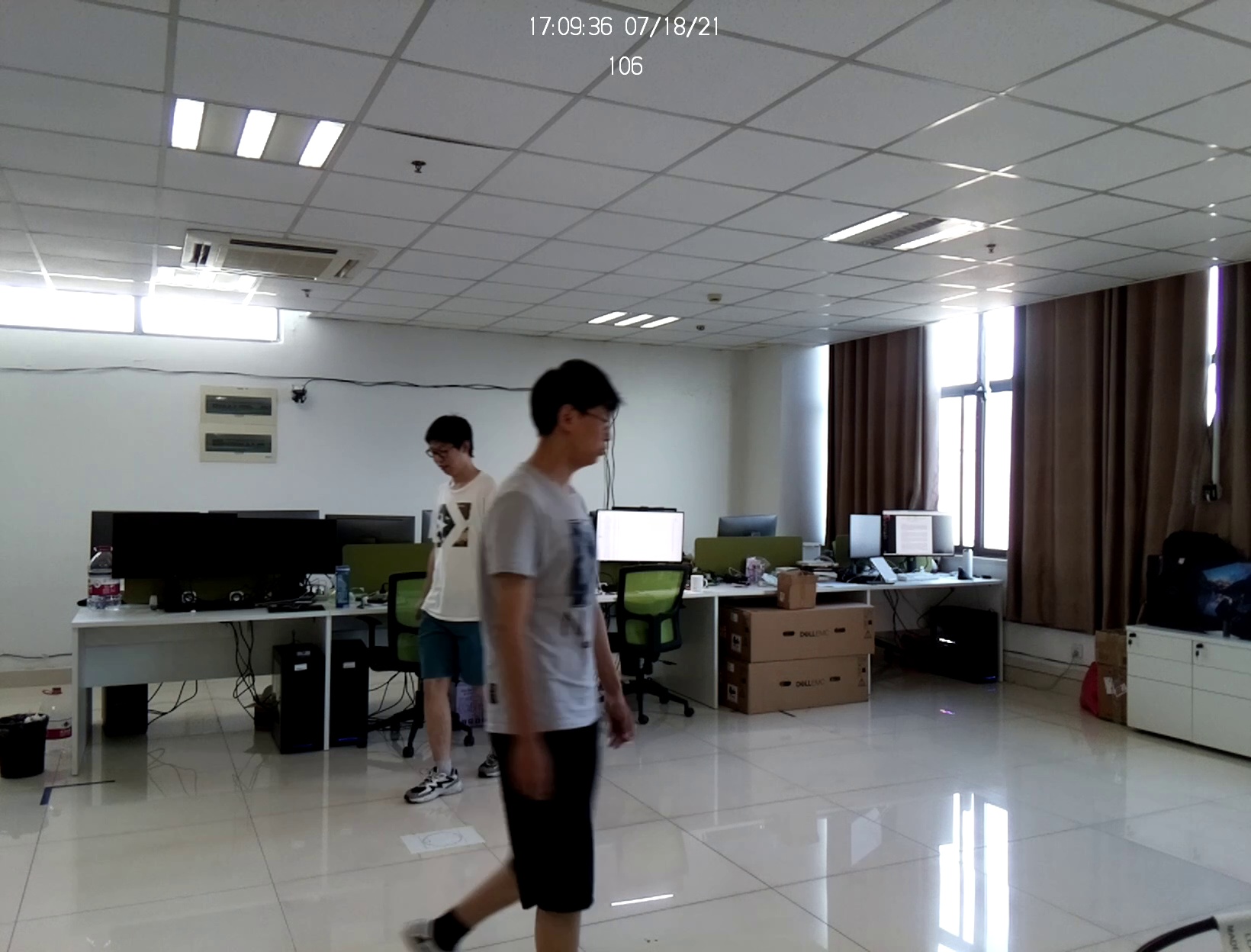}\\
            \vspace{0.03cm}
            \includegraphics[width=1\textwidth]{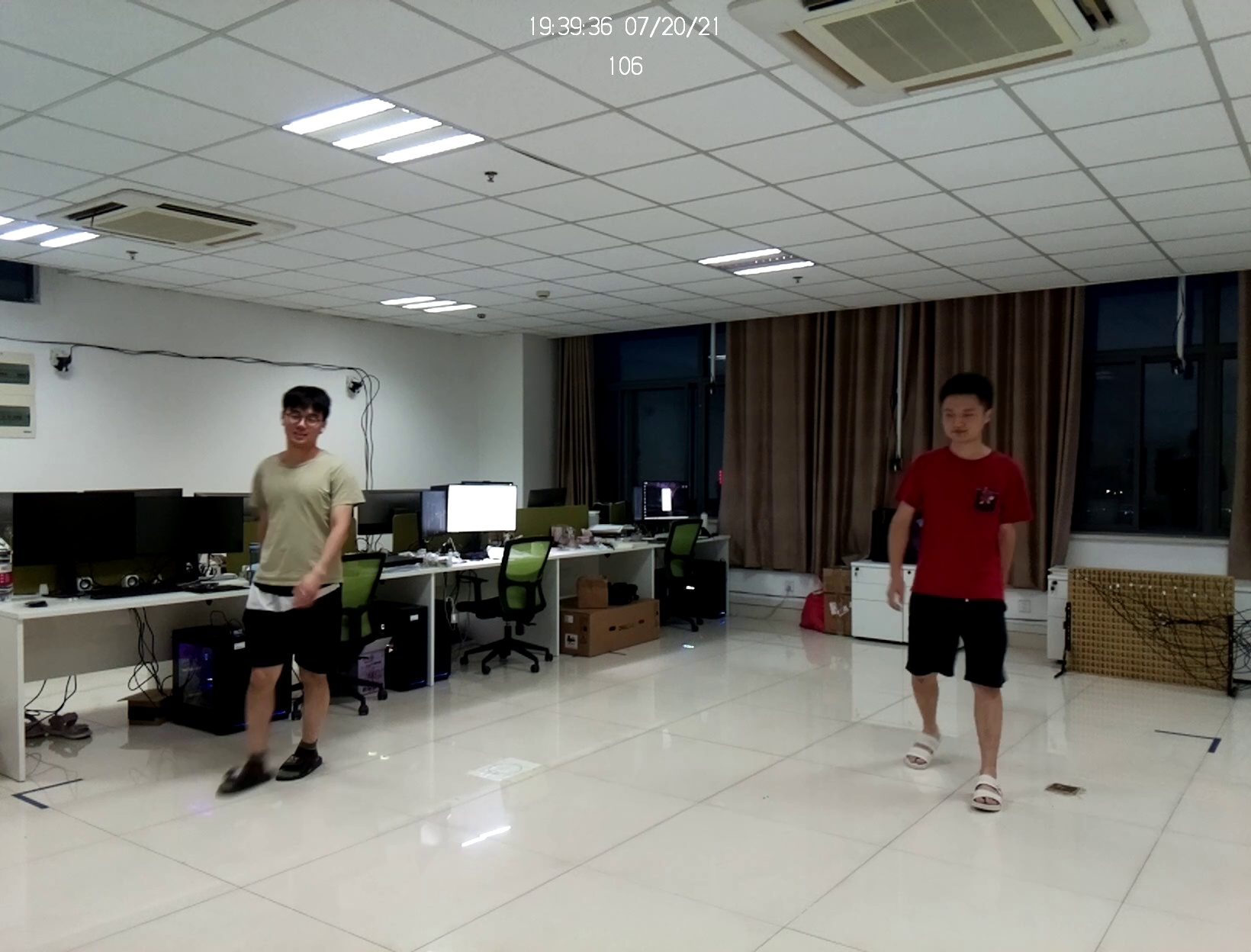}
            \label{multi_walk}
            \vspace{-0.3cm}
        \end{minipage}
        \begin{minipage}{0.156\linewidth}
            \centering
            \includegraphics[width=1\textwidth]{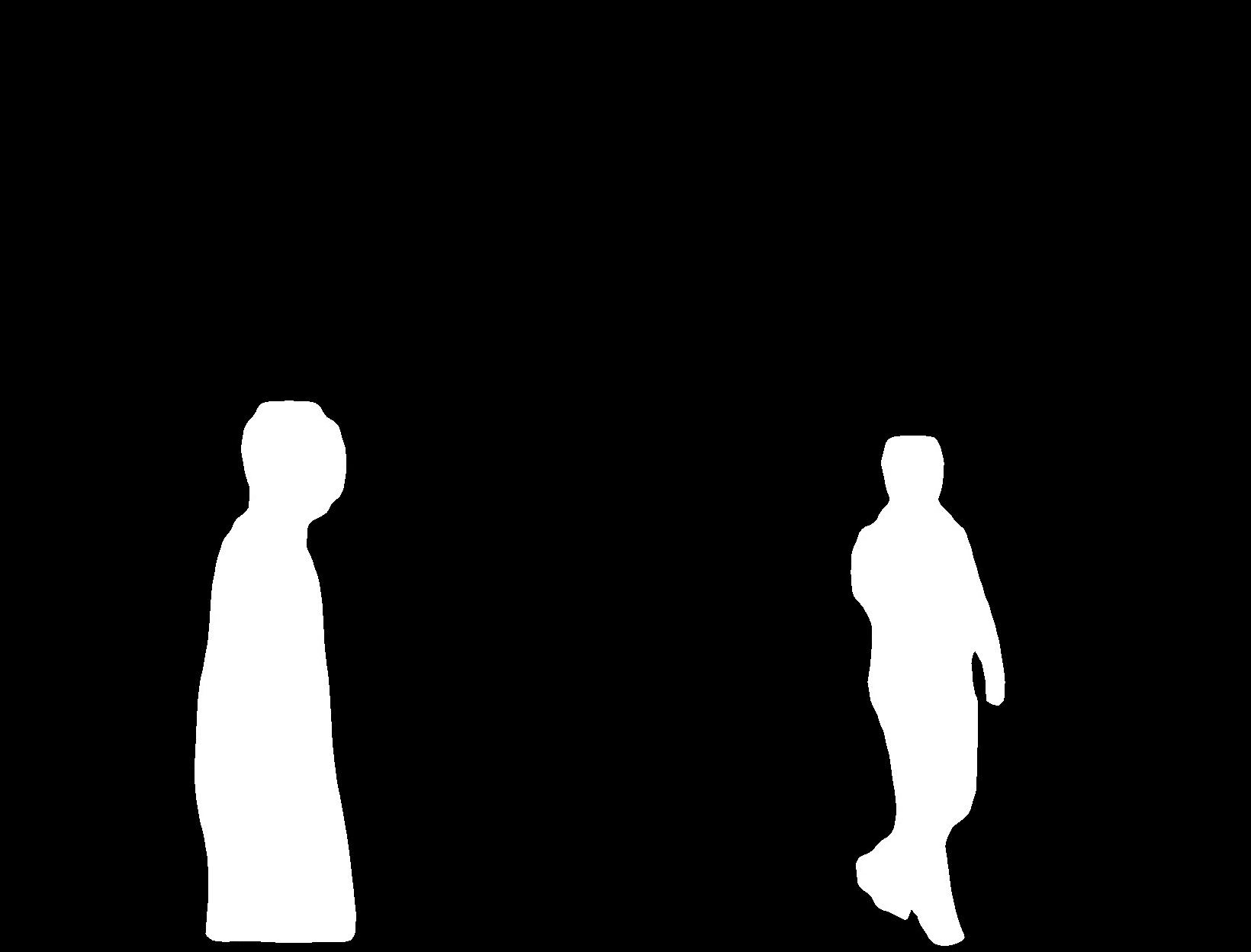}\\
            \vspace{0.03cm}
            \includegraphics[width=1\textwidth]{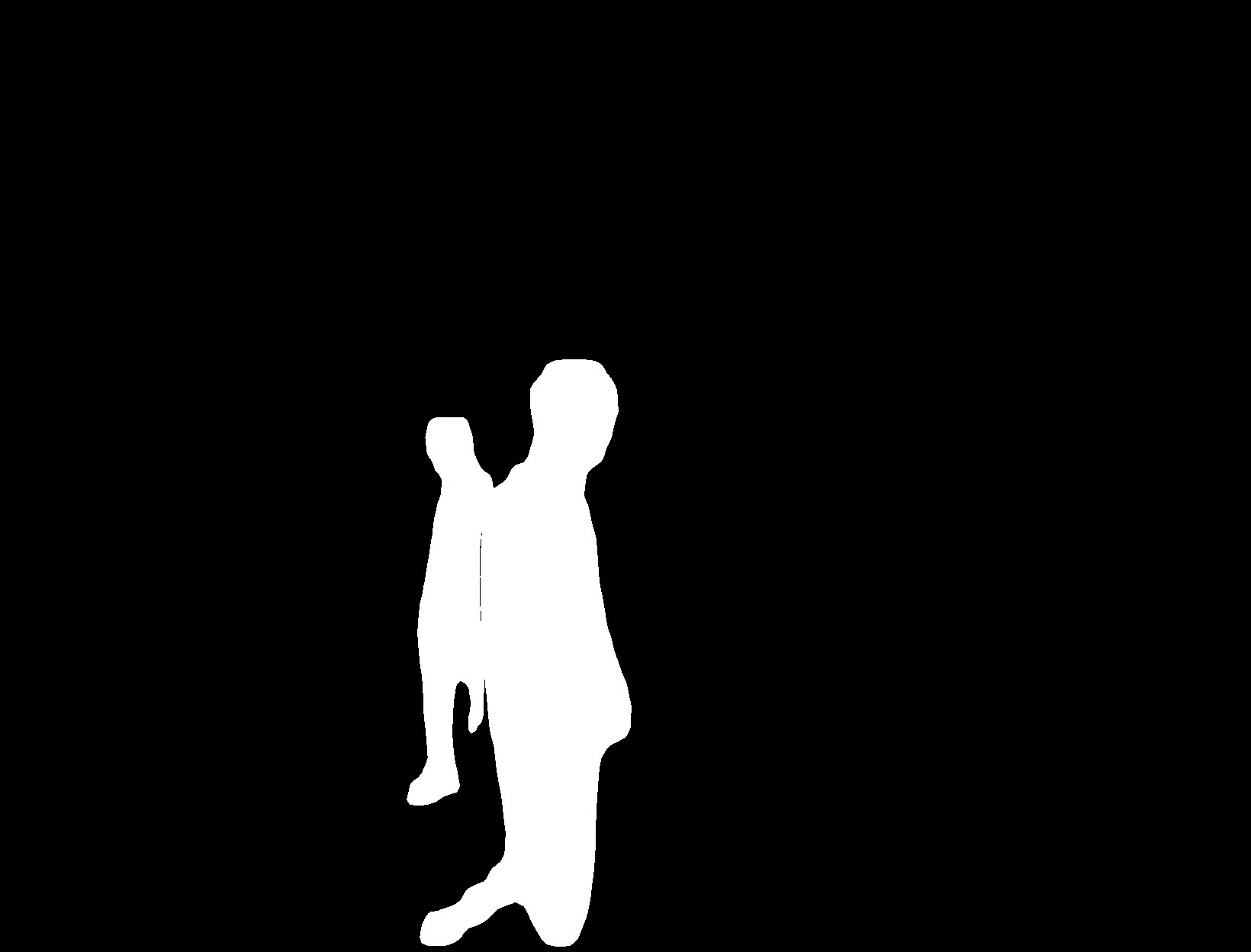}\\
            \vspace{0.03cm}
            \includegraphics[width=1\textwidth]{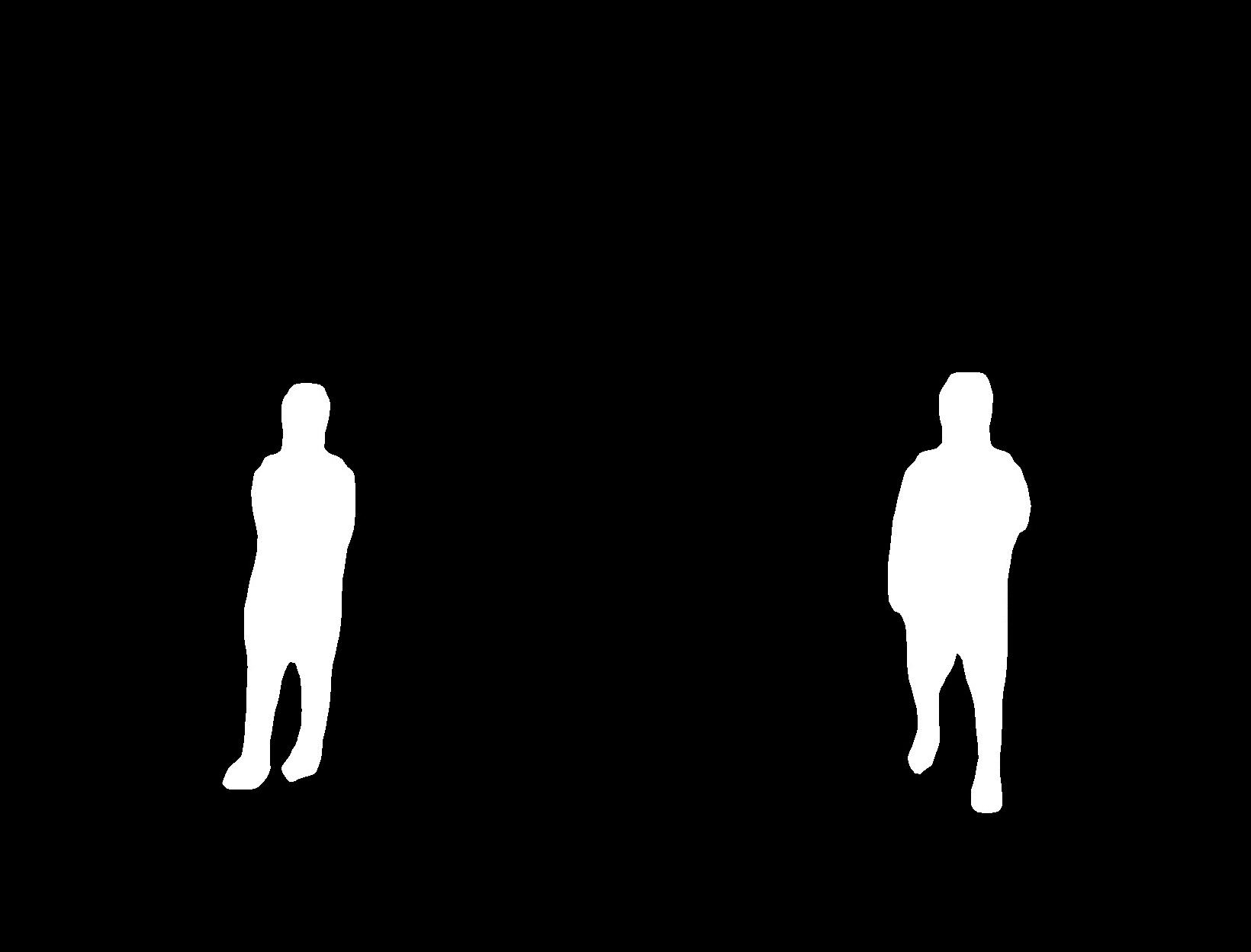}
            \label{multi_walk}
            \vspace{-0.3cm}
        \end{minipage}
    }
    \subfloat[]{
        \begin{minipage}{0.156\linewidth}
            \centering
            \includegraphics[width=1\textwidth]{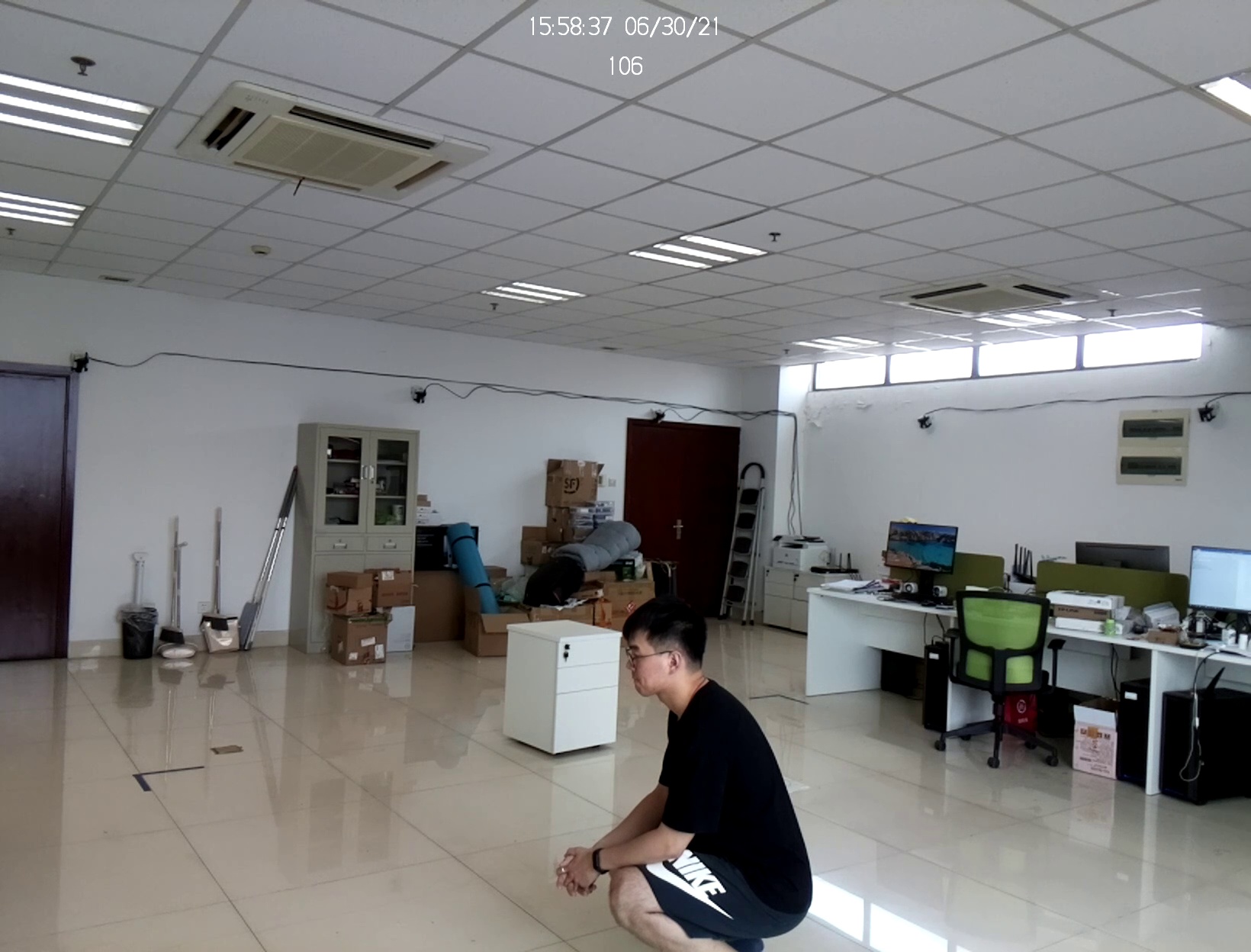}\\
            \vspace{0.03cm}
            \includegraphics[width=1\textwidth]{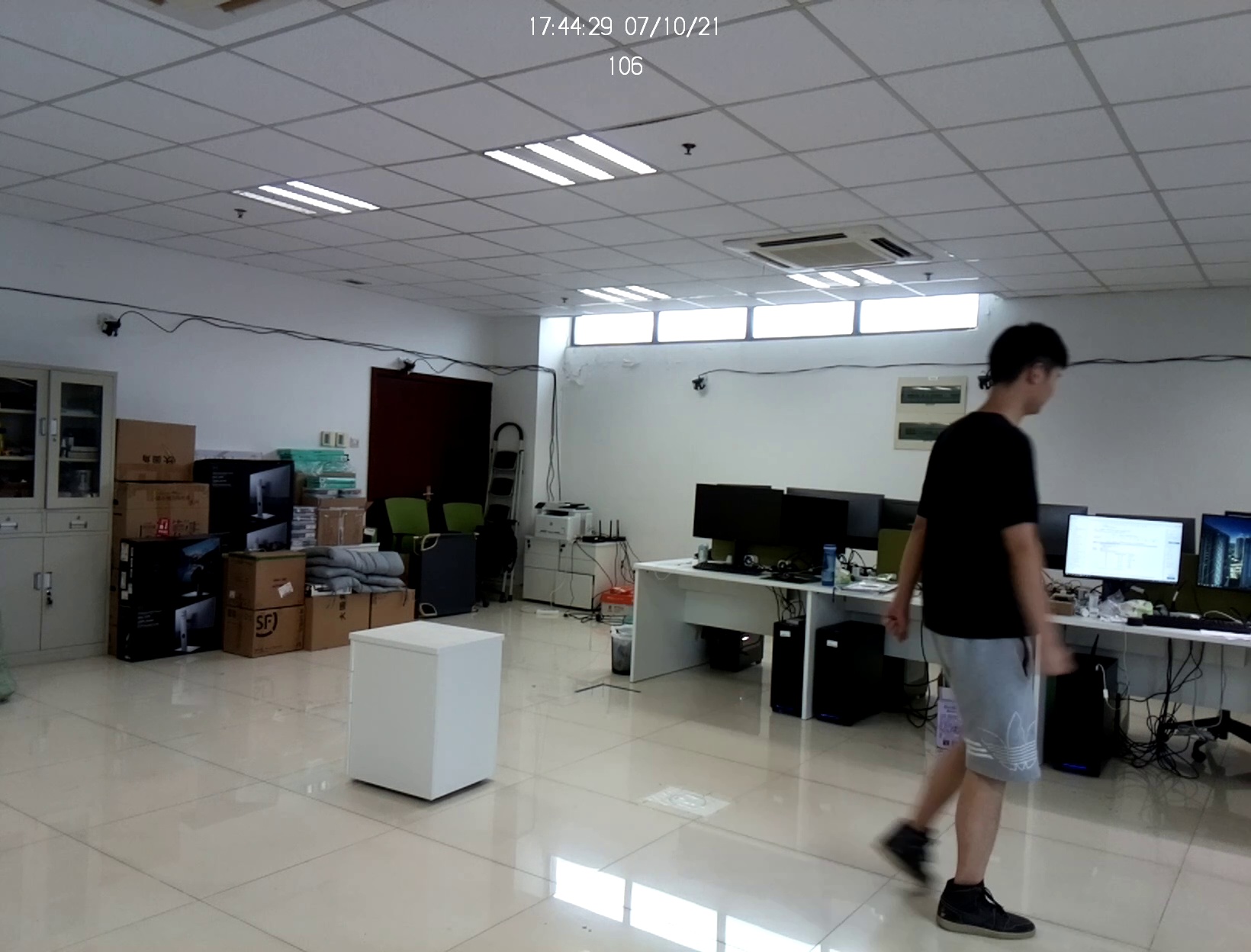}\\
            \vspace{0.03cm}
            \includegraphics[width=1\textwidth]{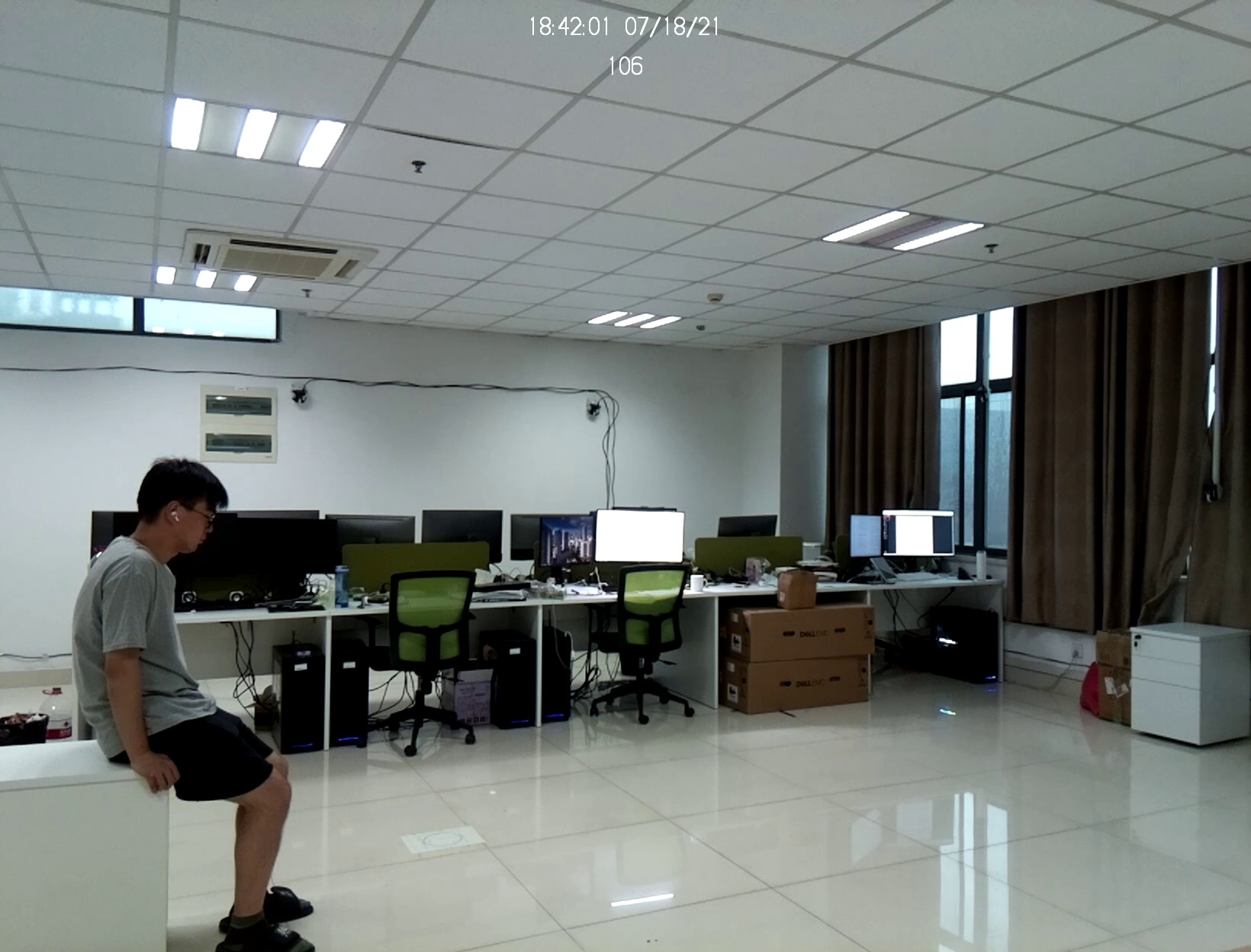}
            \label{action}
            \vspace{-0.3cm}
        \end{minipage}
        \begin{minipage}{0.156\linewidth}
            \centering
            \includegraphics[width=1\textwidth]{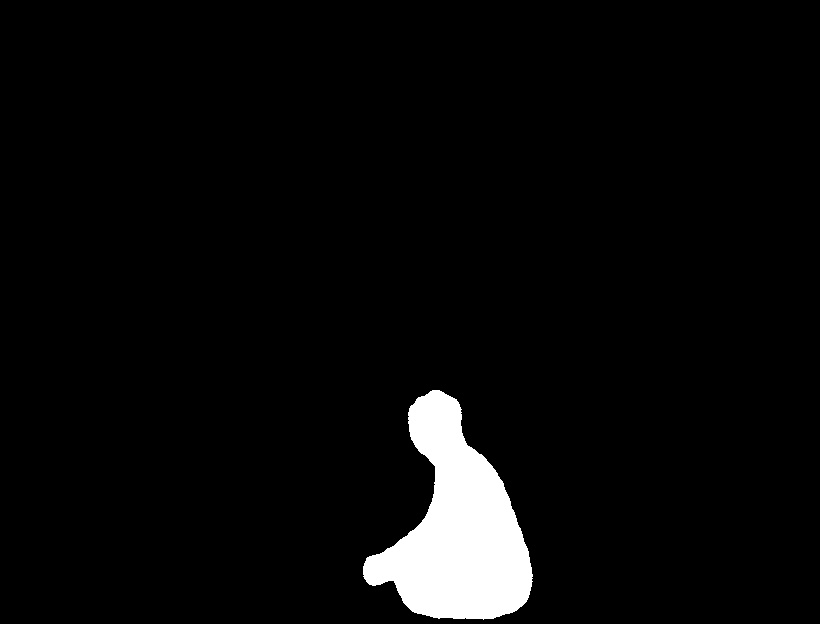}\\
            \vspace{0.03cm}
            \includegraphics[width=1\textwidth]{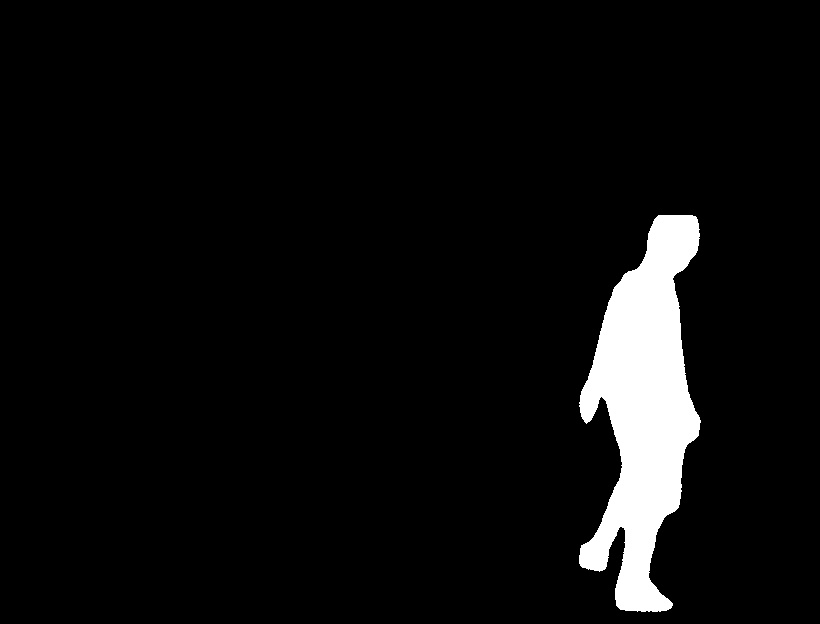}\\
            \vspace{0.03cm}
            \includegraphics[width=1\textwidth]{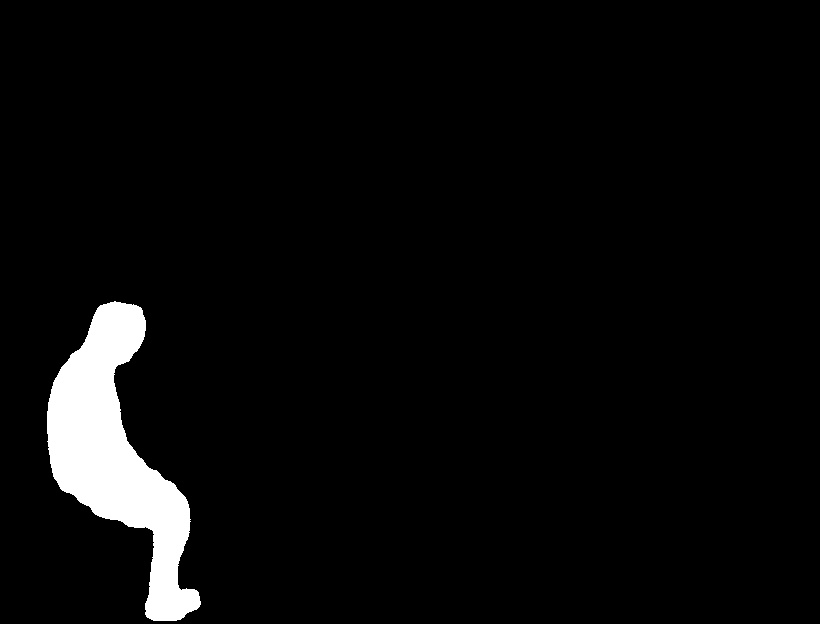}
            \label{action}
            \vspace{-0.3cm}
        \end{minipage}
    }
	
    \caption{Typical data frames in the dataset: (a) one person walking randomly in different views and their corresponding ground-truth segmentation results; (b) multiple persons walking randomly in different views and their corresponding ground-truth segmentation results; (c) one person performing actions in different views and their corresponding ground-truth segmentation results.}
    \label{dataset}
    \vspace{-3mm}
\end{figure*}

\begin{figure*}[ht]
    \centering
    
    \subfloat[]{
        \begin{minipage}[b]{0.11\linewidth}
            \includegraphics[width=1\textwidth]{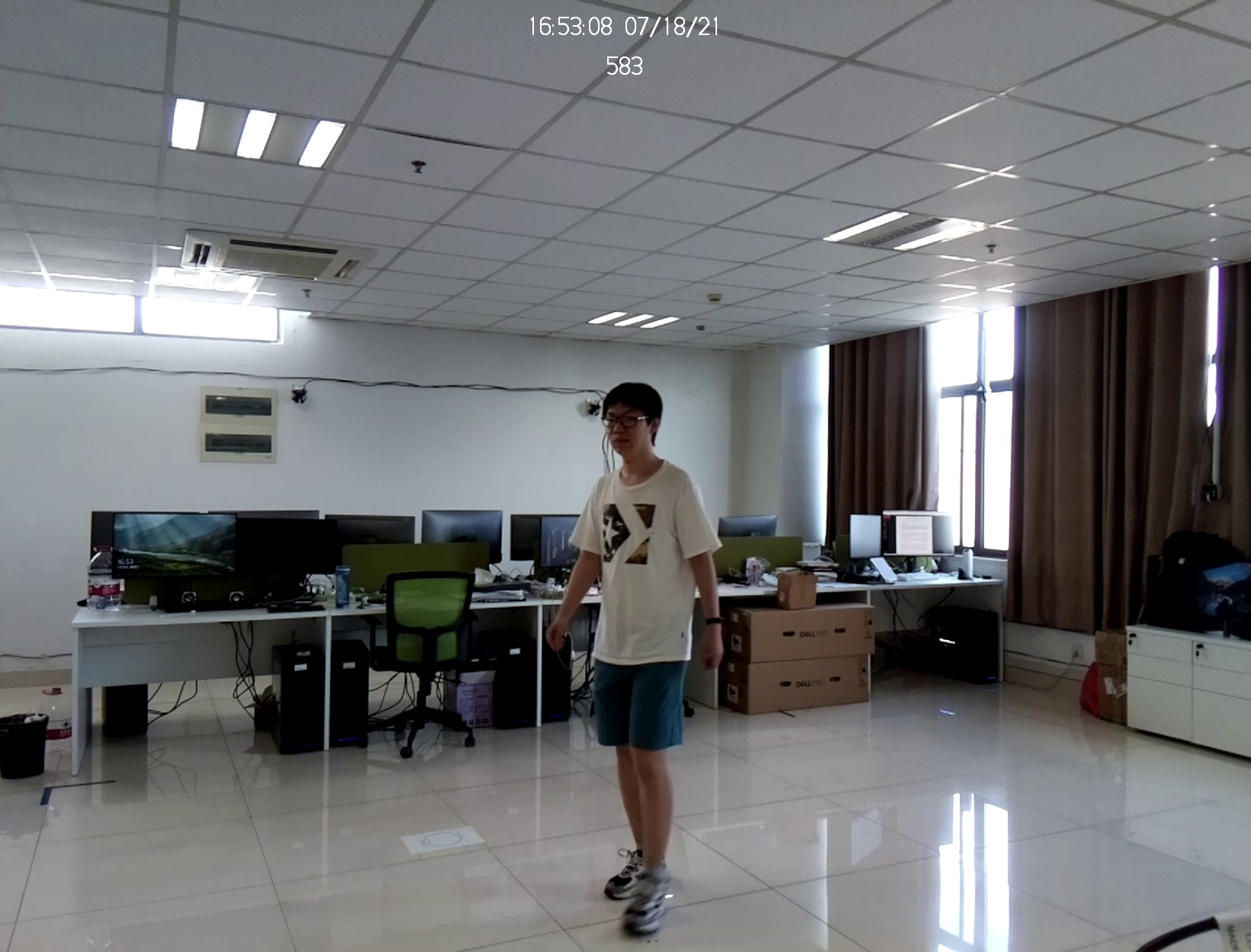} \\
            \vspace{-0.4cm}
            \includegraphics[width=1\textwidth]{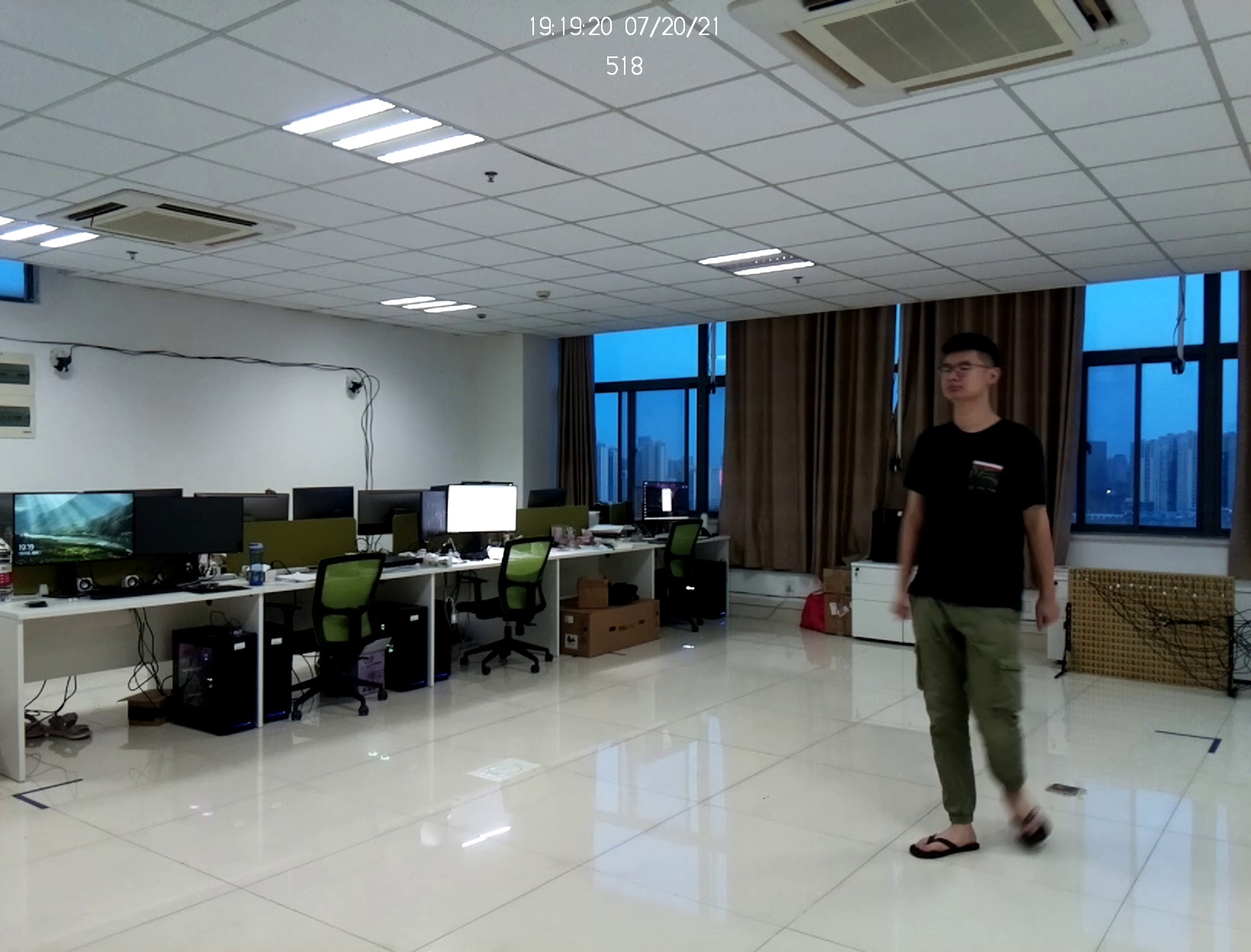} \\
            \vspace{-0.4cm}
            \includegraphics[width=1\textwidth]{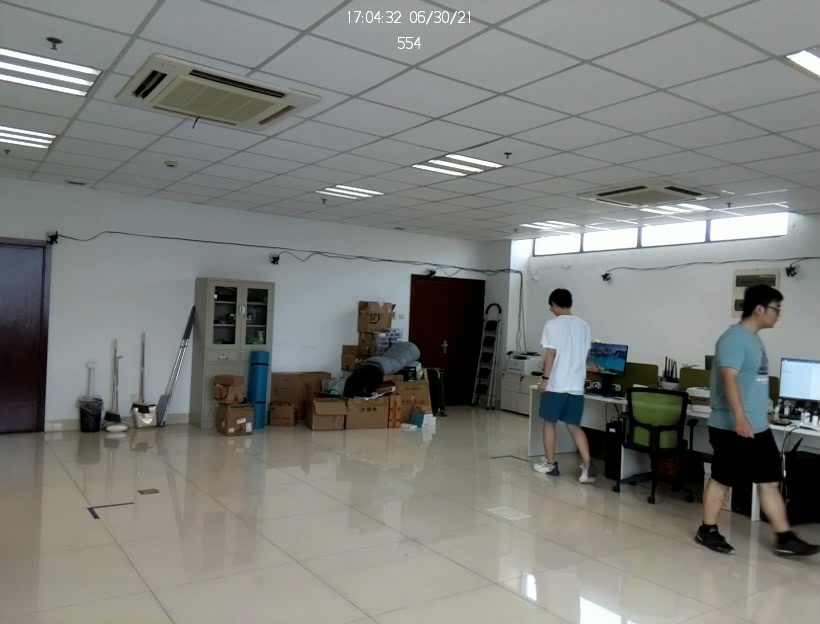} \\
            \vspace{-0.4cm}
            \includegraphics[width=1\textwidth]{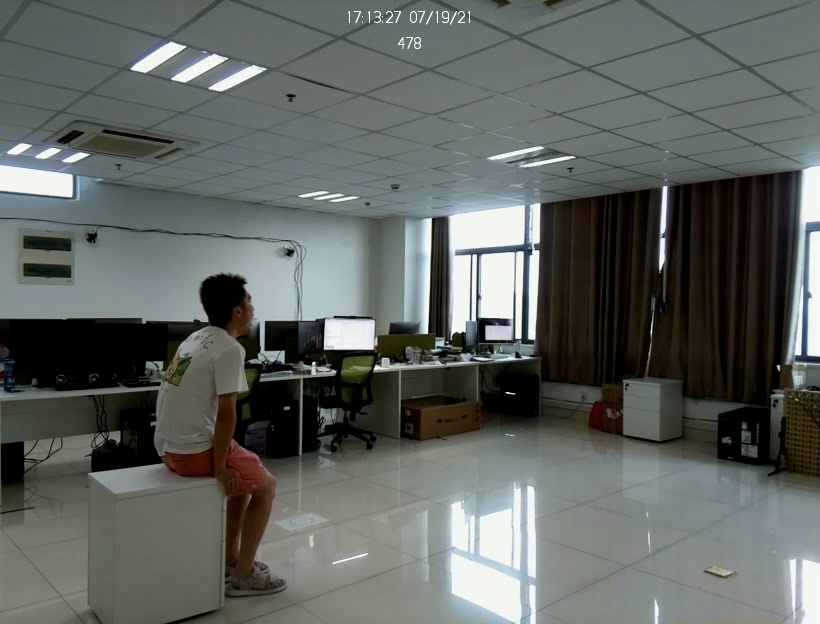} \\
            \vspace{-0.3cm}
        \end{minipage}
    }
    \subfloat[]{
        \begin{minipage}[b]{0.11\linewidth}
            \includegraphics[width=1\textwidth]{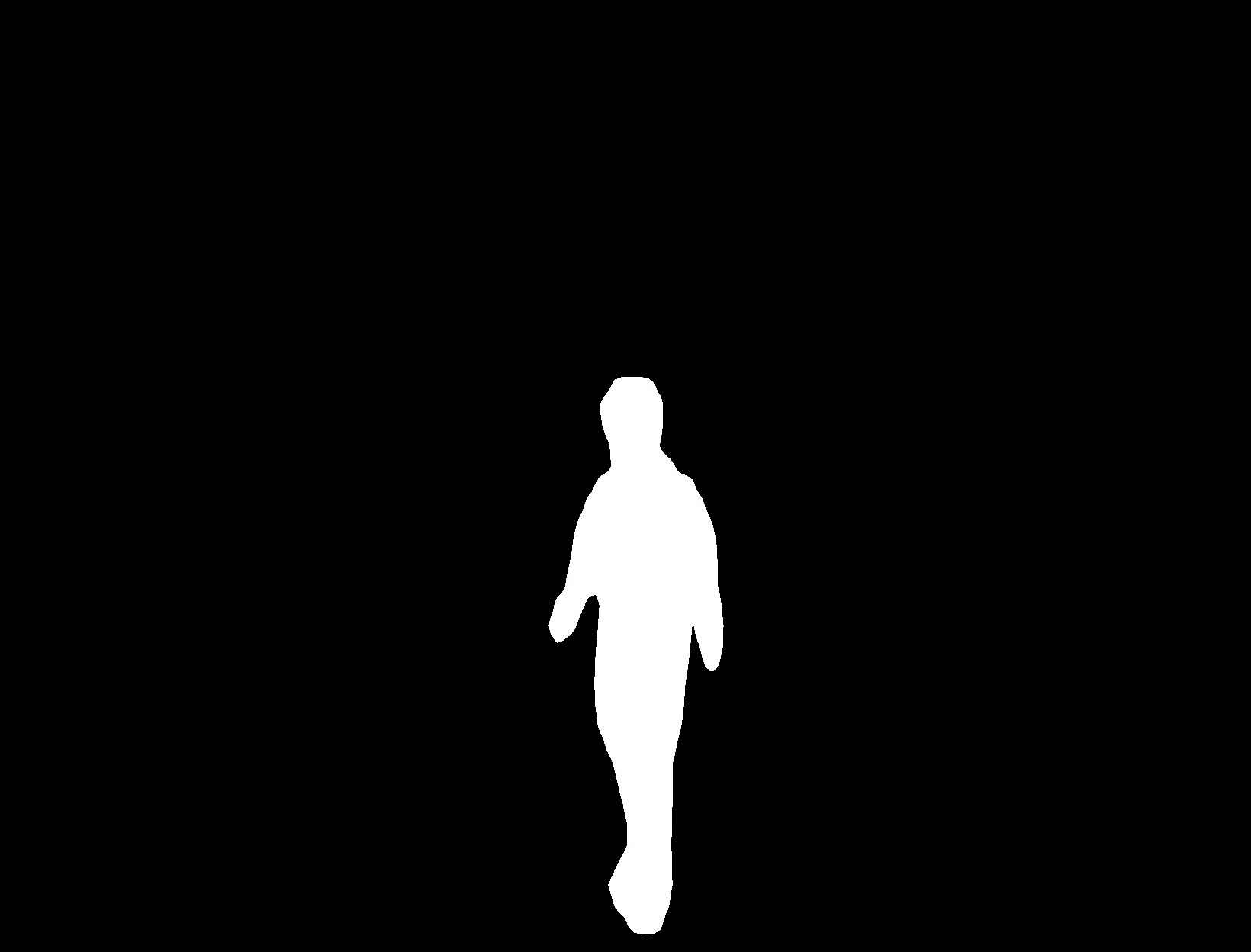} \\
            \vspace{-0.4cm}
            \includegraphics[width=1\textwidth]{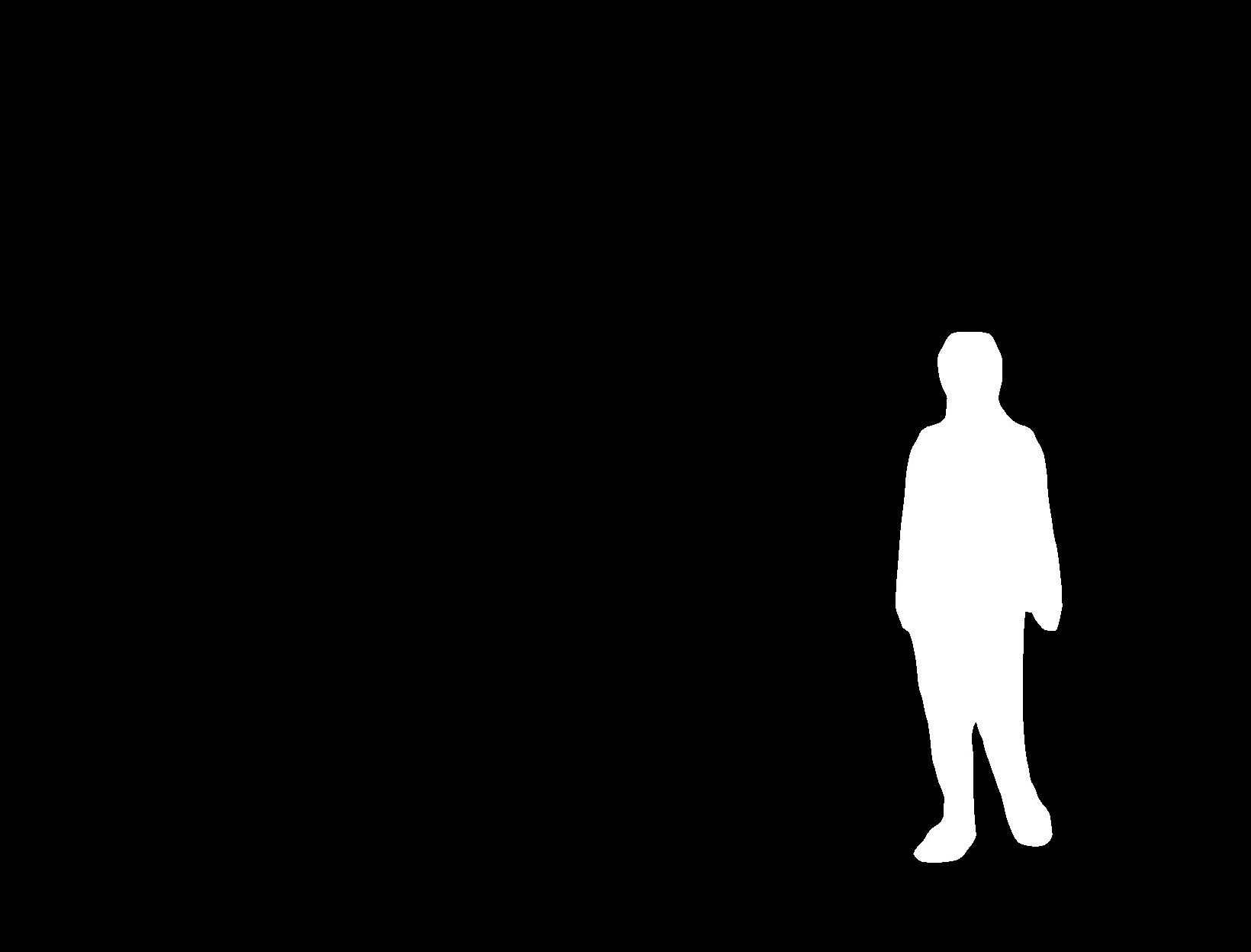} \\
            \vspace{-0.4cm}
            \includegraphics[width=1\textwidth]{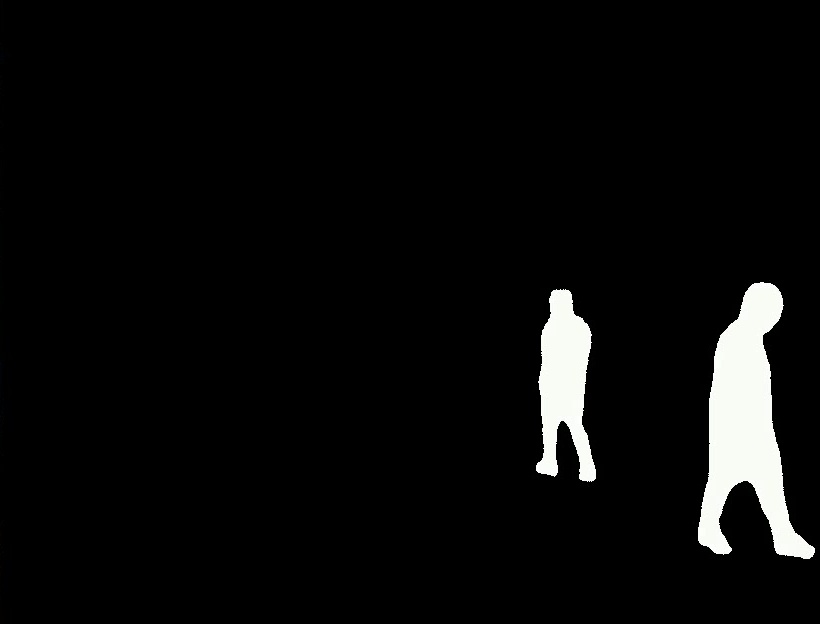} \\
            \vspace{-0.4cm}
            \includegraphics[width=1\textwidth]{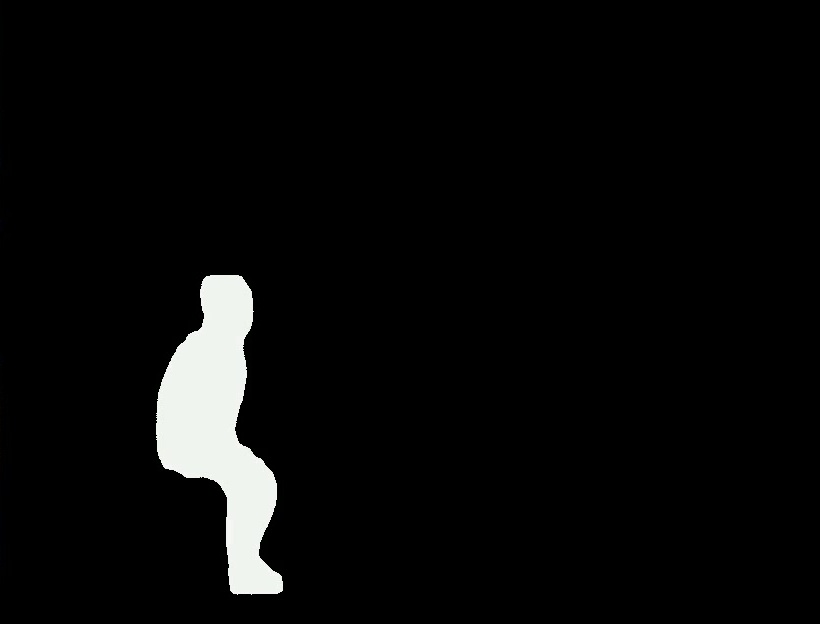} \\
            \vspace{-0.3cm}
        \end{minipage}
    }
    \subfloat[]{
        \begin{minipage}[b]{0.11\linewidth}
            \includegraphics[width=1\textwidth]{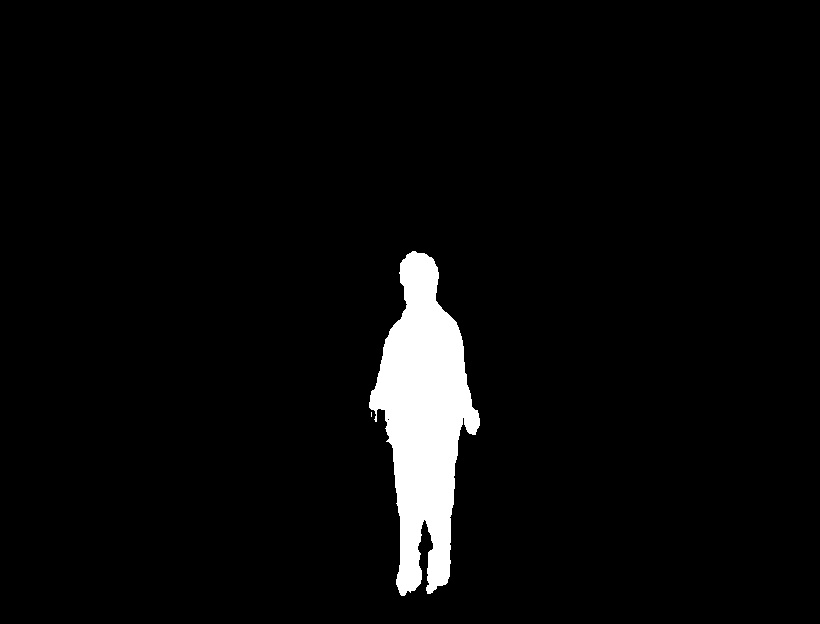} \\
            \vspace{-0.4cm}
            \includegraphics[width=1\textwidth]{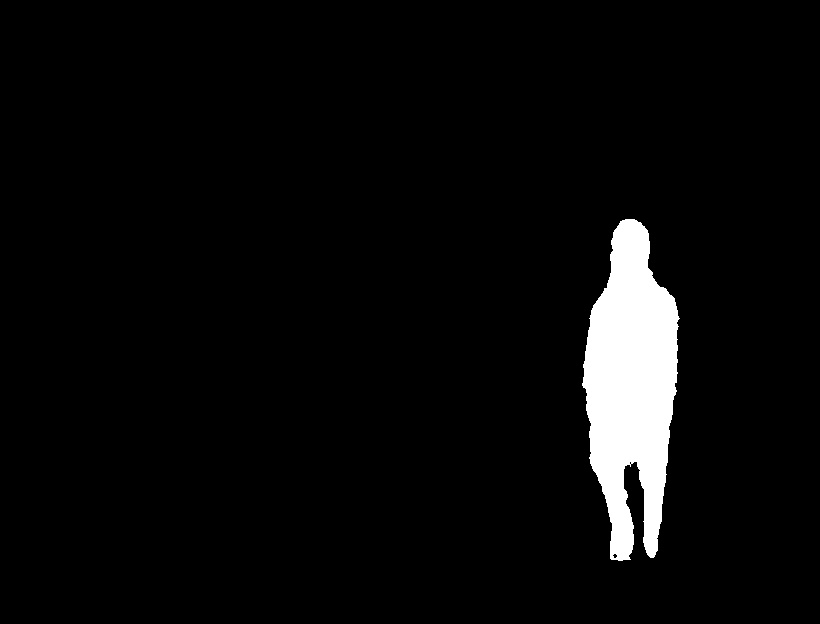} \\
            \vspace{-0.4cm}
            \includegraphics[width=1\textwidth]{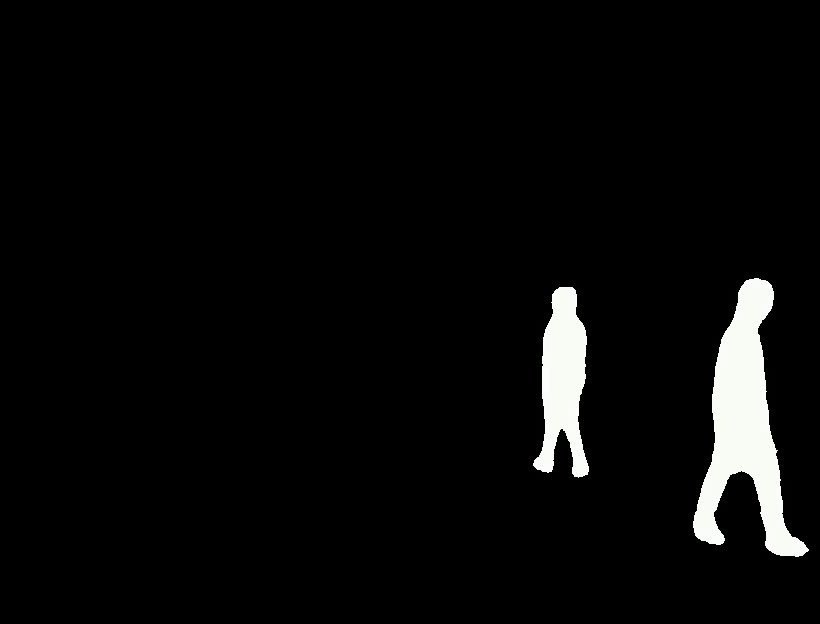} \\
            \vspace{-0.4cm}
            \includegraphics[width=1\textwidth]{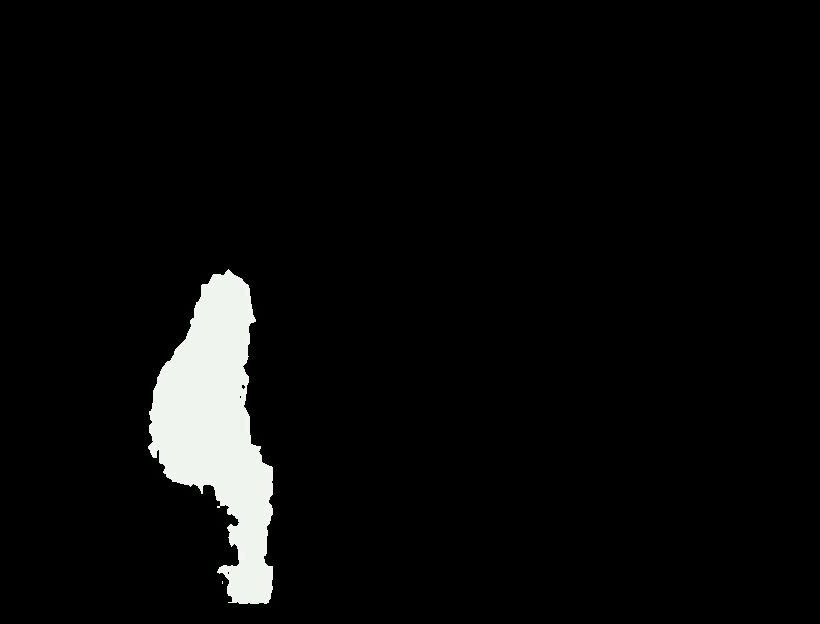} \\
            \vspace{-0.3cm}
        \end{minipage}
    }
    \subfloat[]{
        \begin{minipage}[b]{0.11\linewidth}
            \includegraphics[width=1\textwidth]{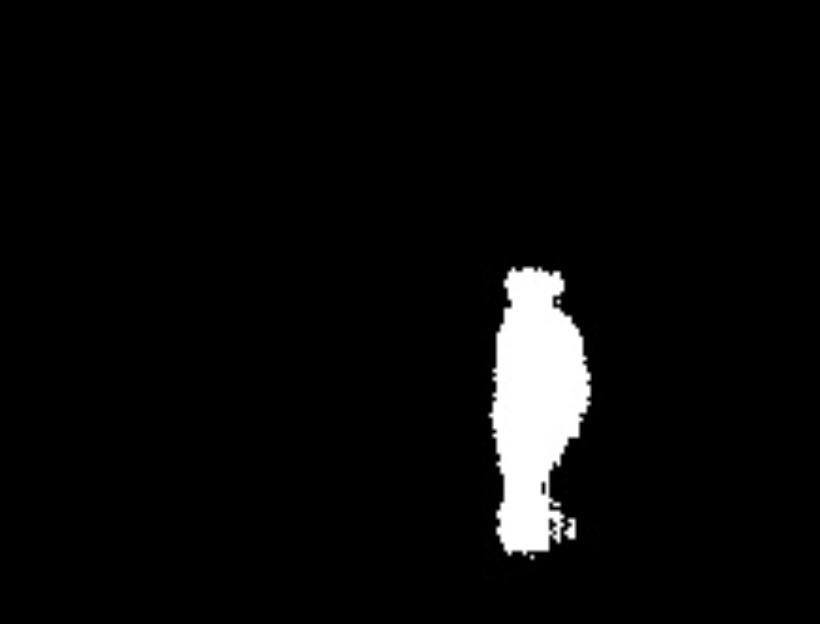} \\
            \vspace{-0.4cm}
            \includegraphics[width=1\textwidth]{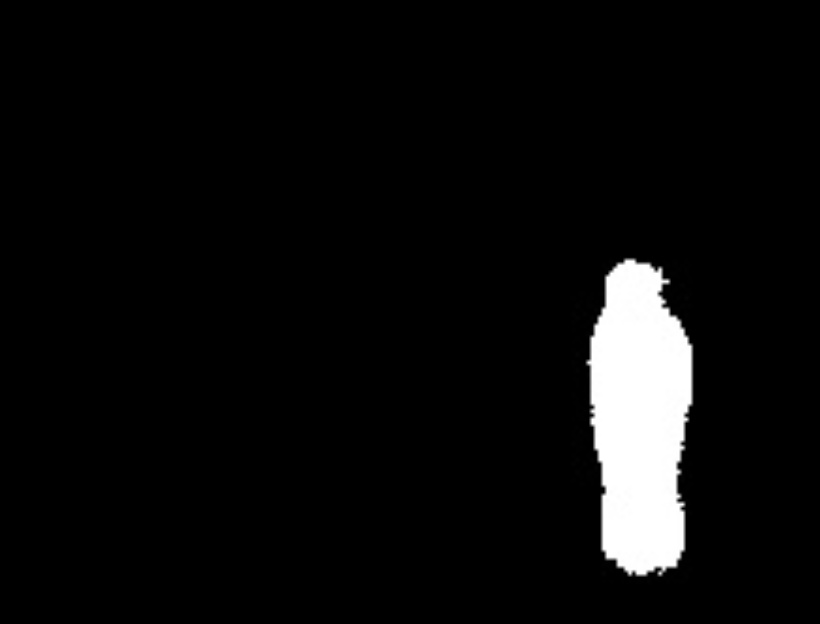} \\
            \vspace{-0.4cm}
            \includegraphics[width=1\textwidth]{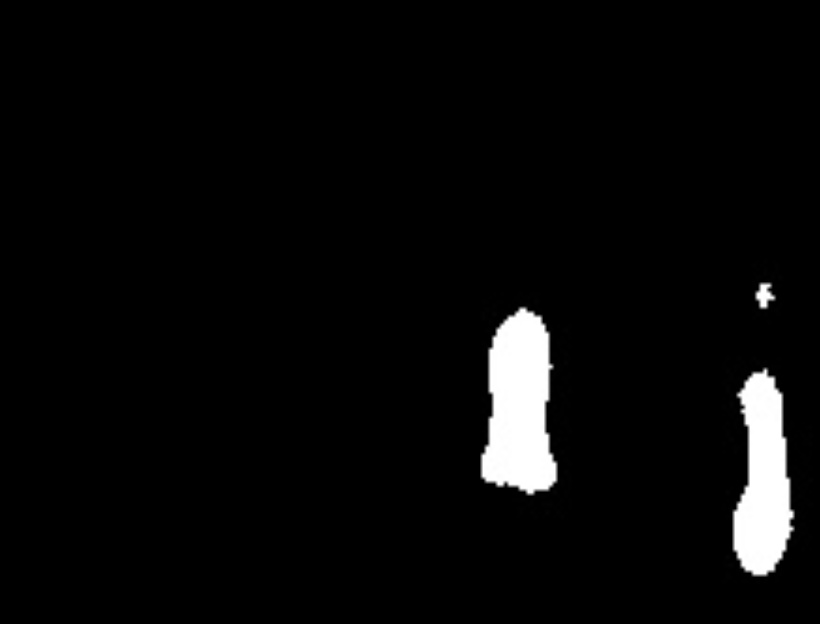} \\
            \vspace{-0.4cm}
            \includegraphics[width=1\textwidth]{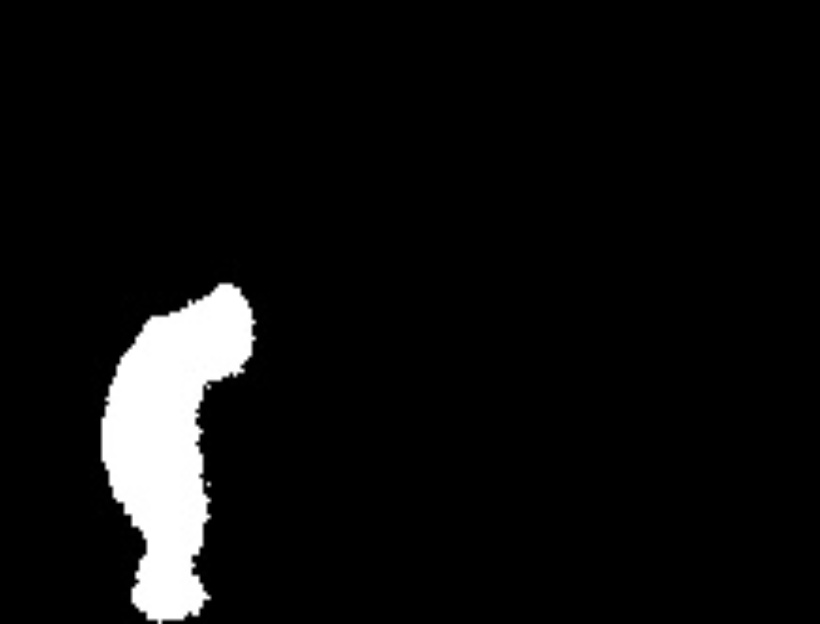} \\
            \vspace{-0.3cm}
        \end{minipage}
    }
    \subfloat[]{
        \begin{minipage}[b]{0.11\linewidth}
            \includegraphics[width=1\textwidth]{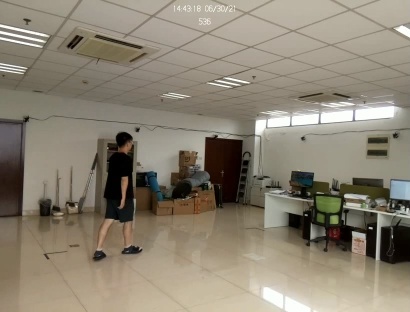} \\
            \vspace{-0.4cm}
            \includegraphics[width=1\textwidth]{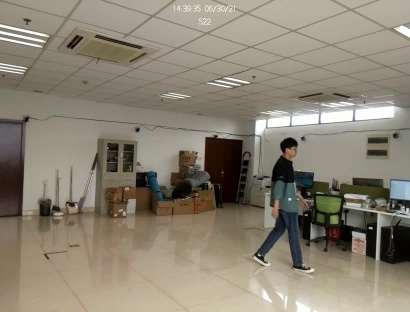} \\
            \vspace{-0.4cm}
            \includegraphics[width=1\textwidth]{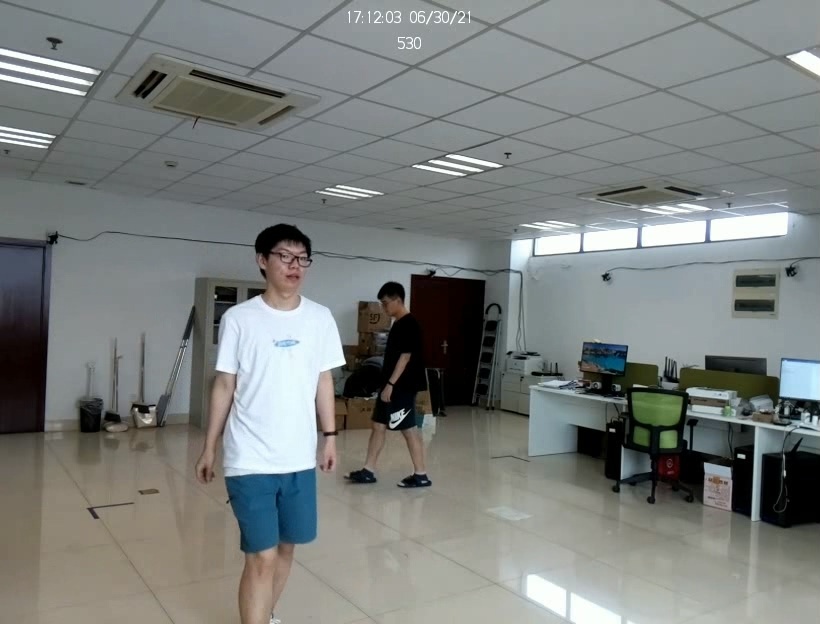} \\
            \vspace{-0.4cm}
            \includegraphics[width=1\textwidth]{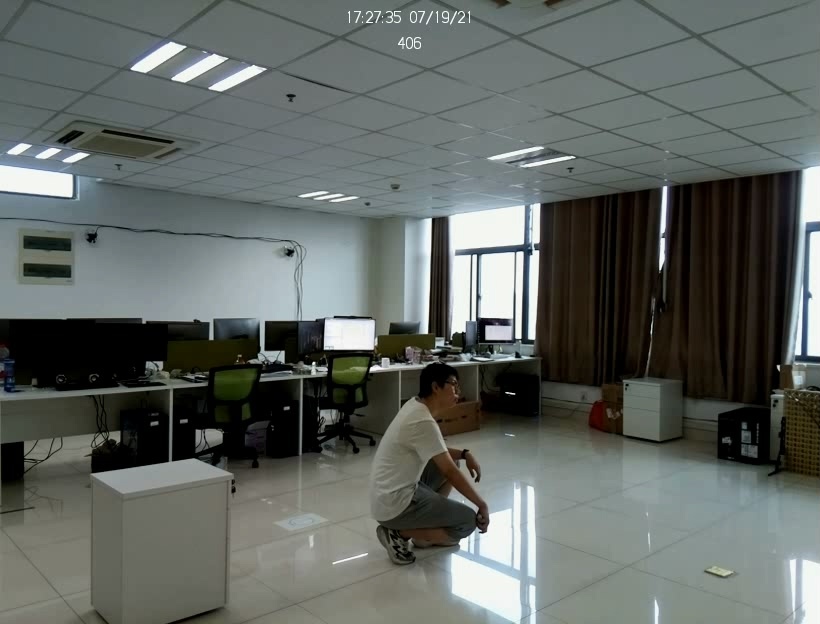} \\
            \vspace{-0.3cm}
        \end{minipage}
    }
    \subfloat[]{
        \begin{minipage}[b]{0.11\linewidth}
            \includegraphics[width=1\textwidth]{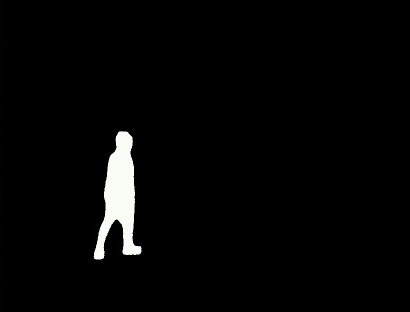} \\
            \vspace{-0.4cm}
            \includegraphics[width=1\textwidth]{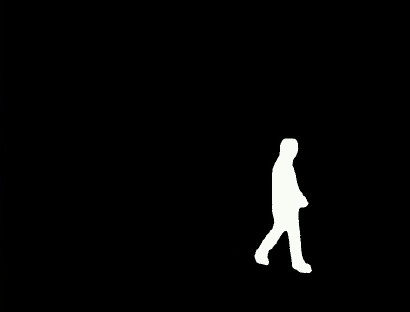} \\
            \vspace{-0.4cm}
            \includegraphics[width=1\textwidth]{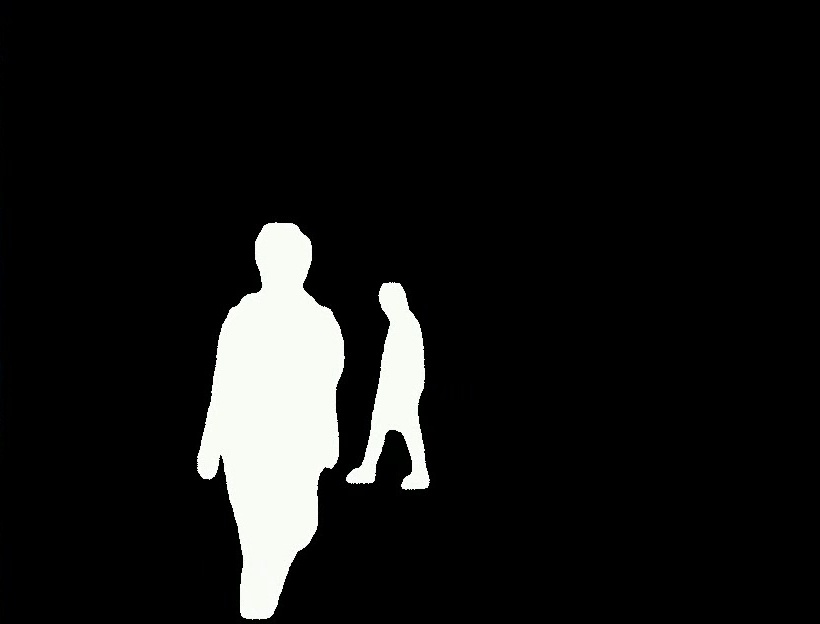} \\
            \vspace{-0.4cm}
            \includegraphics[width=1\textwidth]{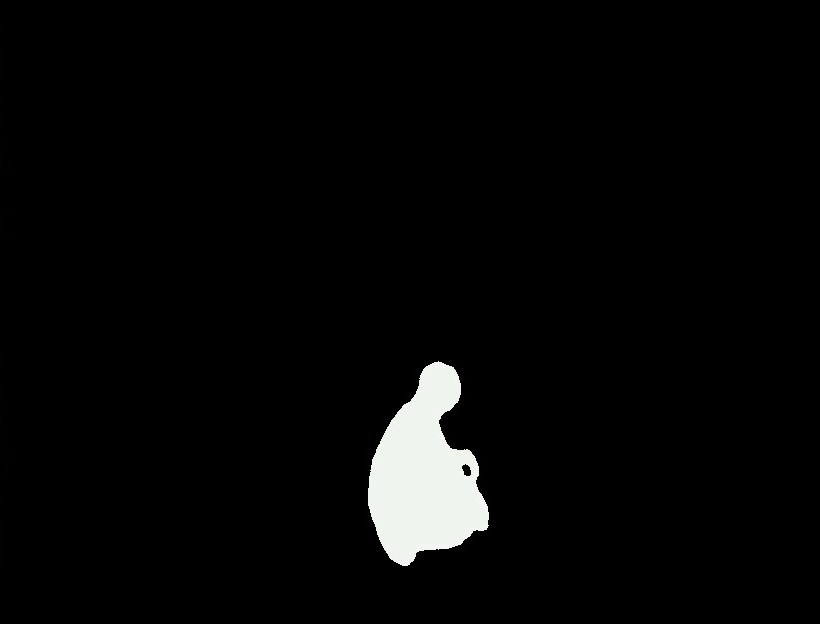} \\
            \vspace{-0.3cm}
        \end{minipage}
    }
    \subfloat[]{
        \begin{minipage}[b]{0.11\linewidth}
            \includegraphics[width=1\textwidth]{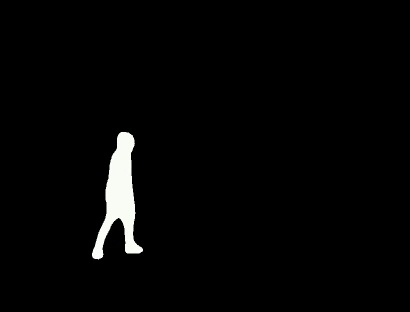} \\
            \vspace{-0.4cm}
            \includegraphics[width=1\textwidth]{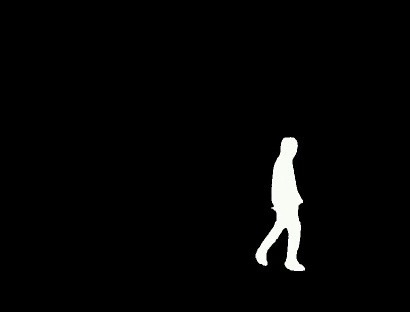} \\
            \vspace{-0.4cm}
            \includegraphics[width=1\textwidth]{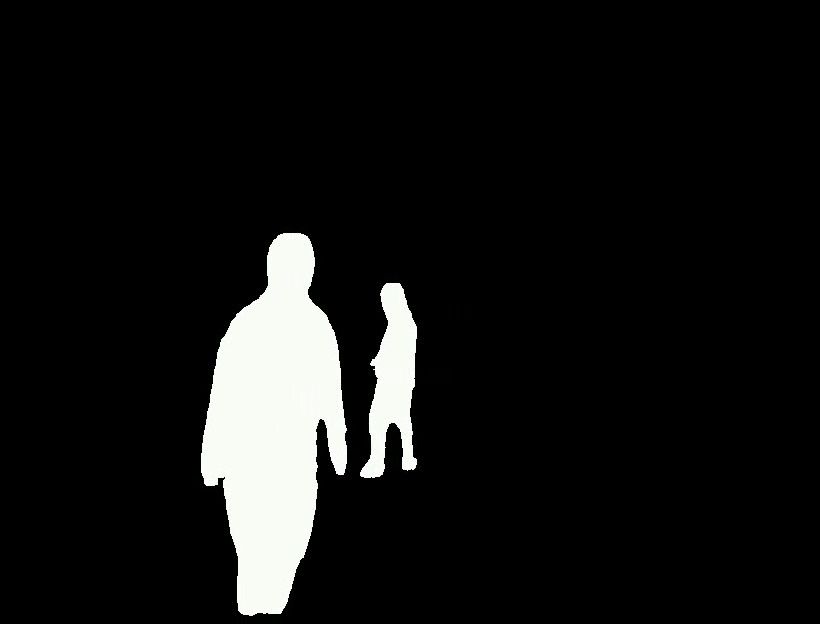} \\
            \vspace{-0.4cm}
            \includegraphics[width=1\textwidth]{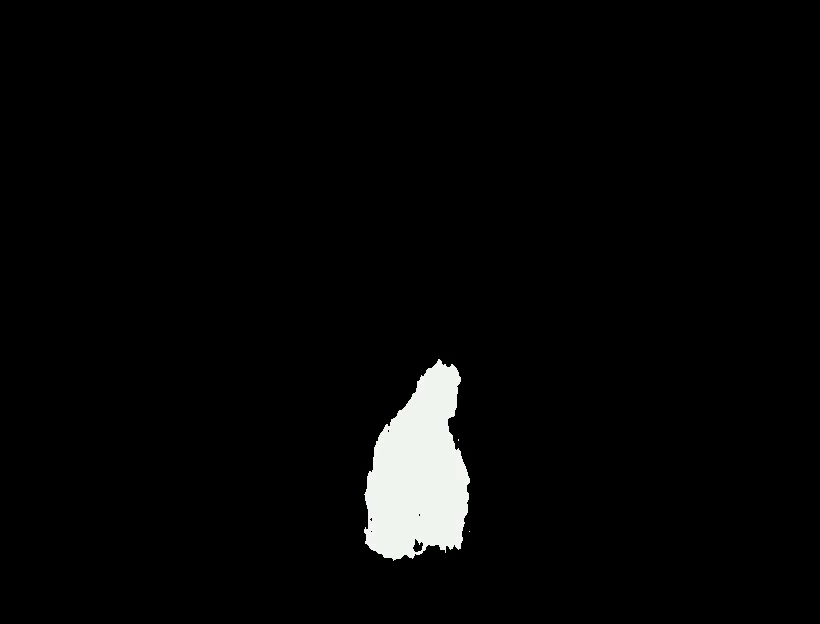} \\
            \vspace{-0.3cm}
        \end{minipage}
    }
    \subfloat[]{
        \begin{minipage}[b]{0.11\linewidth}
            \includegraphics[width=1\textwidth]{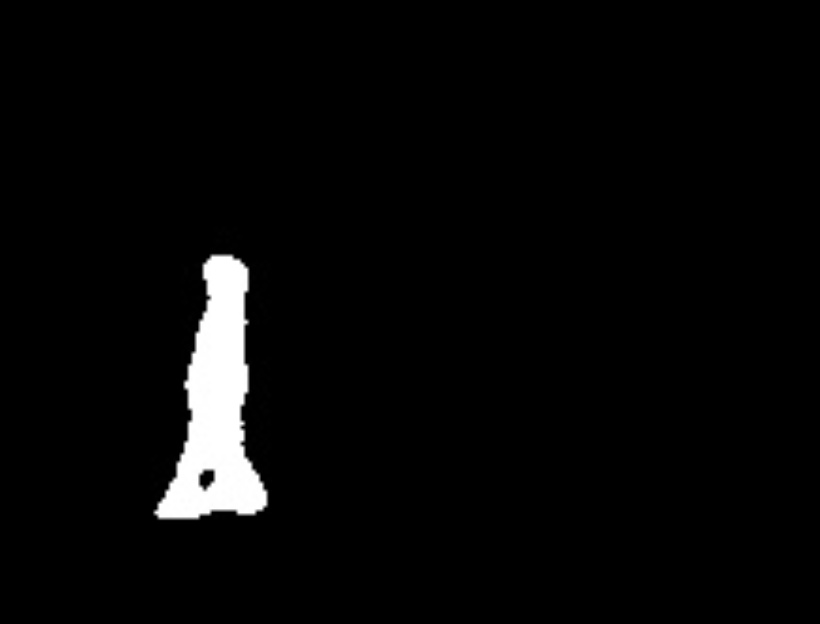} \\
            \vspace{-0.4cm}
            \includegraphics[width=1\textwidth]{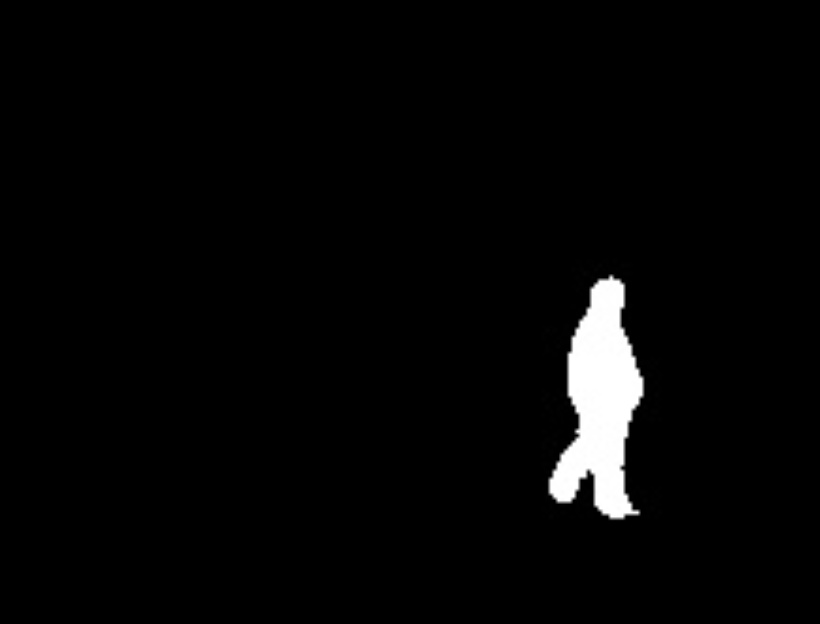} \\
            \vspace{-0.4cm}
            \includegraphics[width=1\textwidth]{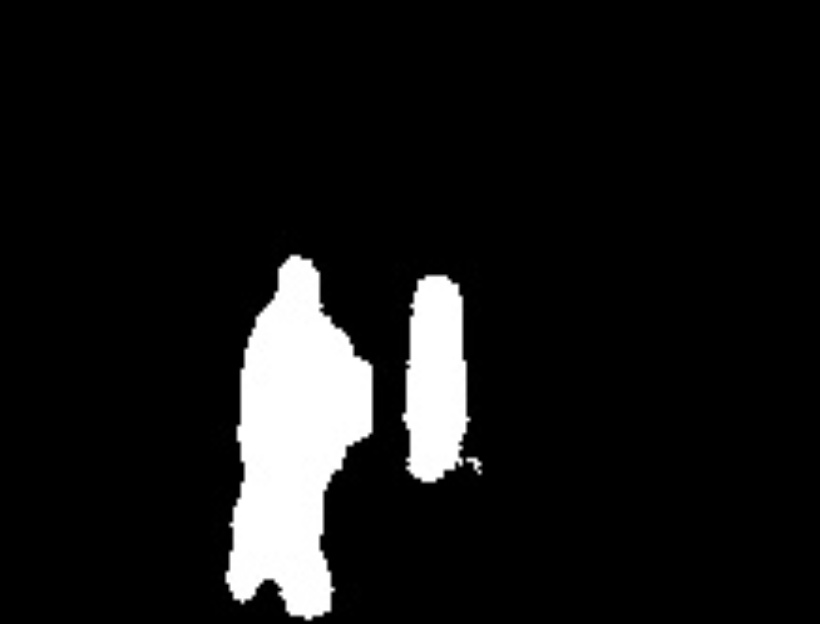} \\
            \vspace{-0.4cm}
            \includegraphics[width=1\textwidth]{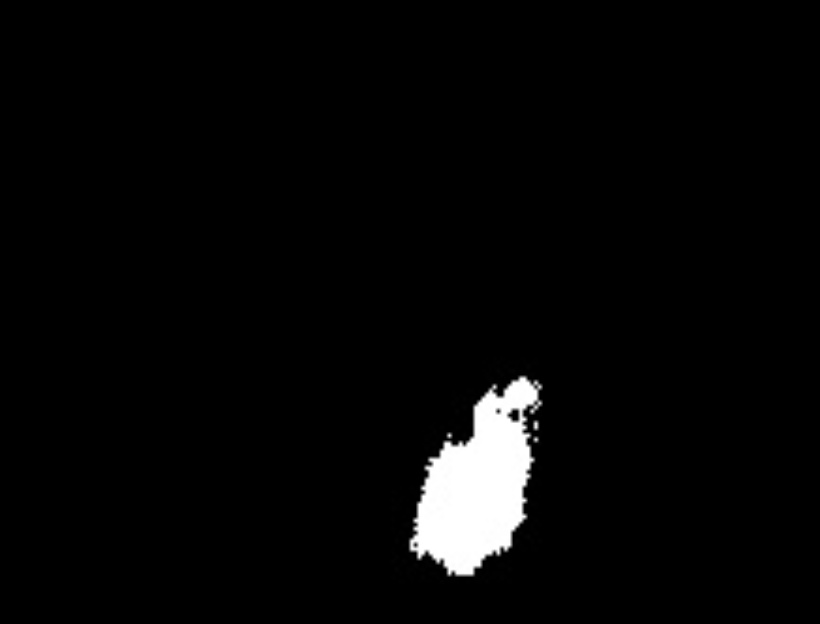} \\
            \vspace{-0.3cm}
        \end{minipage}
    }
	
    \caption{Comparison of results under single-person, multi-person, and action scenarios: (a) the camera view; (b) the ground-truth; (c) the results of RFMask; (d) the results of RFPose(12); (e) the camera view; (f) the ground-truth; (g) the results of RFMask; (h) the results of RFPose(12). RFMask output more fine-grained silhouette results compared with RFPose in single-person, multi-person, and action scenarios.}
    \label{result}
\end{figure*}

\section{RFMask}
\label{rfmask}
RFMask is a framework which can generate human silhouette with RF signals. 
As shown in Figure \ref{framework}, RFMask is composed of three components: signal processing, human detection, and mask generation. 
It takes RF signals as input, processes RF signals into frame sequences, locates 
human positions, and then encodes surrounding signals into intermediate feature representation, and finally generates human silhouettes.
In the following, we will introduce these three components in detail, respectively. 
\subsection{Signal Processing}
Our RF-based system relies on transmitting RF signals and receiving the reflections. 
To separate signals from different spatial locations, we adopt Frequency Modulated Continuous Wave (FMCW) and linear antenna arrays for signal transceiving. By combing FMCW and antenna array, the RF signals from the location $(x, y, z)$ can be extracted as
\begin{equation}
y(x,y,z,t)= \sum_{k=1}^{K}\sum_{m=1}^{M} s_{k,m,t} \cdot e^{j2\pi \frac{d_m(x,y,z)}{\lambda_k}}
\end{equation}
where $s_{k,m,t}$ denotes the $k$-th sample of FMCW sweep on $m$-th antenna at time $t$, $\lambda_k$ is the signal wavelength of $k$-th sample, $d_m(x,y,z)$ denotes the round-trip distance from the transmitting antenna to location $(x, y, z)$ and back to the receiving antenna. As illustrated in \cite{rfpose}, the time and space complexity of processing raw 4D tensor is extremely high, which is not supported by major machine learning platforms. 
Hence, inspired by \cite{rfpose}, we decompose the signal into horizontal and vertical planes and deal with them in parallel. 
In this case, the input data for neural network takes the form of two-dimensional maps, which can be expressed as 
\begin{equation}
    y_{hor}(x,y,t)= \sum_{k=1}^{K}\sum_{m=1}^{M} s_{k,m,t} \cdot e^{j2\pi \frac{d_m(x,y)}{\lambda_k}} ,
\end{equation}
\begin{equation}
    y_{ver}(y,z,t)= \sum_{k=1}^{K}\sum_{m=1}^{M} s_{k,m,t} \cdot e^{j2\pi \frac{d_m(y,z)}{\lambda_k}} .
\end{equation}
Due to the severe multi-path interference in the indoor environment\cite{MTrack}, human reflection is invisible on the raw extracted signal. 
To address this problem, we have noted that the signal reflected from persons would change over time due to their movement, while the multi-path caused by static objects in the environment keeps the same. Hence, we subtract the signals in time domain to enhance the signals reflected by human. 
\begin{figure}[tbhp]
    \centering
    \includegraphics[width=0.5\linewidth]{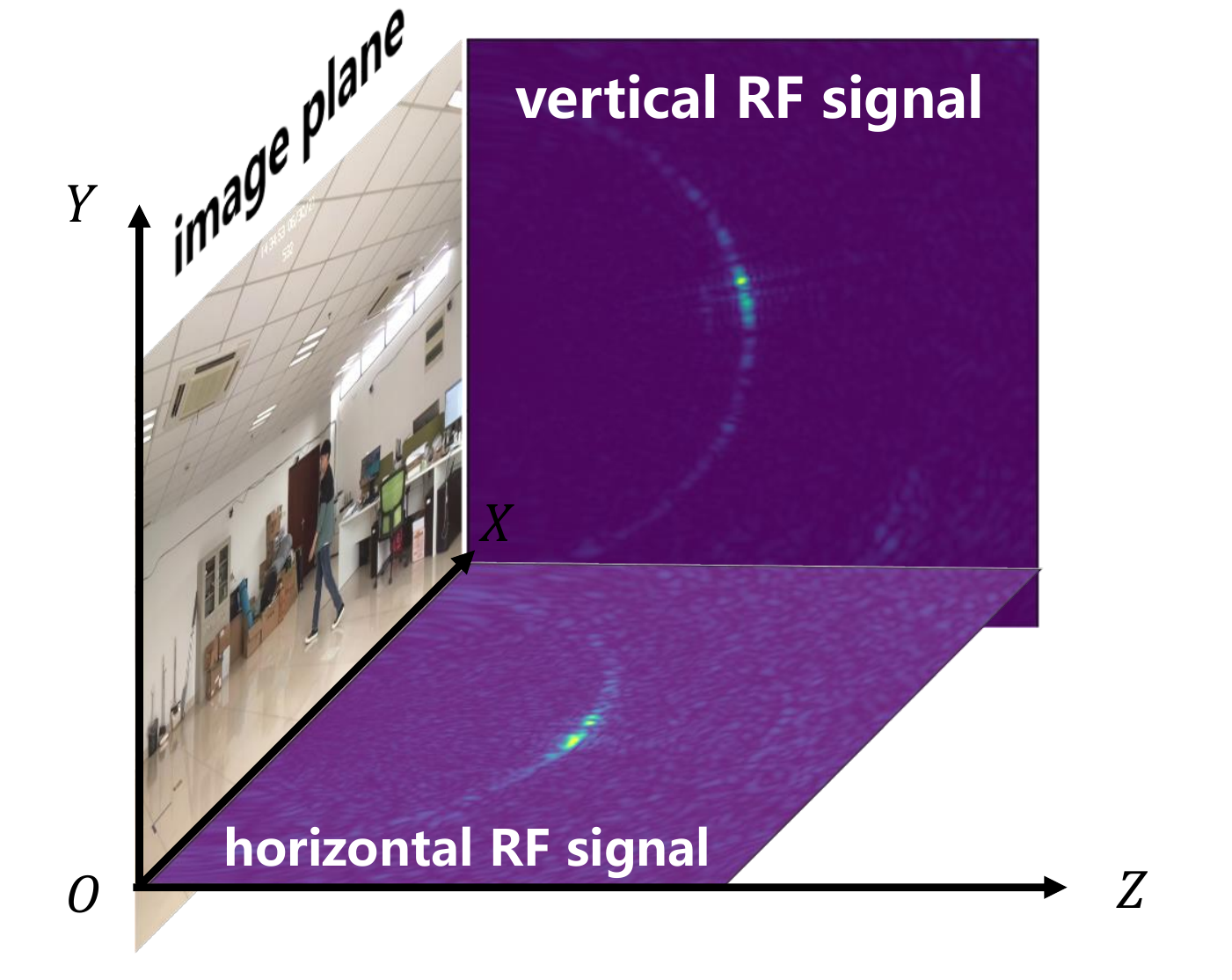}
    \caption{Illustration of the RF signal plane and image plane. The image plane is perpendicular to both the horizontal and vertical signal planes.}
	\label{perpendicular}
    \vspace{-3mm}
\end{figure}

\begin{figure}[tbhp]
    \centering
        \includegraphics[width=0.6\columnwidth]{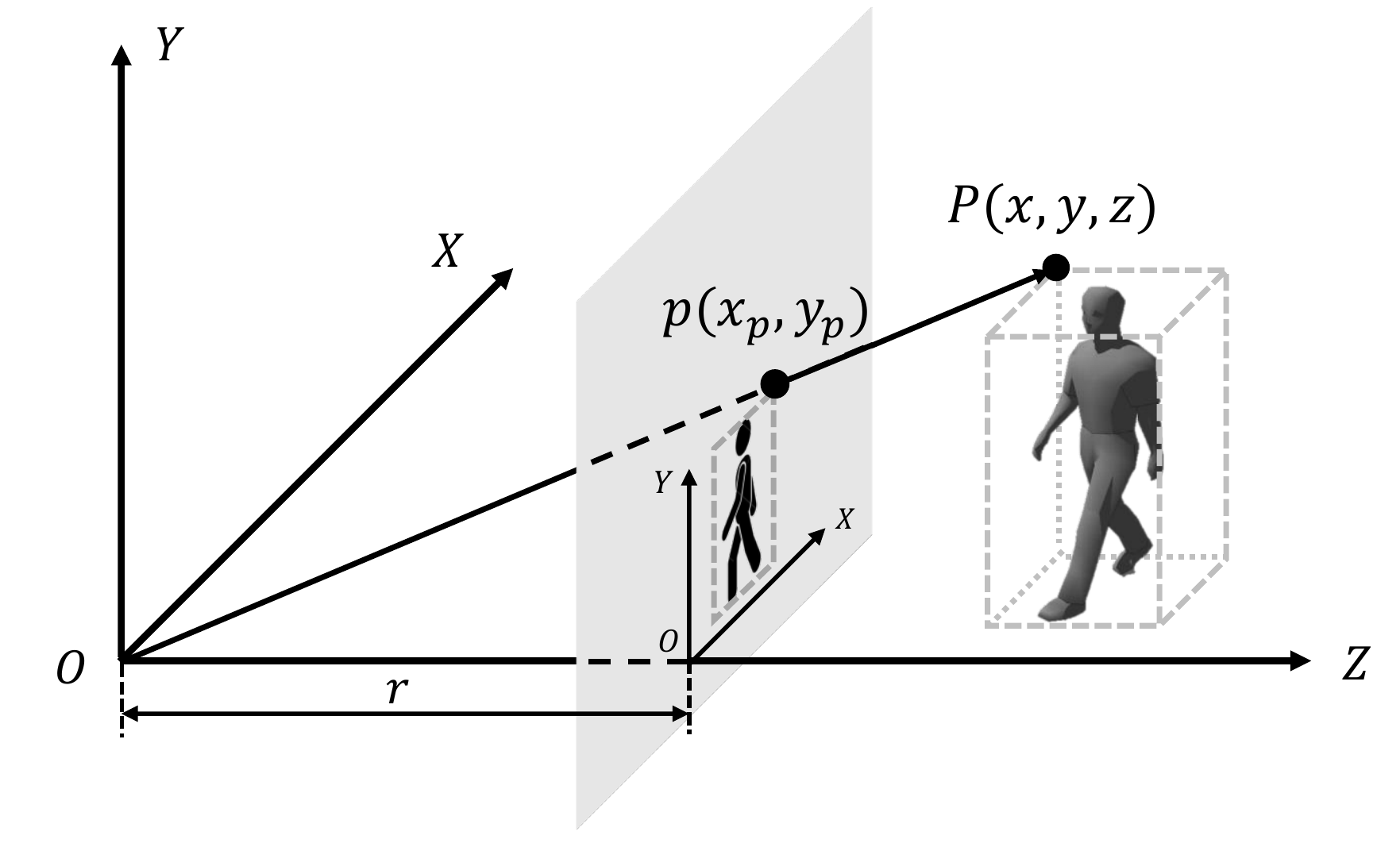}
    \caption{Projection from 3D coordinate system to result plane. A point $P(x,y,z)$ in 3D coordinate can be projected into result plane $\bm{Z}=r$ as $p(x_p,y_p)$.}
    \label{projection}
    \vspace{-3mm}
\end{figure}
\subsection{Human Detection}

As we mentioned in Section \ref{introduction}, RF signals are sparse and susceptible to noise. Feeding the whole AoA-ToF heatmaps into the decoder network will not only make it hard to convergence, but also bring unnecessary computational overhead. Therefore, it is necessary to perform detection first to filter out noise and make model focus on signals around targets.

To achieve this, the two RF sequences generated by signal processing module are input to the human detection module. 
Two identical encoders, which are composed of convolutional layers with skip connections, are utilized to extract horizontal and vertical features. Then, the horizontal feature is fed into a Region Proposal Network (RPN) \cite{ren2015faster} to propose candidate regions. The horizontal box regression module outputs refined bounding-box coordinates with confidence value.

Here we only apply RPN on the horizontal feature, and corresponding vertical regions are determined according to the combination of horizontal positions and a fixed height range. The reason is that: in most scenarios, human activities are in a fixed range of height which makes it simple to locate in vertical. 
Each horizontal region and its vertical counterpart indicate a 3D bounding box of human. We then crop horizontal and  vertical features by applying RoIAlign respectively to make following modules zoom into the vicinity of the human targets.

\begin{figure}[ht]
    \centering
    \includegraphics[width=1\columnwidth]{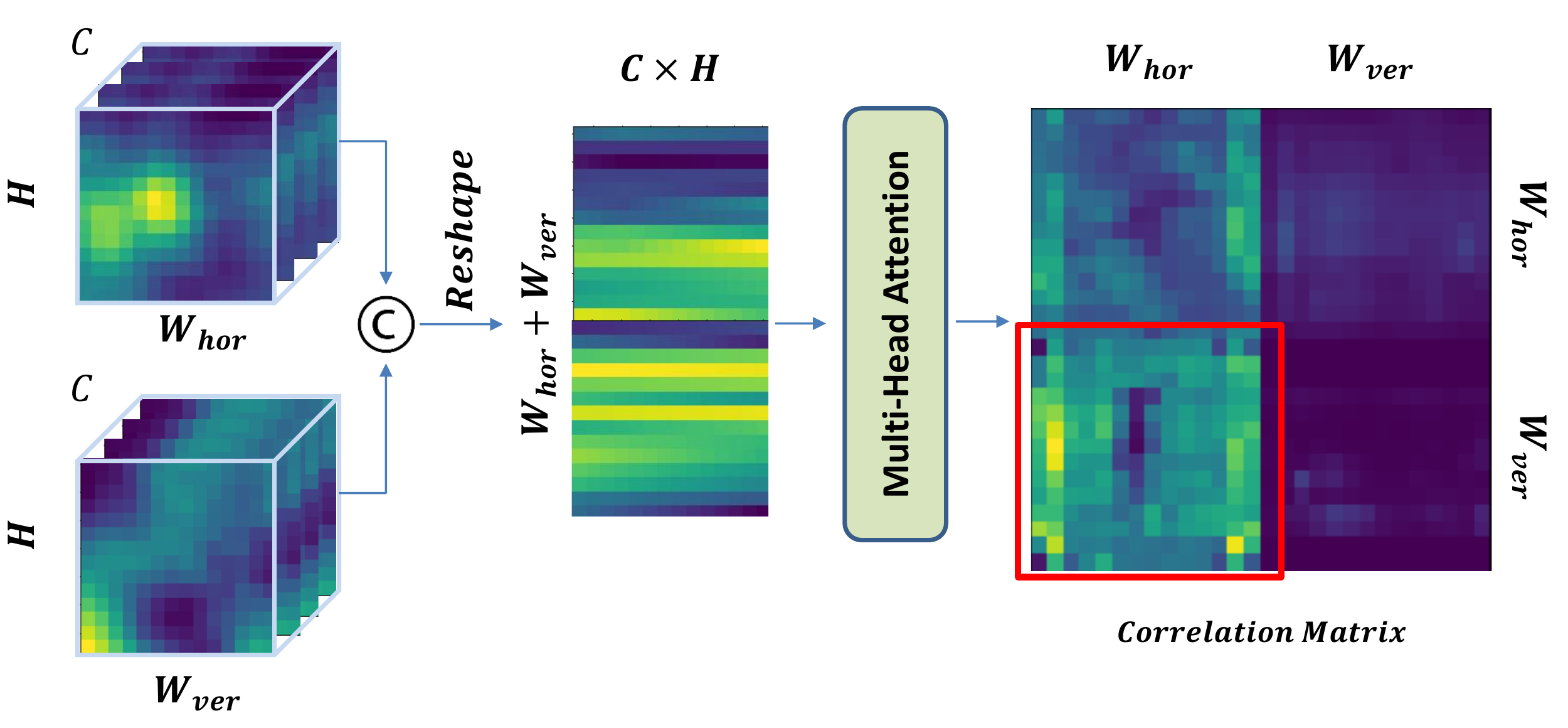}
    \caption{The pipeline of our Multi-Head Fusion module: our proposed Multi-Head Fusion module takes features of shape ($C \times H \times W_{hor/ver} $) which are extracted from horizontal and vertical signal plane as input, reshapes and concatenates the features into $(W_{hor} + W_{ver}) \times (C \times W)$, and then feeds the reshaped features into multi-head attention layers to obtain correlation matrix. We take lower-left part of the correlation matrix as output.}
    \label{fusion}
    \vspace{-3mm}
\end{figure}

\subsection{Mask Generation}
Our mask generation module aims to generate vision-like silhouette results. In common RGB-image-based segmentation tasks, the silhouette results are fundamentally projections of targets from real-world coordinates to imaging plane. In our situation, this process is different. We use two FMCW radars perpendicular to each other to collect horizontal and vertical RF signals. Meanwhile, our expected result imaging plane is orthogonal to two signal planes, as illustrated in Figure \ref{perpendicular}. Thus, in order to locate target positions and generate corresponding silhouette maps in imaging plane, a projection process from real world coordinates to imaging plane, similar to pin-hole camera model, should be conducted. 
Specifically, assuming that $\bm{P}$ is our result plane expressed as $\bm{Z}=r$ in 3D coordinate system, the projection process from point $(x, y, z)$ in 3D space to $(x_p, y_p)$ of 2D plane $\bm{P}$ can be expressed as follows
\begin{equation}
    \label{transformation}
    \begin{pmatrix}
    x_p \\ y_p \\
    \end{pmatrix}
    =
    \begin{bmatrix}
    r & & p_x \\ 
    & r &p_y \\
    & & 1
    \end{bmatrix}
    \begin{pmatrix}
    x \\ y \\ z \\
    \end{pmatrix} ,
\end{equation}
where $p_x$ and $p_y$ are the offset in plane $P$.
Using Equation \ref{transformation}, vertices of 3D bounding boxes can be projected into imaging plane which indicate target human positions.


The aforementioned procedure converts detection results from two signal planes into imaging plane, which indicates correct human positions in imaging plane. To depict detailed silhouette maps, a conversion and fusion process should also be applied in feature level. Therefore, as depicted in Figure \ref{fusion}, we adopt a decoder structure with Multi-Head Fusion module which is composed of four multi-head attention layers inspired by \cite{vaswani2017attention}. 
As depicted in Figure \ref{framework}, horizontal and vertical features are passed to a Multi-Head Fusion module to get fused features which will be taken by decoder network as input to generate silhouette results. 
Specifically, as illustrated in Figure \ref{fusion}, horizontal/vertical features of shape ($C \times H \times W_{hor/ver}$) are flattened on height and channel dimensions to get $W_{hor}/W_{ver}$ vectors of shape ($ 1 \times (C \times H) $), which represent features of each position of along width dimension, and concatenated together. Then, the Multi-Head Fusion module takes reshaped and concatenated horizontal and vertical features as input and output low-resolution silhouette heatmaps. Multi-head layers learn the correlation weight which represents correlations between any two elements in the vector sequence. We only need correlations between all the horizontal and vertical positions. Therefore, we crop out the lower-left part of the correlation matrix and feed them into decoder network which is composed of three deconvolution-convolution blocks to increase the output resolution. 
The experimental results demonstrate that our well-designed Multi-Head Fusion module shows great power in combining horizontal and vertical features. It improves the performance of RFMask on our dataset and wipes out the accuracy gap between simple and complex subsets. Qualitative demonstrations can be find in Section \ref{experiment}.
Finally, we combine two parts of results together by pasting mask results into corresponding projection bounding boxes in result plane to get final silhouette maps. 

\begin{figure*}[ht]
    \centering
    \subfloat[]{
        \begin{minipage}[b]{0.11\linewidth}
            \includegraphics[width=1\textwidth]{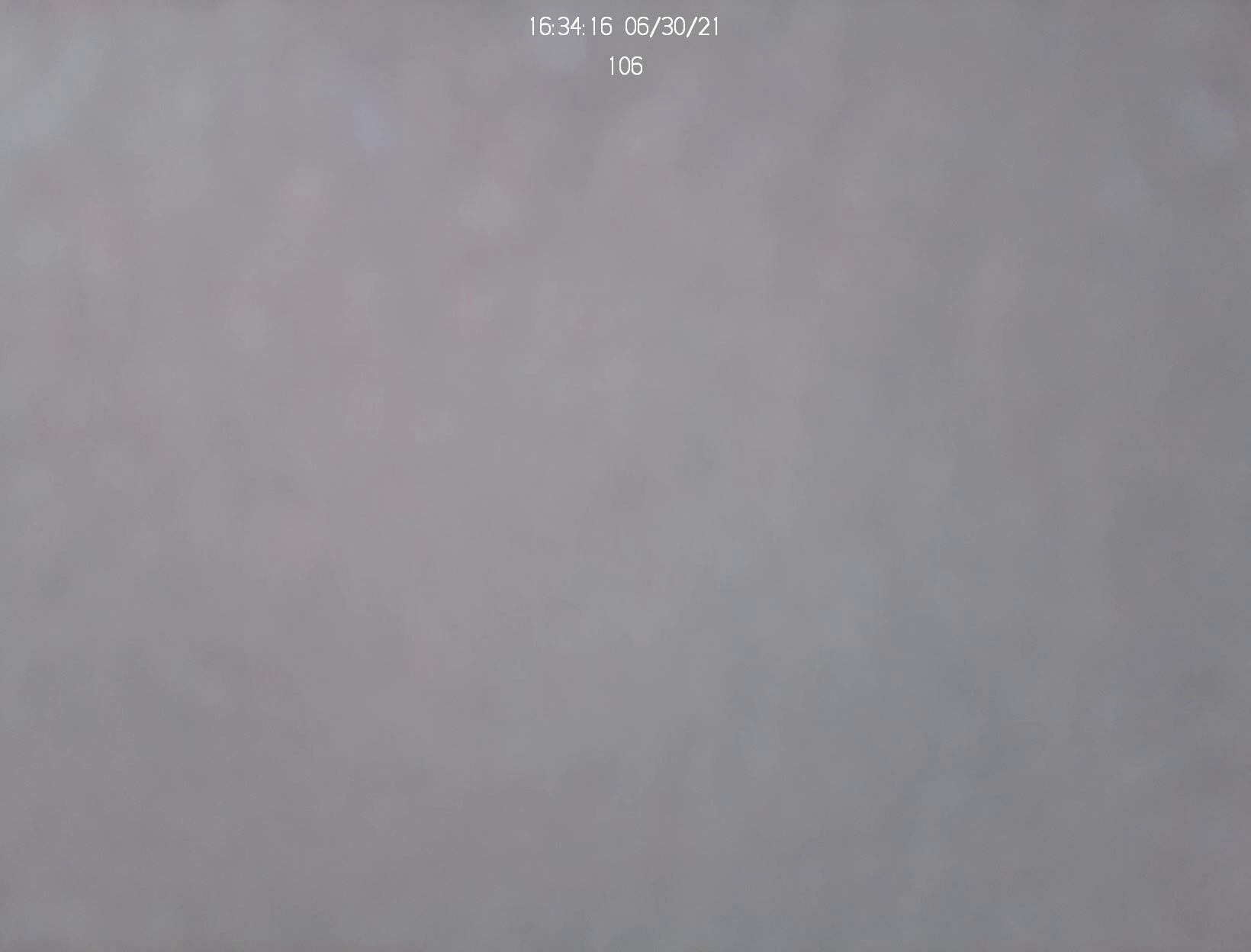} \\
            \vspace{-0.4cm}
            \includegraphics[width=1\textwidth]{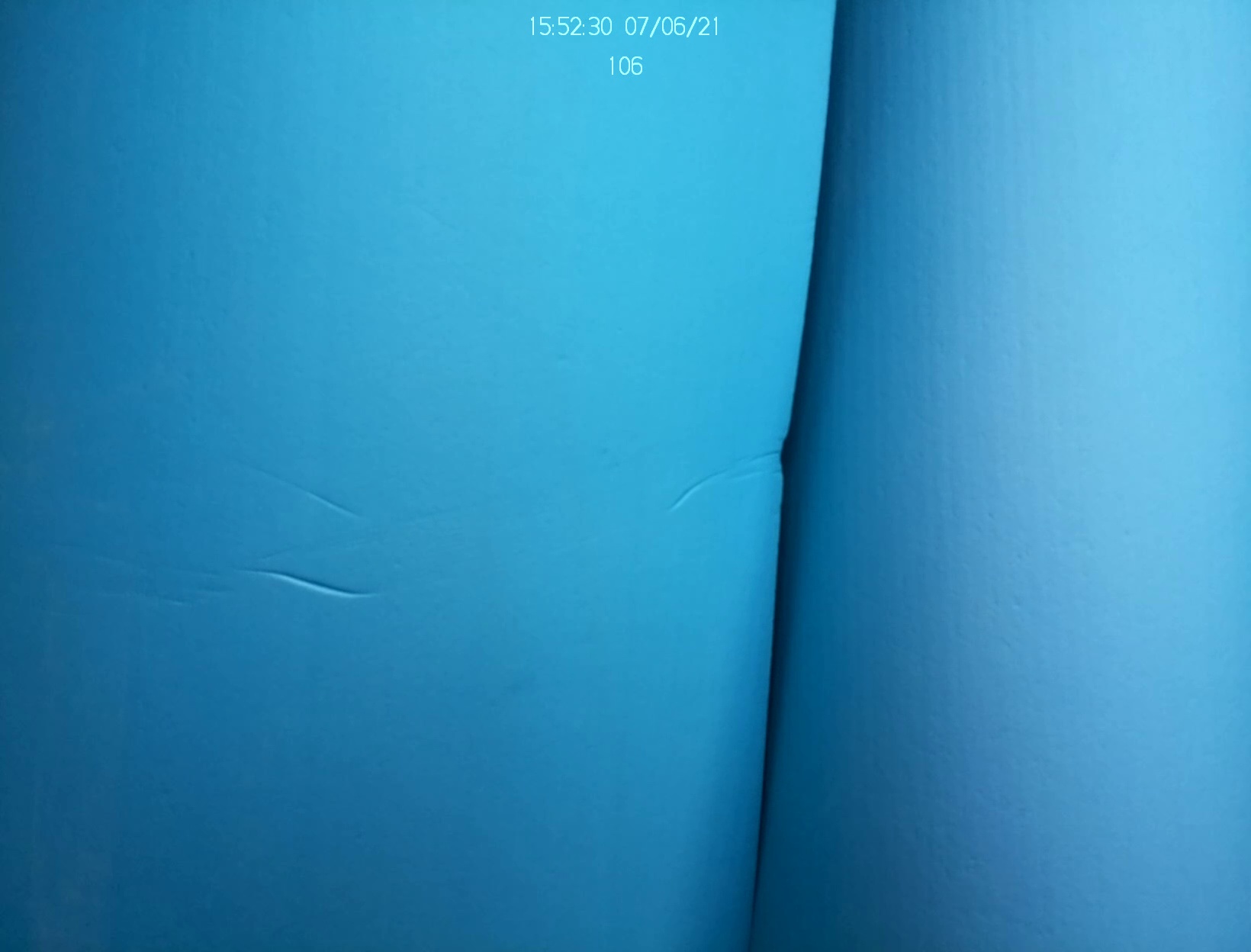} \\
            \vspace{-0.4cm}
            \includegraphics[width=1\textwidth]{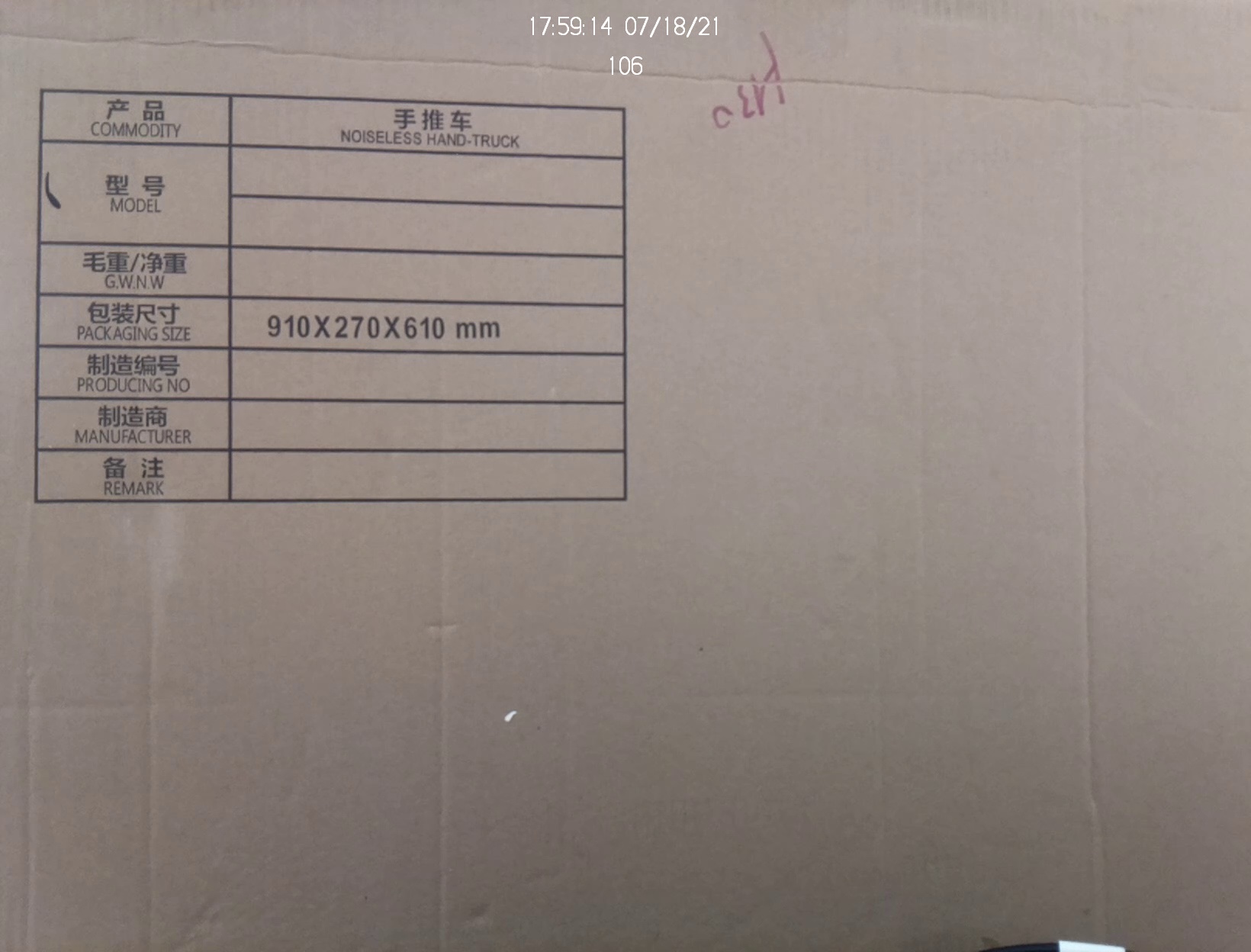} \\
            \vspace{-0.4cm}
            \includegraphics[width=1\textwidth]{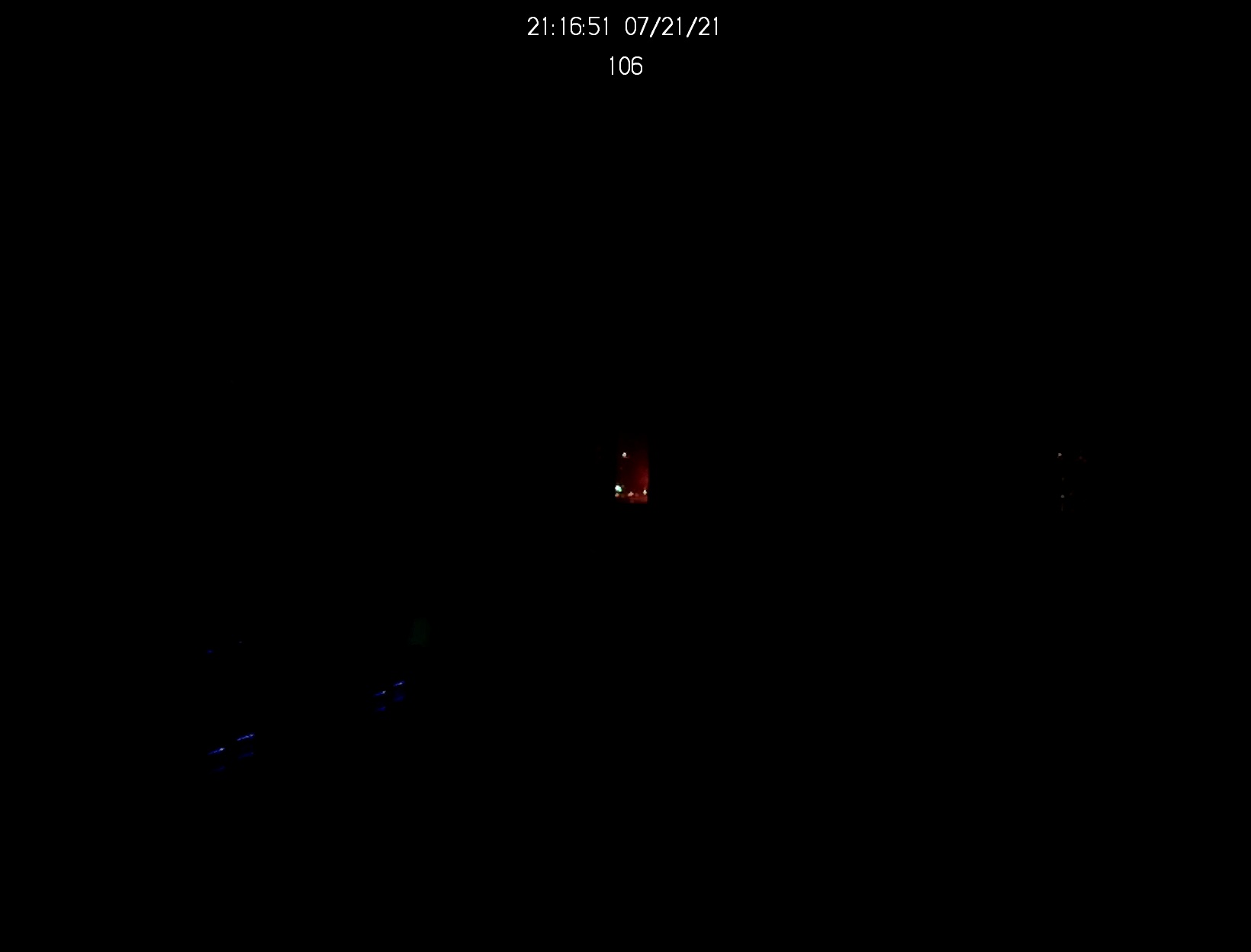} \\
            \vspace{-0.3cm}
        \end{minipage}
    }
    \subfloat[]{
        \begin{minipage}[b]{0.11\linewidth}
            \includegraphics[width=1\textwidth]{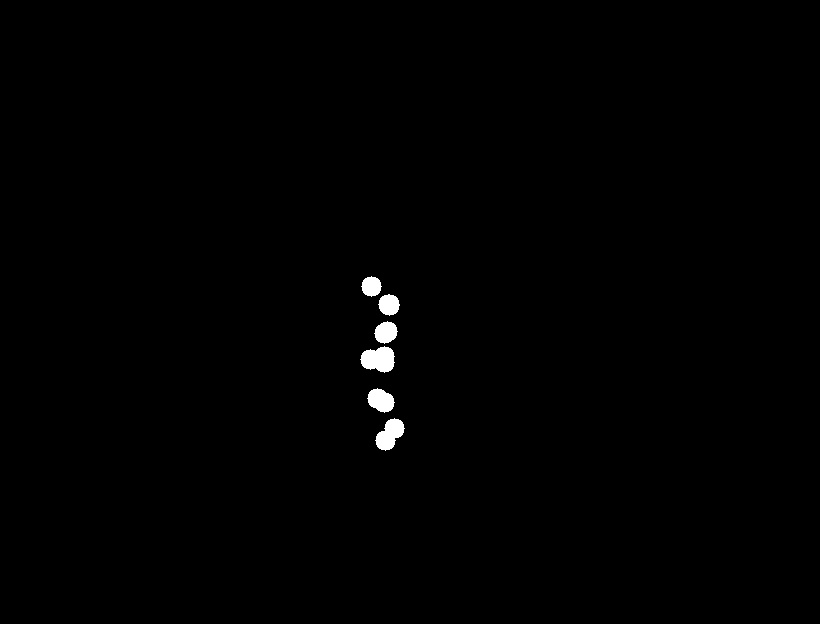} \\
            \vspace{-0.4cm}
            \includegraphics[width=1\textwidth]{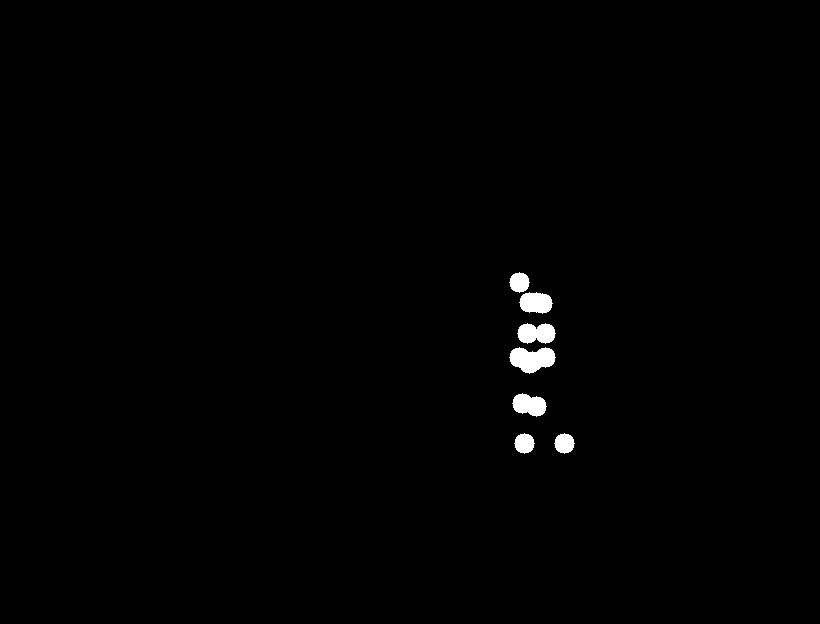} \\
            \vspace{-0.4cm}
            \includegraphics[width=1\textwidth]{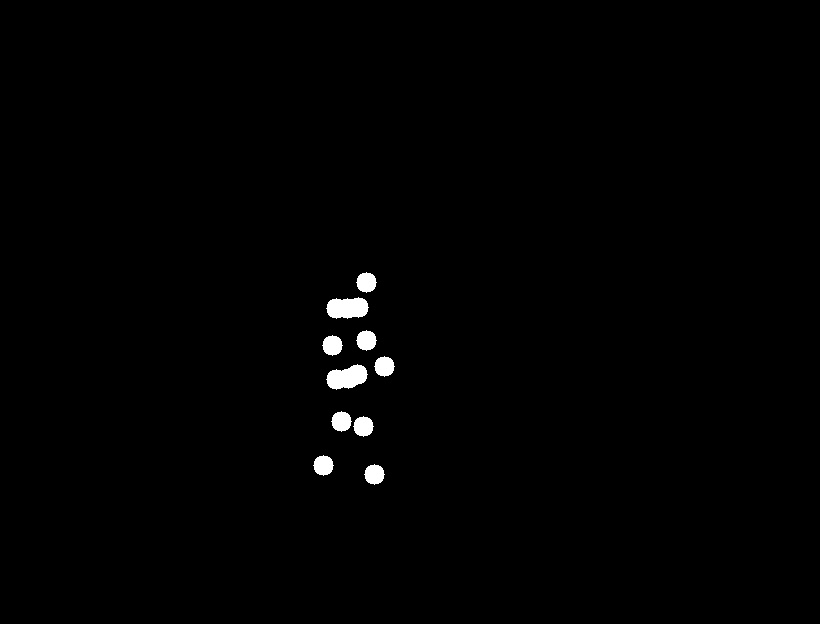} \\
            \vspace{-0.4cm}
            \includegraphics[width=1\textwidth]{pictures/dark_cv_mask.jpg} \\
            \vspace{-0.3cm}
        \end{minipage}
    }
    \subfloat[]{
        \begin{minipage}[b]{0.11\linewidth}
            \includegraphics[width=1\textwidth]{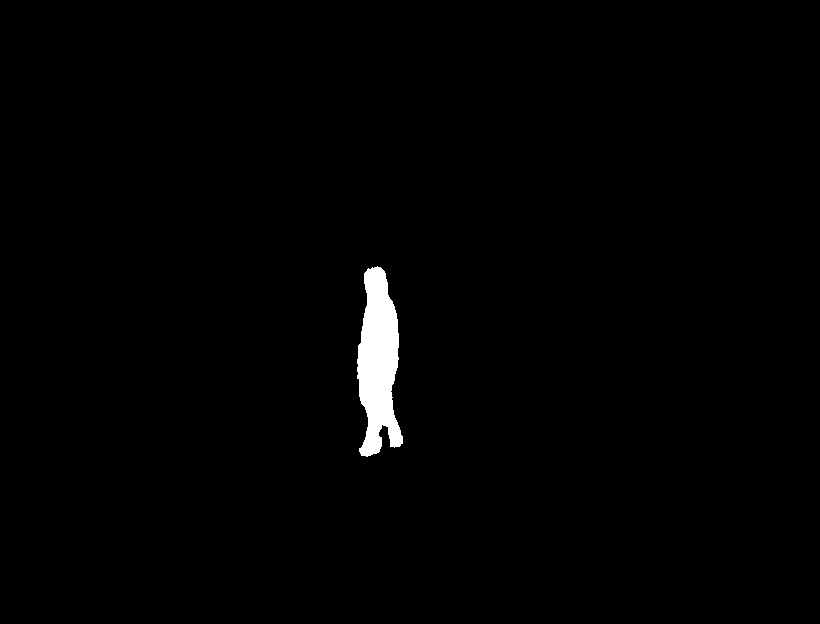} \\
            \vspace{-0.4cm}
            \includegraphics[width=1\textwidth]{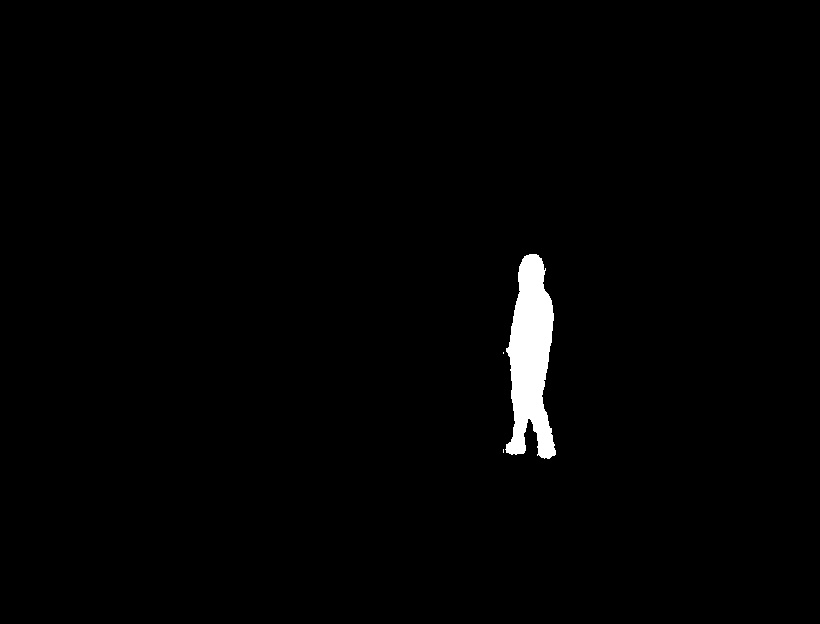} \\
            \vspace{-0.4cm}
            \includegraphics[width=1\textwidth]{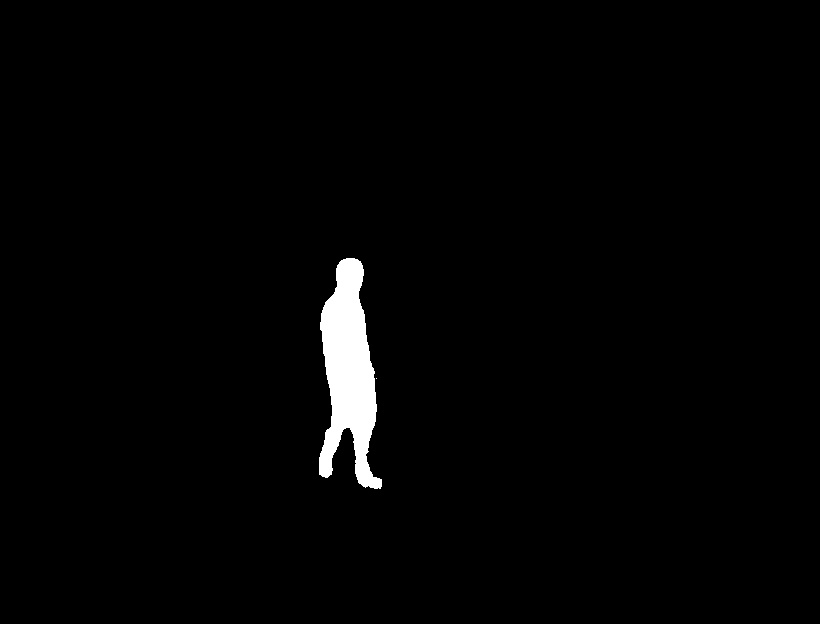} \\
            \vspace{-0.4cm}
            \includegraphics[width=1\textwidth]{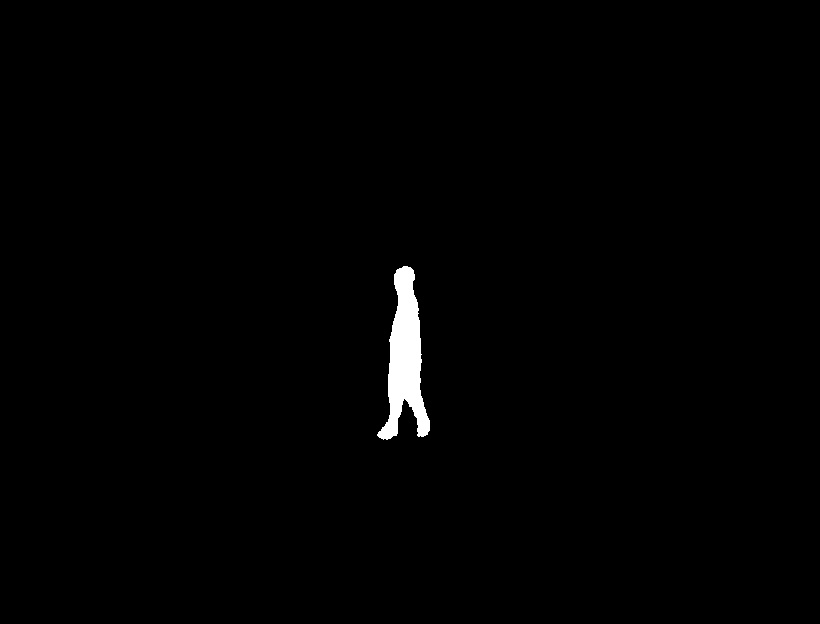} \\
            \vspace{-0.3cm}
        \end{minipage}
    }
    \subfloat[]{
        \begin{minipage}[b]{0.11\linewidth}
            \includegraphics[width=1\textwidth]{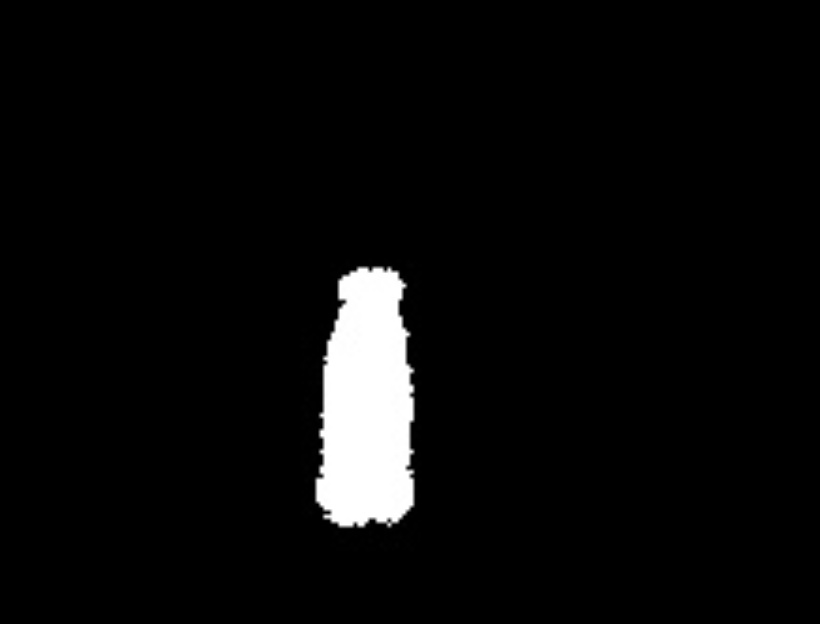} \\
            \vspace{-0.4cm}
            \includegraphics[width=1\textwidth]{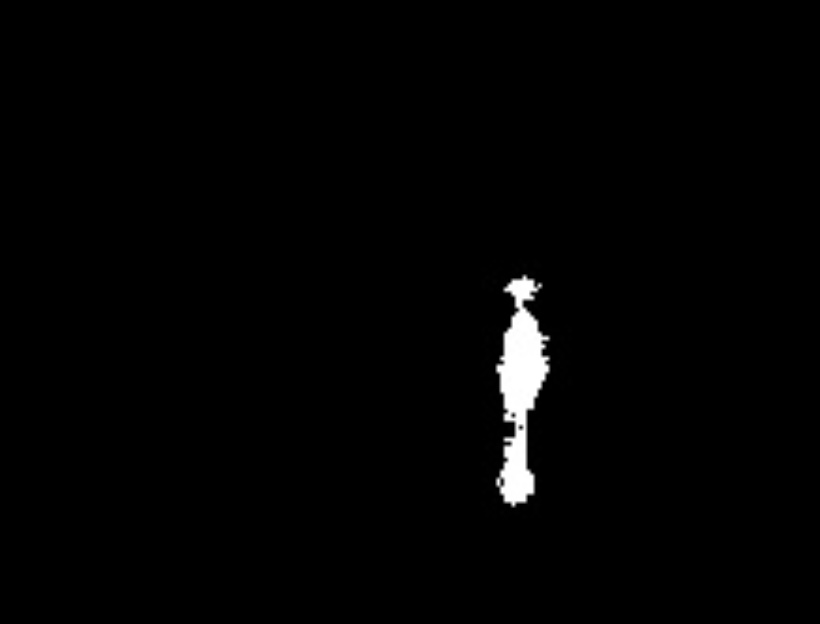} \\
            \vspace{-0.4cm}
            \includegraphics[width=1\textwidth]{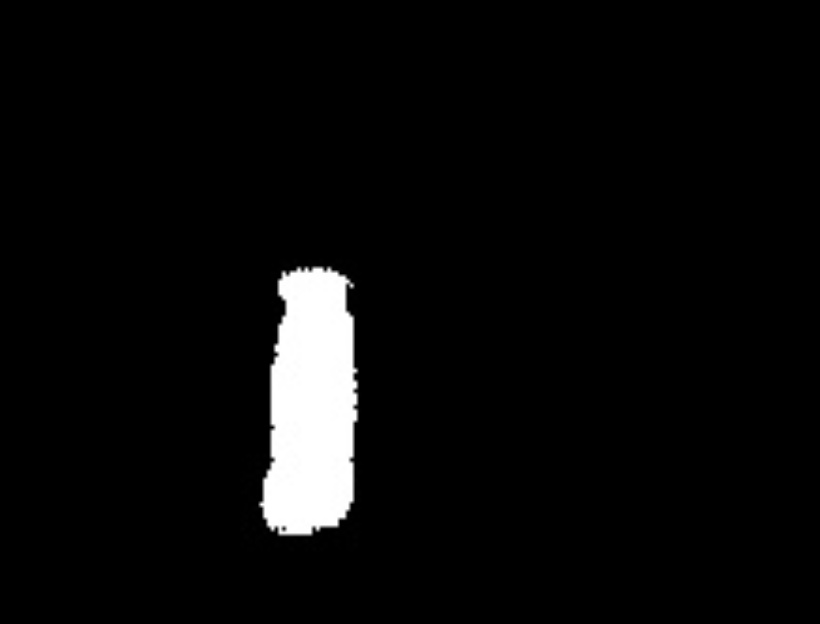} \\
            \vspace{-0.4cm}
            \includegraphics[width=1\textwidth]{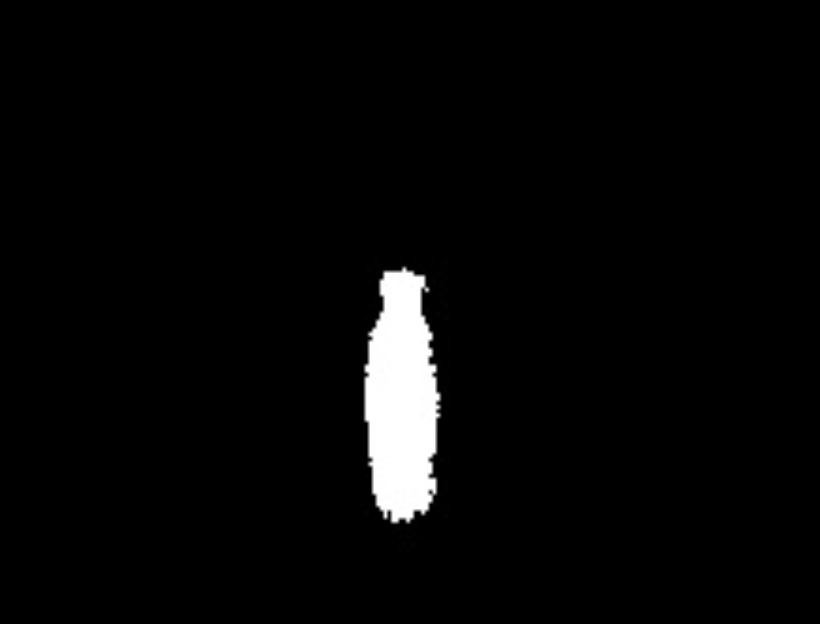} \\
            \vspace{-0.3cm}
        \end{minipage}
    }
    \subfloat[]{
        \begin{minipage}[b]{0.11\linewidth}
            \includegraphics[width=1\textwidth]{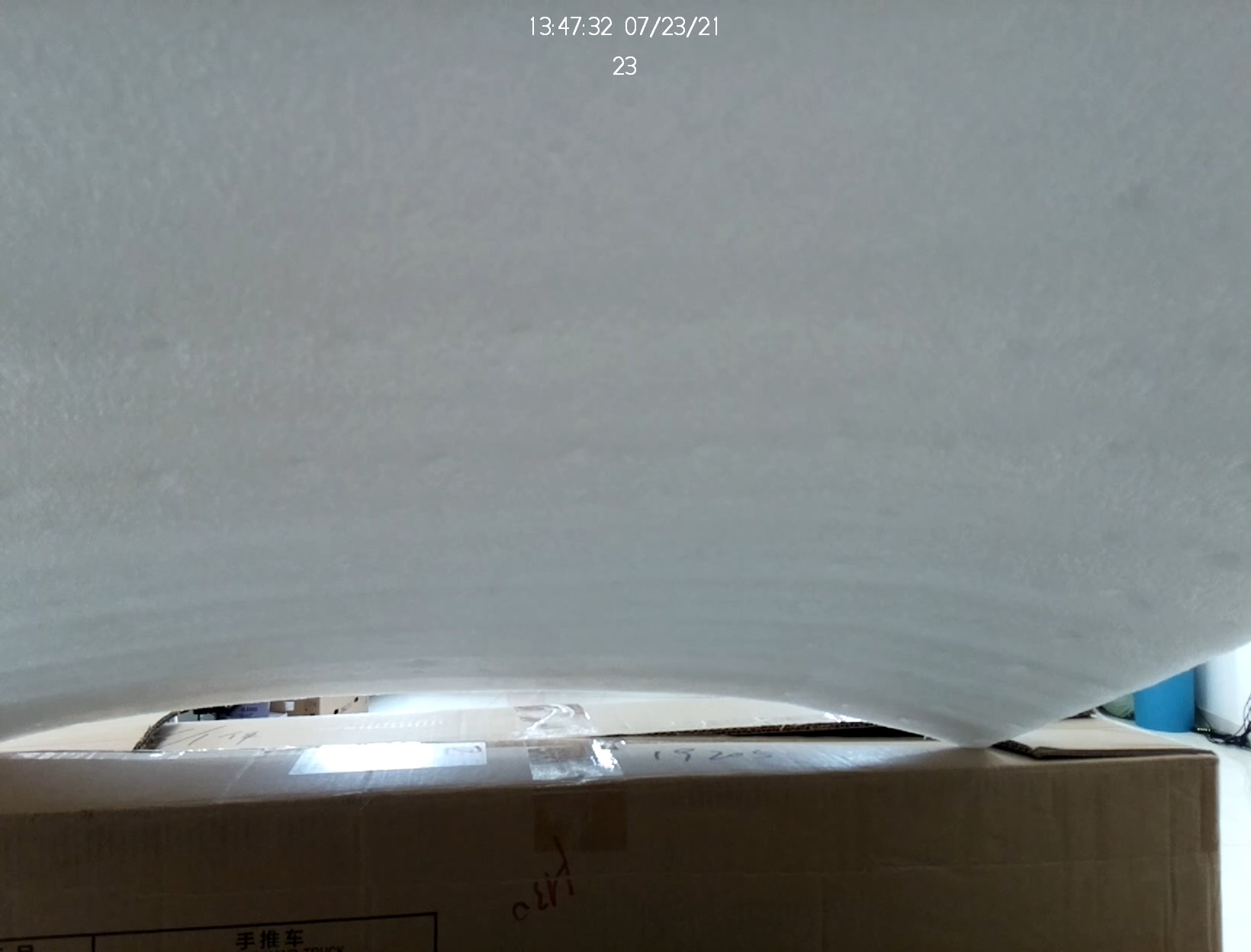} \\
            \vspace{-0.4cm}
            \includegraphics[width=1\textwidth]{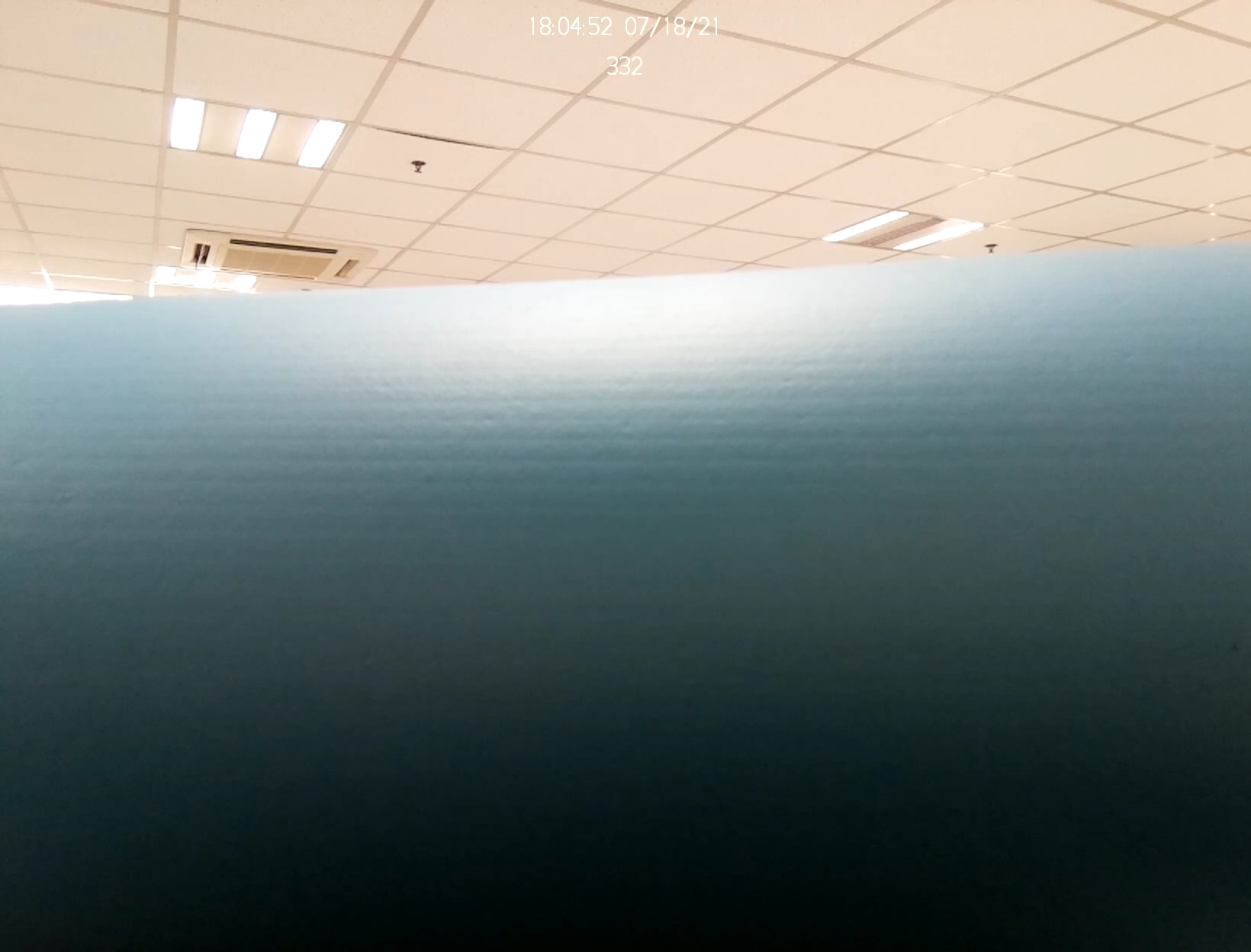} \\
            \vspace{-0.4cm}
            \includegraphics[width=1\textwidth]{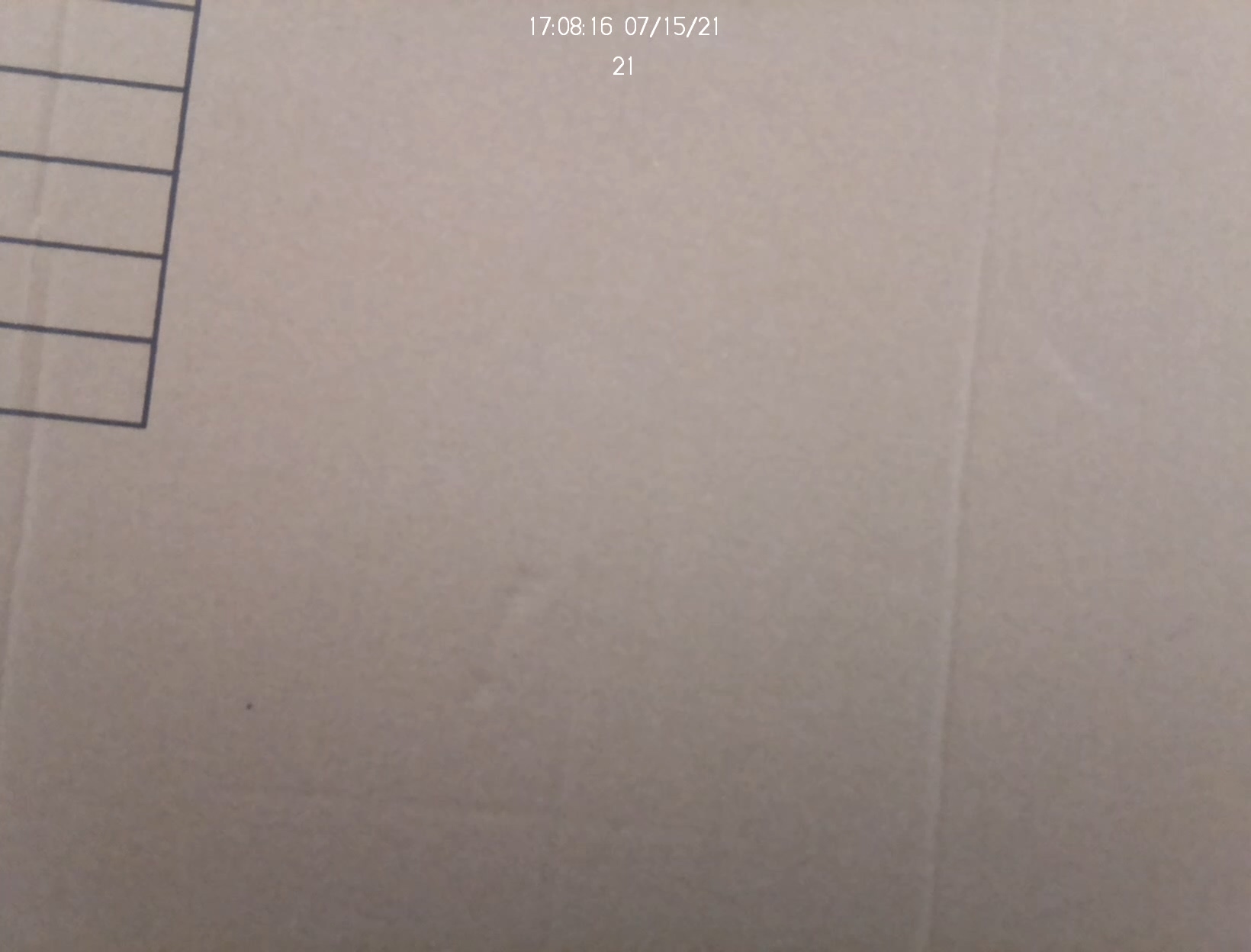} \\
            \vspace{-0.4cm}
            \includegraphics[width=1\textwidth]{pictures/d14.jpg} \\
            \vspace{-0.3cm}
        \end{minipage}
    }
    \subfloat[]{
        \begin{minipage}[b]{0.11\linewidth}
            \includegraphics[width=1\textwidth]{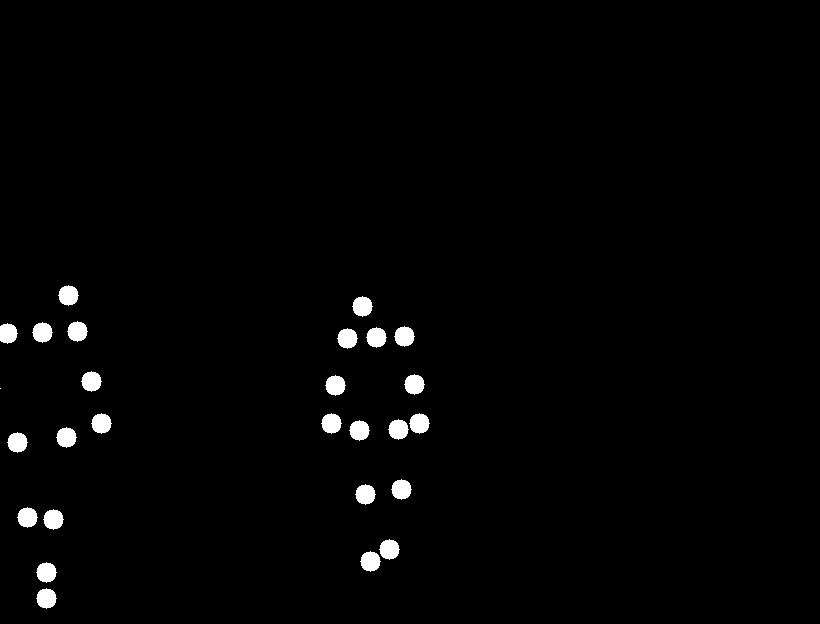} \\
            \vspace{-0.4cm}
            \includegraphics[width=1\textwidth]{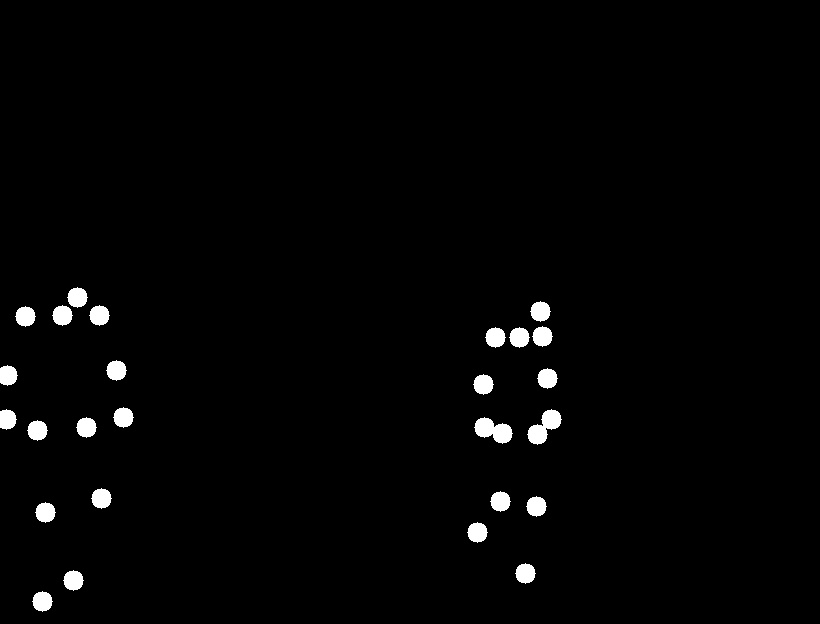} \\
            \vspace{-0.4cm}
            \includegraphics[width=1\textwidth]{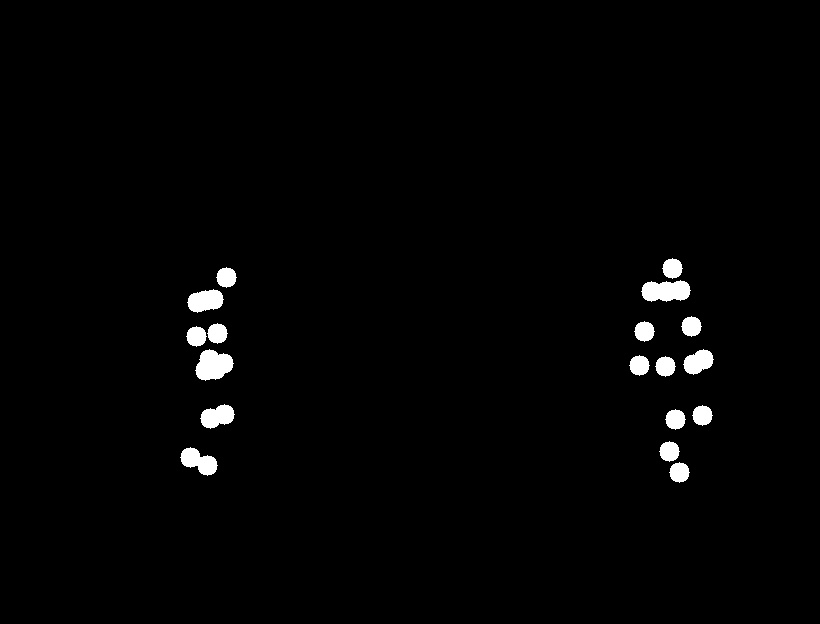} \\
            \vspace{-0.4cm}
            \includegraphics[width=1\textwidth]{pictures/dark_cv_mask.jpg} \\
            \vspace{-0.3cm}
        \end{minipage}
    }
    \subfloat[]{
        \begin{minipage}[b]{0.11\linewidth}
            \includegraphics[width=1\textwidth]{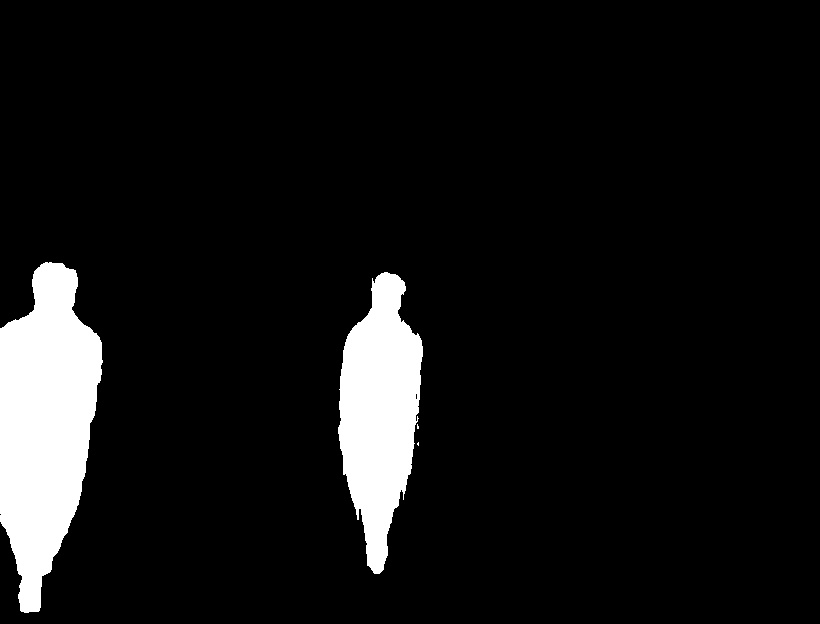} \\
            \vspace{-0.4cm}
            \includegraphics[width=1\textwidth]{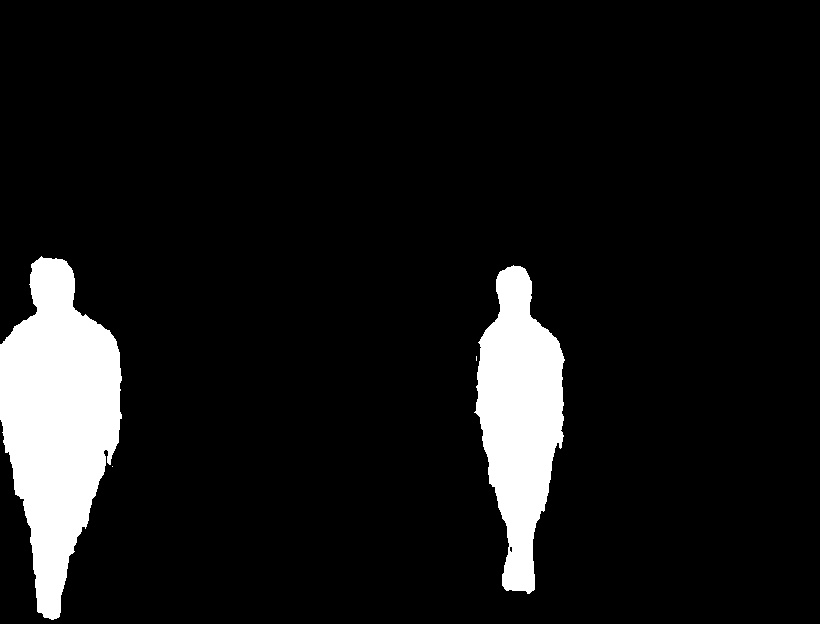} \\
            \vspace{-0.4cm}
            \includegraphics[width=1\textwidth]{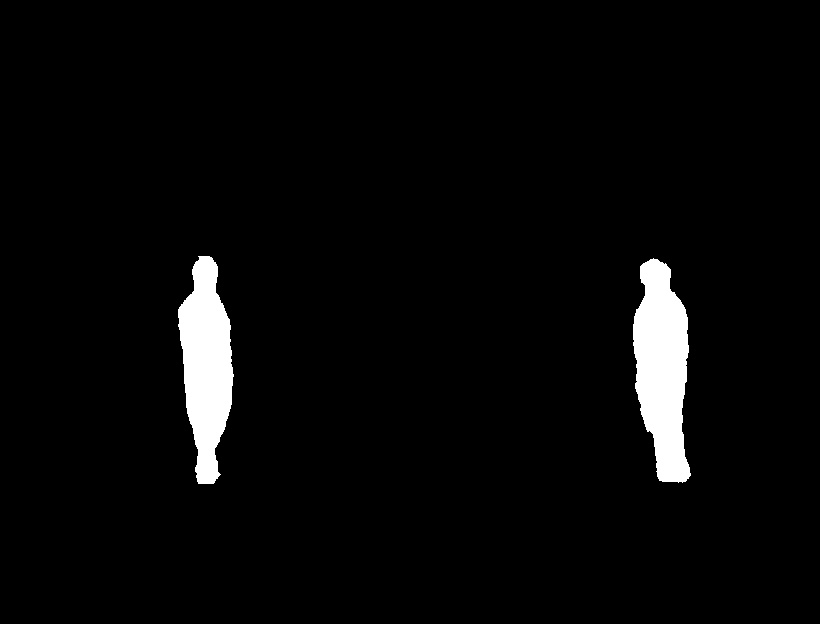} \\
            \vspace{-0.4cm}
            \includegraphics[width=1\textwidth]{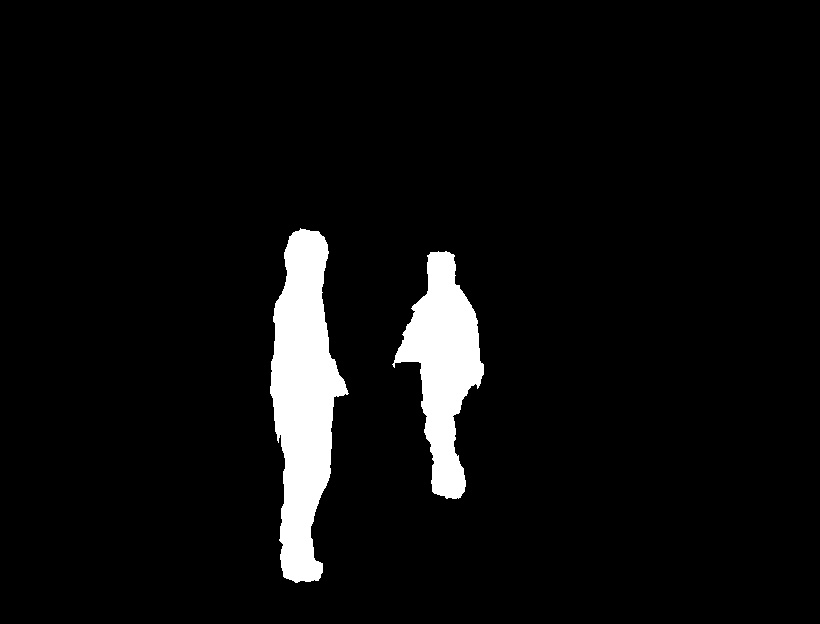} \\
            \vspace{-0.3cm}
        \end{minipage}
    }
    \subfloat[]{
        \begin{minipage}[b]{0.11\linewidth}
            \includegraphics[width=1\textwidth]{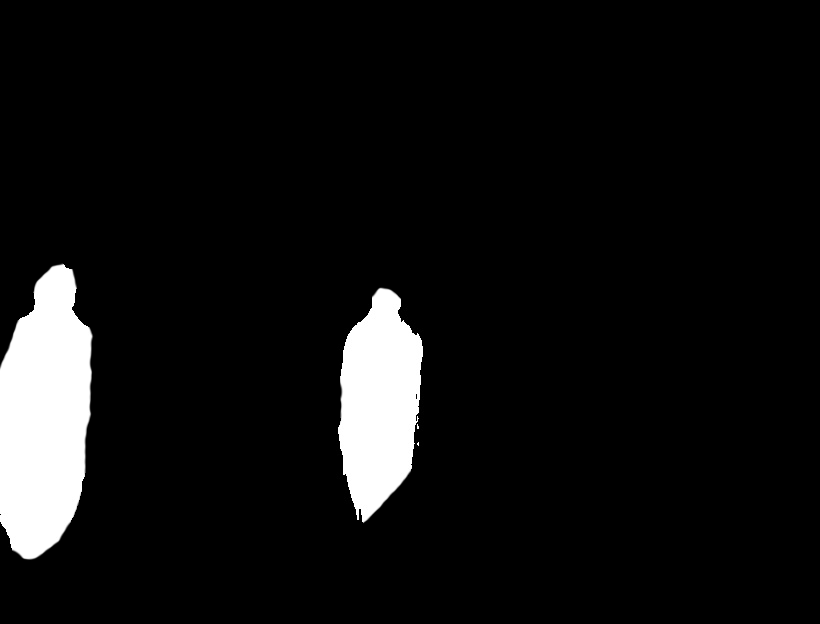} \\
            \vspace{-0.4cm}
            \includegraphics[width=1\textwidth]{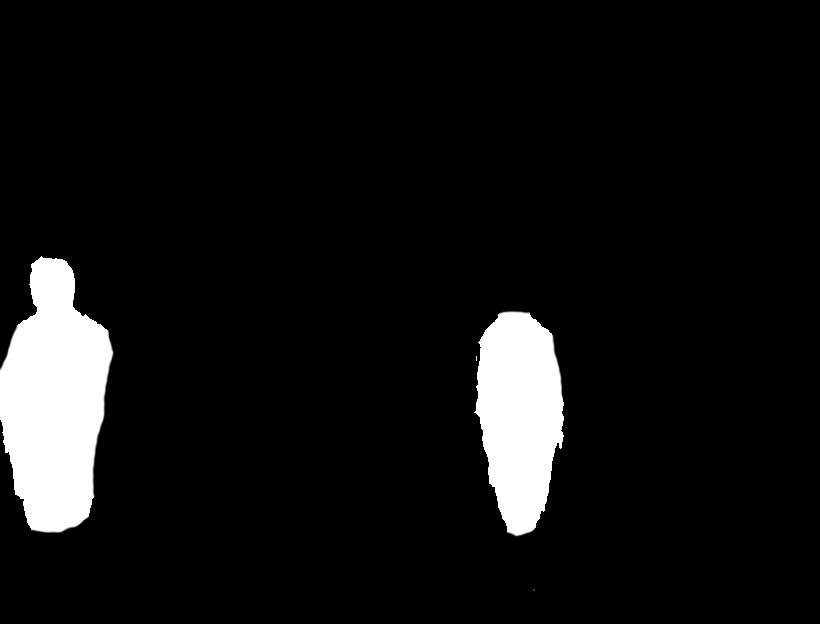} \\
            \vspace{-0.4cm}
            \includegraphics[width=1\textwidth]{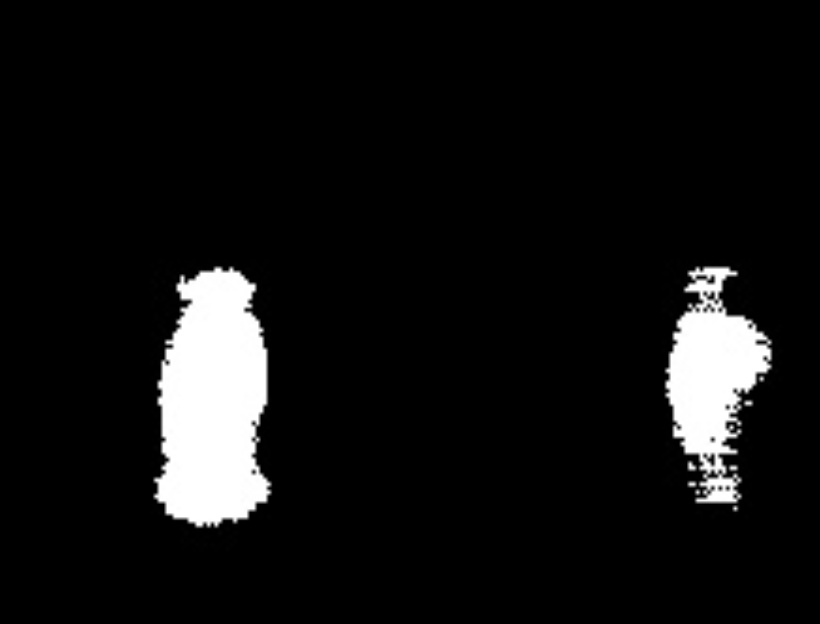} \\
            \vspace{-0.4cm}
            \includegraphics[width=1\textwidth]{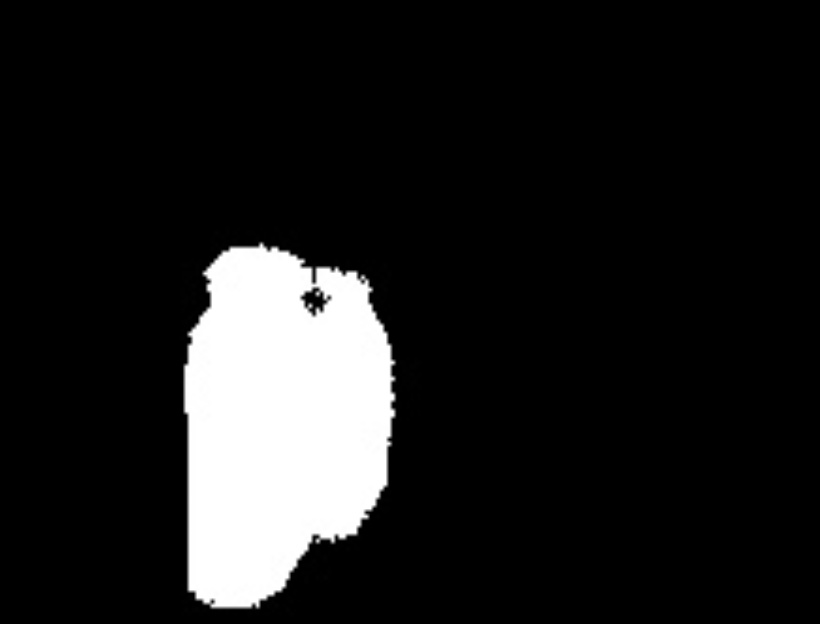} \\
            \vspace{-0.3cm}
        \end{minipage}
    }
    \caption{Results under occlusion and low illumination scenarios: (a) the camera view; (b) the reference ground-truth obtained by reprojecting 3D human keypoints into occluded camera plane; (c) the results of RFMask; (d) the results of RFPose(12) (e) the camera view; (f) the reference ground-truth obtained by reprojecting 3D human keypoints into occluded camera plane; (g) the results of RFMask. (h) the results of RFPose(12)}
    \label{occlution_res}
    \vspace{-3mm}
\end{figure*}

\subsection{Loss Function}
RFMask is a typical two stage network, which first outputs human bounding boxes, and then predicts a binary mask for each bounding box. Therefore, two loss functions are involved in RFMask to supervise the training procedure. 

To supervise the regression of human positions, we define a multi-task loss on each predicted bounding-box as follows
\begin{equation}
    \begin{aligned}
    \label{detect_loss}
    L_{detect}(p,p^u,v, t^u)&=L_{cls}(p,p^u) \\
    &+\lambda_{det}\ [u \ge 1]L_{box}(t^u, v) ,
    \end{aligned}
\end{equation}
where $p$ and $p^u$ are the predicted class scores and ground-truth class scores, $u$ is the ground-truth class labels, $v$ and $t^u$ are the predicted and ground-truth coordinates. $L_{cls}$ is \textit{binary-cross-entropy} loss over two categories: background and human, $L_{box}$ is the \textit{smooth-}$l_1$ loss. $[u \ge 1]$ means only foreground classes have contributions to total loss. $\lambda_{det}$ is the balancing weight. 

Our mask generation branch has one $Km^2$ dimensional output for each predicted bounding-box ($K$ denotes the number of all possible classes), which represents $K$ binary masks with resolution $m \times m$. The corresponding loss function for the mask generation branch is
\begin{equation}
    \label{mask_loss}
    L_{mask}=\frac{1}{N_{box}} \sum_{i=1}L_m(m_{i,k}, m_k^*) ,
\end{equation}
where $i$ is the index of predicted boxes, $N_{box}$ denotes the number of detected boxes, $k$ is the ground-truth class label, $m_{i, k}$ denotes the predicted mask result on the $k$-th mask, $m_k^*$ is the corresponding ground-truth, $L_m$ is the \textit{binary-cross-entropy} loss.

In summary, the overall loss function can be written as follows
\begin{equation}
    \label{loss}
    L=L_{detect}+L_{mask} .
\end{equation}

\section{Experiment}
\label{experiment}
Since this is, to our best knowledge, the first work which achieves human silhouette generation from the millimeter wave radio signals, there is no existing public dataset that can be used to evaluate our method. 
In this paper, we create a multi-modal dataset that contains thousands of radio frames and corresponding optical camera images of human activity. In the following, we first introduce our dataset in detail and then demonstrate the performance of the proposed simple baseline on segmenting human silhouette from RF signals under different scenarios including single-person, multi-person, low illumination, and occlusion. Due to our two stage design principle, which first detects target positions and then generates corresponding mask results, we will also present our detection result in the form of average precision (AP) to demonstrate the location error. Meanwhile, the effectiveness of our proposed Multi-Head Fusion module will also be demonstrated.
\begin{figure}[h]
    \centering
    \includegraphics[width=1\linewidth]{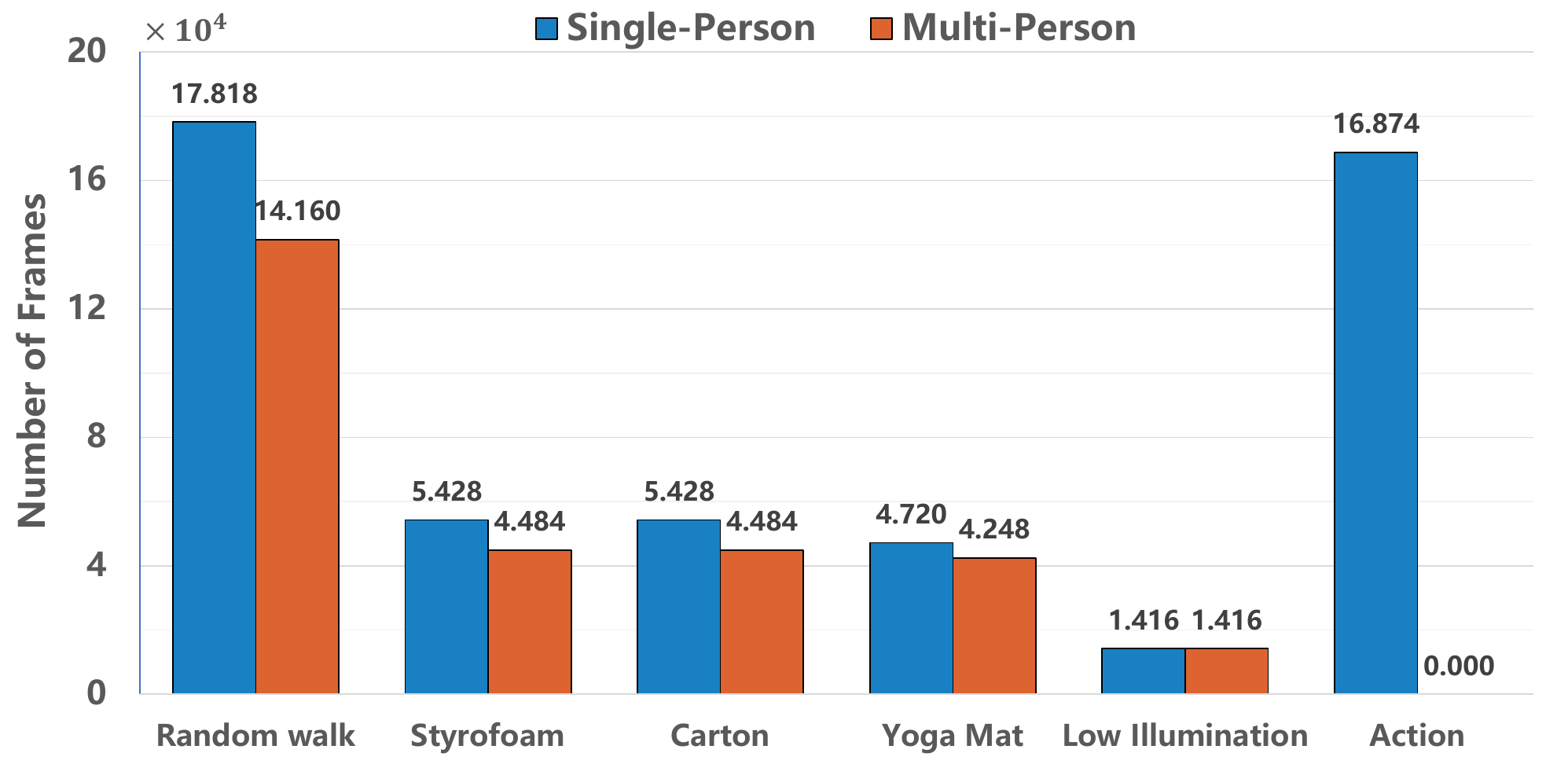}
    \caption{Statistics of our dataset. The label of x-axis represents the environment settings or human actions when capturing data: Random walk represents people walking randomly in the scene; Styrofoam, Carton, and Yoga Mat represent people walking randomly with the 13th camera and dual-radars occluded by corresponding materials; Low Illumination represents people walking randomly with no illumination; Action represents people performing stand, walk, squat, and sit in the scene. The label of y-axis represents the number of frames. }
	\label{dataset_category}
    \vspace{-3mm}
\end{figure}

\subsection{Dataset}
We aim to create a multi-modal dataset including the optical images and radio signals to study the influence of human behaviors on the RF signals, which can be utilized for not only our segmentation task but also other tasks such as human pose estimation, action recognition, and position tracking. To achieve this, we build a multi-camera system with 13 camera nodes to capture visual information and a dual-radar system with two FMCW radars to capture RF signals.  

\subsubsection{Multi-camera system}
In our multi-camera system, each camera node is composed of a Raspberry Pi, a camera module, and a Power-over-Ethernet (PoE) module, which makes each node portable and easy to deploy. Among the 13 camera nodes, 12 nodes are in fixed locations, while the last one is always along with the FMCW radars. 
We perform pairwise calibration between adjacent cameras with respect to the same real-world coordinates using the calibration method proposed in \cite{zhang2000flexible}.
The multi-camera system is mainly utilized to calculate 3D human keypoints. Once the 3D keypoints are detected, the bounding-box of human target can be obtained by calculating smallest bounding cube of 3D keypoints belonging to the same person, which is used to supervise the training of detection stage in our RFMask framework. Meanwhile, although the ground-truth silhouette results can not be obtained in occlusion scenarios, we can still calculate reference ground-truth by re-projecting 3D keypoints back to the occluded imaging plane to indicate correct human positions and actions. With well-calibrated camera system, we can also provide more forms of annotations, which makes our own collected dataset applicable to many other wireless sensing tasks.

\subsubsection{Dual-radar system}
We use two identical TI MMWCAS-RF-EVM FMCW radars to transmit and receive RF signals, and each radar is equipped with 12 transmitters and 16 receivers with Multiple-In Multiple-Out (MIMO) antenna array. The time division multiplexing is utilized to achieve orthogonality among different transmitters. Within one frame, 12 transmitters transmit the RF signals successively to all 16 receivers. The antennas are designed in the way that the azimuth resolution is high, i.e., the shape of virtual antenna array equipped by each radar is $86 \times 1$. We place the two radars perpendicularly to maintain high resolution in both horizontal and vertical directions. To avoid mutual interference, the sweep range of the first radar is set as from $77\ GHz$ to $78.23\ GHz$ while that of the second radar is set as from $79\ GHz$ to $80.23\ GHz$, both with $1.23\ GHz$ bandwidth. 

\subsubsection{Synchronization}
The radars and multi-camera system are synchronized using the network time protocol (NTP). All devices including radars and camera nodes are synchronized with a local time server in advance. When capturing, a timestamp of 15 seconds later from now is sent to radars and each camera node of multi-camera system simultaneously through TCP connection, and all devices will keep waiting until the timestamp comes. In this way, the time misalignment is mainly caused by the synchronization error between each device and local time server, which is independent of each device and achieves millisecond-level synchronization error. 
\subsubsection{Data collection}
While we aim to create a large-scale dataset under various environments, deploying the multi-camera system when encountering a new environment is extremely time-consuming. 
Due to the consideration of maintaining data diversity and reducing the burden of data collection, we choose to place radars at different locations in the same room which deployed multi-camera system in advance, to simulate different environments for RF signals. After each change of radar location, we only need to move the 13th camera node along with the radars, and re-calibrate the multi-camera system.
The frame rate of the camera system is set to 10 while that of the radar is set to 20. 
We collect the data at 10 different environments under 11 different conditions, including random walk with no occlusion, random walk under styrofoam, carton and yoga mat occlusion, respectively, random walk under low illumination, random action such as stand, walk, squat and sit, for both single-person and multi-person scenarios. The RF frame statistics of our dataset is shown in Figure \ref{dataset_category}. We also illustrate some examples of our dataset in Figure \ref{dataset}. 

\subsubsection{Ground-truth generation}
In order to generate silhouette results, we perform Mask R-CNN \cite{He_2017_ICCV} frame by frame on the video stream captured by the 13th camera.
We also adopt OpenPose \cite{fang2017rmpe} to generate 2D human keypoints, which is applied to video streams of all 13 camera nodes. Once 2D skeletons of the same person from different views is obtained, triangulation can be used to generate corresponding 3D skeletons. Assuming that there is a particular 3D keypoint $k$ and its 2D projection of $i$-th camera node $k_i$, the triangulation process can be formulated as:
\begin{equation}
    k = \mathop{\arg\min}\limits_{k} \sum_{i \in I}||M_ik-k_i||_2^2,
\end{equation}
where $M_i$ is the homography matrix of $i$-th camera node, $I$ is the collection of camera index which is in $[0, 13]$. When dealing with multi-person scenario, we adopt $K$-means algorithm to identify 3D keypoints of different person using euclidean distance as similarity metrics.

\subsection{Results}
To the best of our knowledge, this is the first attempt to perform silhouette generation from millimeter wave radio signals. 
Hence, there is no existing works which can be utilized for comparison.   
To better illustrate the advantages of the proposed framework, we have noted that 2D keypoints detection has been achieved in \cite{zhao2018through}.
Therefore, the RFPose framework in \cite{zhao2018through} is adapted for silhouette generation. Specifically, we decrease some of the strides of spatial-temporal convolutions to maintain the resolution of the feature maps. We also modify the decoder of the RFPose, making it output human silhouette sequence instead of keypoint heatmap sequence. 

\subsubsection{Implementation Details}
In our implementation, our model is built with input length equal to 12. Residual network \cite{2016Deep} with feature pyramid module \cite{Lin_2017_CVPR} is adopted as our feature extraction backbone. We train our model on a single NVIDIA A100 GPU with batch size 90, initial learning rate $1.5^{-4}$. The cyclic cosine annealing strategy with cycle period 4 is adopted as our training schedule. 

\subsubsection{General Performance}
The quantitative results of RFMask are demonstrated in Table \ref{comparisons}, the proposed method achieves impressive performance. 
Specifically, RFMask achieves mask IoU of 0.706, 0.711 and 0.705 on single-person, multi-person and action subset, respectively. 
We also illustrate some qualitative results in Figure \ref{result}. As we expected, RFMask generates reasonable silhouette maps, where human actions including moving direction, arm swing, and complex movements are extracted from raw RF signals and converted to vision-like results. 

For comparison, we compare the proposed RFMask with the modified RFPose \cite{zhao2018through}. 
As zhao et al. described In \cite{zhao2018through}, original RFPose is equipped with an encoder of ten 3D convolution layers and a decoder of four 2D deconvolution layers. In our modified version of RFPose, We decrease the kernel stride of some middle 3D convolution layers in encoder to enlarge the resolution of output features which makes it more suitable to silhouette tasks. More importantly, we adopt silhouette sequence instead of keypoint heatmap sequence as ground-truth to supervise the training process of the RFPose. We also train the RFPose under four different input sequence length as zhao et al. did in \cite{zhao2018through}. Concretely, we adopt sequence length of 6, 20, 50, 100 frames corresponding to 0.2s, 0.6s, 1.6s, and 3.3s time elapse which is the same time elapse as the original RFPose settings. As for the proposed RFMask, we adopt sequence length of 4, 12 as input. In the following, we denote RFPose with input length 4 as RFPose(4) for brevity, and the same applies to RFPose(12), RFPose(32), RFPose(64), RFMask(4), RFMask(12).

Compared with RFPose, it is obvious that our RFMask generates more precise silhouette maps. In complex scenarios, such as multi-person subset (third row of Figure \ref{result}) and action subset (fourth row of Figure \ref{result}), the output of RFPose shows severe performance degradation and the integrity of generated silhouette is affected.
As illustrated in Table \ref{comparisons}, our proposed method outperforms RFPose at all subset (single-person, multi-person, and action subset). 
When the input sequence length is 4, our proposed RFMask(4) achieves mask IoU of 0.681 on single-person subset, 0.682 on multi-person subset, and 0.681 on action subset, which shows 0.017, 0.056, and 0.065 improvement on three subset respectively than its RFPose(4) counterpart. Meanwhile, we also noticed that our RFMask(4) even outperforms RFPose(12), RFPose(32), and RFPose(64) in all subsets. When we increase the sequence length of our model to 12, RFMask achieves mask IoU of 0.706 on single-person subset, 0.711 on multi-person subset, and 0.705 on action subset. Our RFMask(12) brings 0.031, 0.08, and 0.091 improvement than its RFPose(12) counterpart.  


\begin{table}[ht]
    \small
    \vspace{-4mm}
    \caption{Comparisons with RFPose}
    \label{comparisons}
    \centering
    \setlength{\tabcolsep}{3.6mm}{
        \begin{tabular}{cccc}
            \toprule
            Model & Single-Person & Multi-Person & Action \\
            \midrule
            RFPose(4) & 0.664 & 0.626 & 0.616 \\
            RFPose(12) & 0.675 & 0.631 & 0.614 \\
            RFPose(32) & 0.661 & 0.617 & 0.598 \\
            RFPose(64) & 0.641 & 0.589 & 0.604 \\
            RFMask(4) & 0.681 & 0.682 & 0.681 \\
            RFMask(12) & \textbf{0.706} & \textbf{0.711} & \textbf{0.705} \\
            \bottomrule
        \end{tabular}
    }
        
\end{table}

We noticed that RFPose has different performance on different subsets. 
For instance, the performance of RFPose on action subset is lower than that on multi-person subset, and the performance on multi-person subset is lower than that on single-person subset. The situation is different for our RFMask, whose performance is consistent and stable in all subsets. 
With the proposed two-stage framework, the noise and meaningless signals are filtered out in detection stage which greatly reduces the computational overhead. 
The proposed Multi-Head Fusion module further focuses on the spatial relationship of reflections from human body. 
In conclusion, the proposed framework could handle more challenges raised by complex scenarios with multiple persons and various actions. 

\subsubsection{Performance Under Occlusion or Low Illumination}
We also present experimental results under occlusion and low illumination scenarios to verify the feasibility of our proposed RFMask.
In the challenging scenarios, such as occlusion and low illumination, the ground-truth human silhouette cannot be obtained through the multi-camera system, and thus the quantitative evaluation cannot be conducted. Here, we only show the qualitative comparisons. 

Specifically, when encountering occlusion scenario, the camera along with the radars would fail to work, but the rest of the camera nodes can still be used to estimate the 3D keypoints of human. Once 3D keypoints is obtained, we can re-project the 3D keypoints back to the occluded view to obtain the 2D keypoints of human as the reference ground-truth to evaluate the accuracy of RFMask qualitatively. On the other hand, when encountering dim environment, all 13 cameras would not work which means the aforementioned re-projection process can not be conducted, hence, there is no ground-truth to be referenced. 

The corresponding results are shown in Figure \ref{occlution_res}. We can see that RFMask works well under the occlusion and low illumination conditions. 
Therefore, it can be used as a good supplementary for the vision-based system to improve the system performance. 

\begin{figure}[h]
    \centering
    \includegraphics[width=0.8\columnwidth]{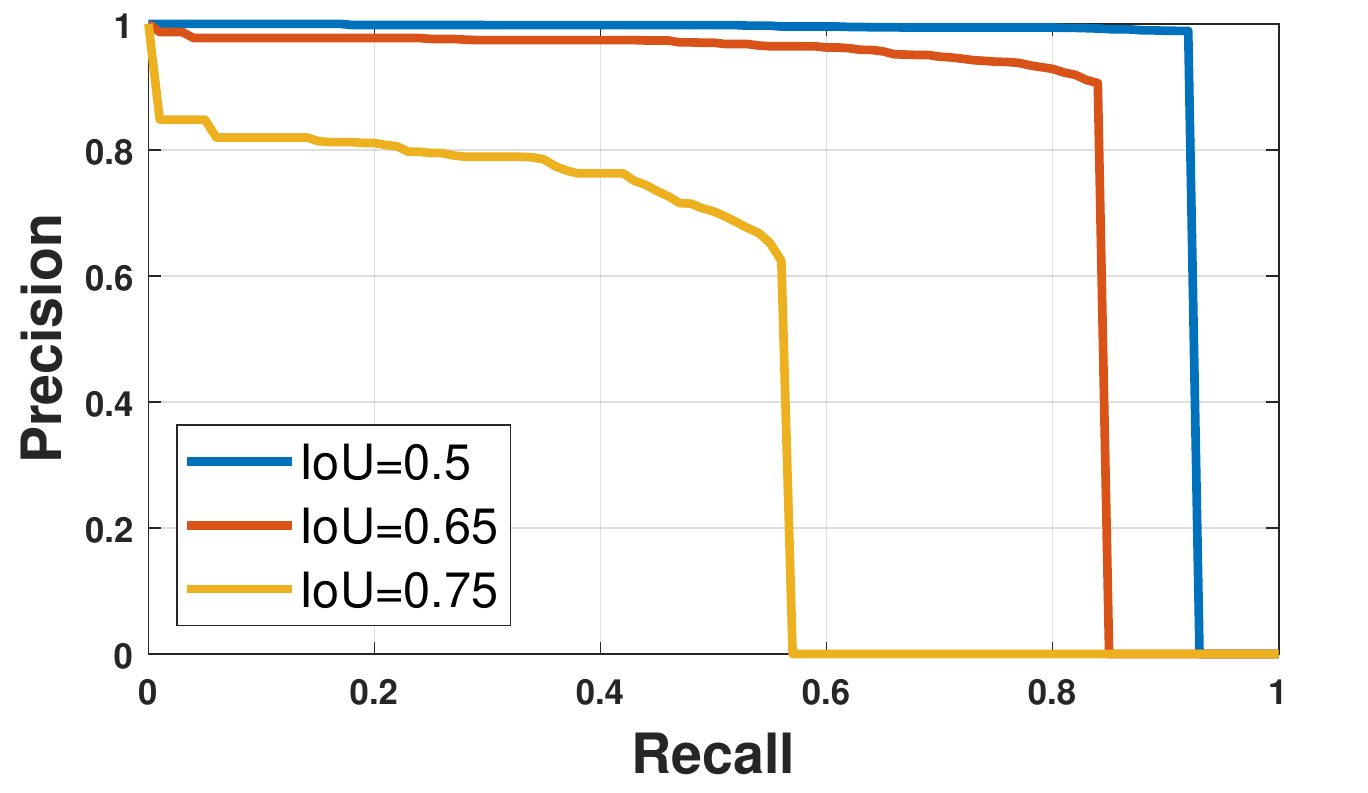}
    \caption{Precision-Recall curve of our location results. We present PR curve under three IoU thresholds: 0.5, 0.65, and 0.75.}
    \label{pr_curve}
\end{figure}

\subsubsection{Localization Performance of Detection Module} 
Our proposed RFMask follows the two-stage detection-segmentation principle. The performance of detection module plays an important role in helping decoder network to focus on the correct position and generate accurate results.
Therefore, in this subsection, we will present the localization performance of human detection module in RFMask. Similar to common object detection algorithms in computer vision, we adopt average precision (AP) under different confidence threshold as our evaluation metrics. The whole evaluation is conducted with input time sequence length of 4 and 12, and each with ResNet-18, ResNet-34, ResNet-50 as feature extraction backbone. 
Our RFMask with ResNet-50 backbone and sequence length of 12 achieves highest AP, which is 0.632 and highest recall which is 0.701. We also illustrate our Precision-Recall curve under the threshold of 0.5, 0.65, and 0.75 in Figure \ref{pr_curve}. 

The experimental results show that our proposed framework is robust enough to achieve human detection. The performance of RFMask(12) outperforms RFMask(4) counterparts, which means longer sequence helps to improve the location accuracy. Although deeper network brings better performance, the improvement is limited. With the input sequence length of 12, the RFMask with ResNet-18 backbone only 0.011 behind ResNet-50 backbone version. 
This is due to the fact that the representation of RF signals is much simpler than RGB images, which enables shallower network to achieve high accuracy.
Quantitative results can be found in Table \ref{location}.

\begin{table}[ht]
    \small
    \vspace{-4mm}
    \caption{Location Accuracy}
    \label{location}
    \centering
    \setlength{\tabcolsep}{2.0mm}{
        \begin{tabular}{cccccc}
            \toprule
            Model & Backbone & $\rm AP_{50:95}$ & $\rm AP_{50}$ & $\rm AP_{75}$ & $\rm Recall$\\
            \midrule
            RFMask(4) & ResNet-18 & 0.586 & 0.966 & 0.678 & 0.662  \\
            RFMask(4) & ResNet-34 & 0.590 & 0.966 & 0.689 & 0.665 \\
            RFMask(4) & ResNet-50 & 0.581 & 0.966 & 0.671 & 0.656 \\
            RFMask(12) & ResNet-18 & 0.621 & 0.967 & 0.783 & 0.691 \\
            RFMask(12) & ResNet-34 & 0.631 & 0.967 & 0.817 & 0.699 \\
            RFMask(12) & ResNet-50 & \textbf{0.632} & \textbf{0.967} & \textbf{0.824} & \textbf{0.701} \\
            \bottomrule
        \end{tabular}
    }
        
\end{table}

\subsubsection{Ablation study}
We conduct an ablation study to demonstrate the effectiveness of modules in our proposed RFMask. We implement a single-branch version of RFMask which only takes horizontal or vertical signal as input. We then verify the effectiveness of our proposed Multi-Head Fusion module by using simple feature concatenation as baseline. 

The quantitative results are presented in Table \ref{ablation}. Our single-branch version of RFMask is feed with horizontal signals, denoted as H. Our original dual-branch RFMask is denoted as H \& V. Here we only evaluate single-branch version of RFMask on horizontal signals because vertical signals only contain limited vertical movement information, where most of the information with respect to human actions is lost.

In Table \ref{ablation}, we can see dual-branch model outperforms the single-branch version. This indicates that although the signal in horizontal plane contains most of the information, the signal in vertical plane also plays an important role for segmenting human silhouette. 
Furthermore, the performance can be improved significantly with the proposed Multi-Head Fusion module as illustrated in Table \ref{ablation}. 
Compared with simple concatenation, our Multi-Head Fusion module could effectively focus on the spatial characteristics on two input sequences with self-attention mechanism . 

\begin{figure}[h]
    \centering
    \subfloat[]{
        \begin{minipage}[b]{0.22\linewidth}
            \includegraphics[width=1\textwidth]{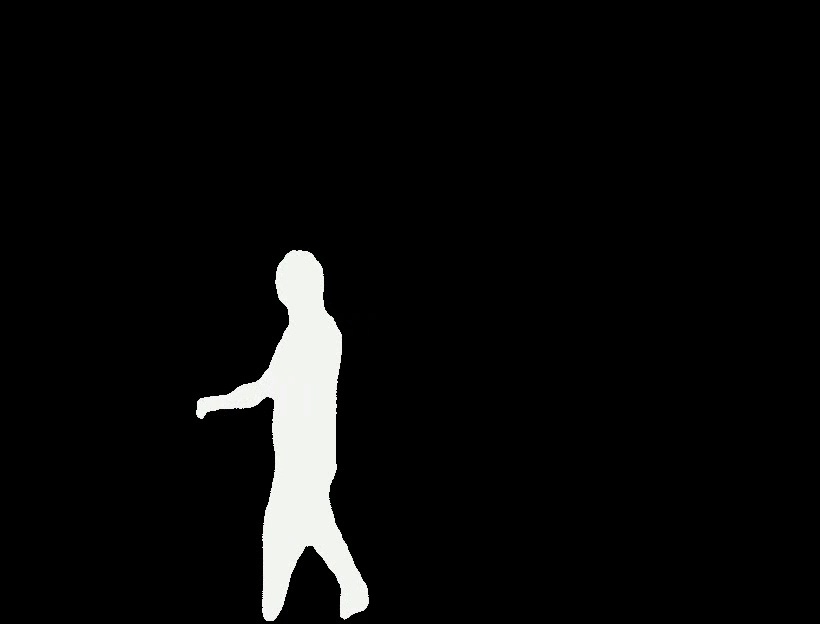} \\
            \vspace{-0.4cm}
            \includegraphics[width=1\textwidth]{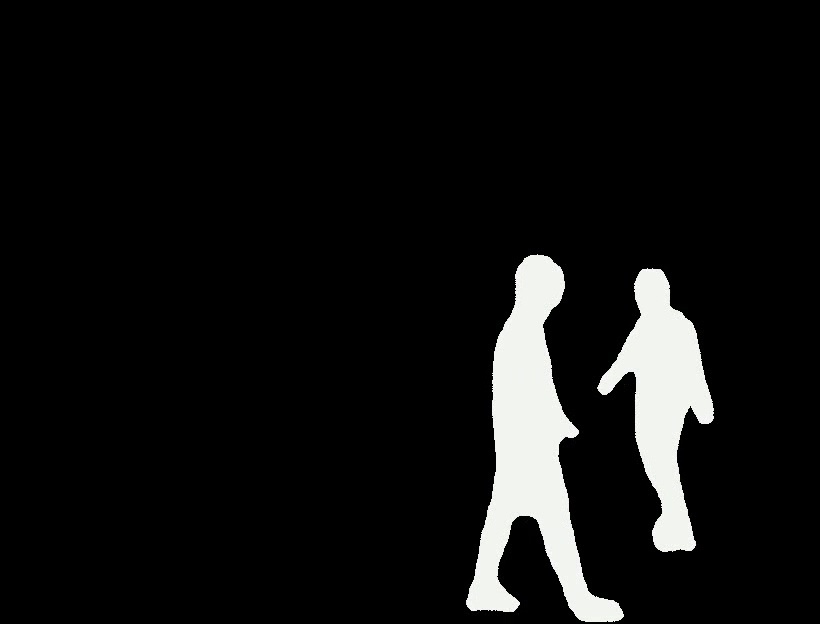} \\
            \vspace{-0.4cm}
            \includegraphics[width=1\textwidth]{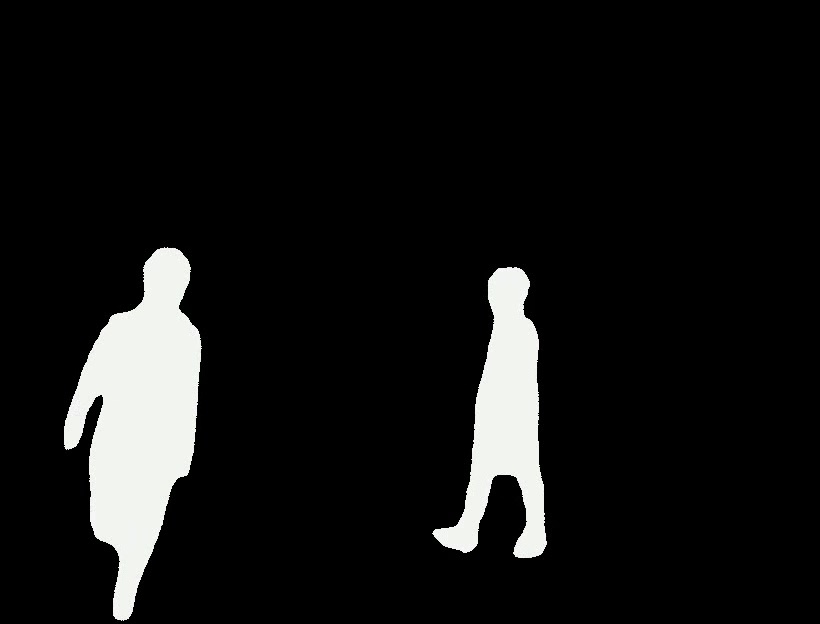} \\
            \vspace{-0.4cm}
            \includegraphics[width=1\textwidth]{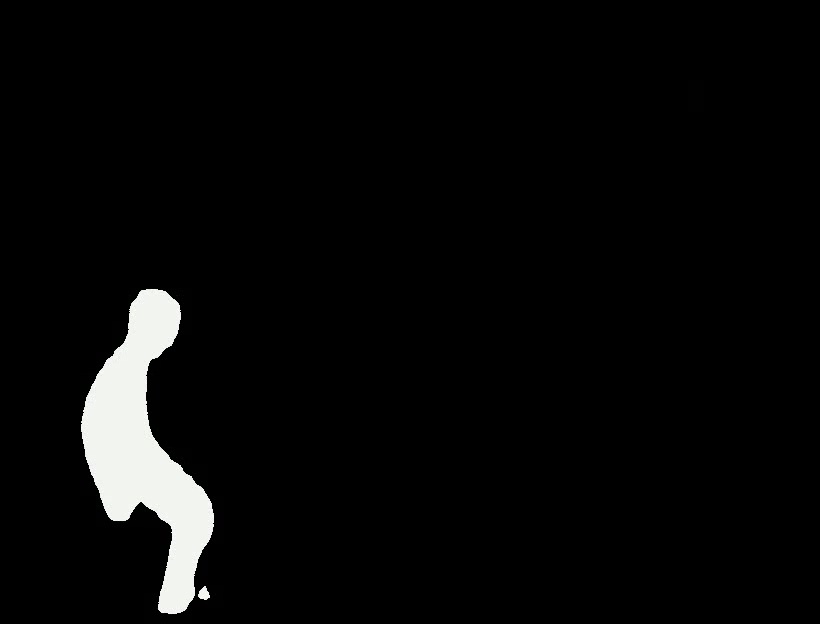} \\
            \vspace{-0.4cm}
            \includegraphics[width=1\textwidth]{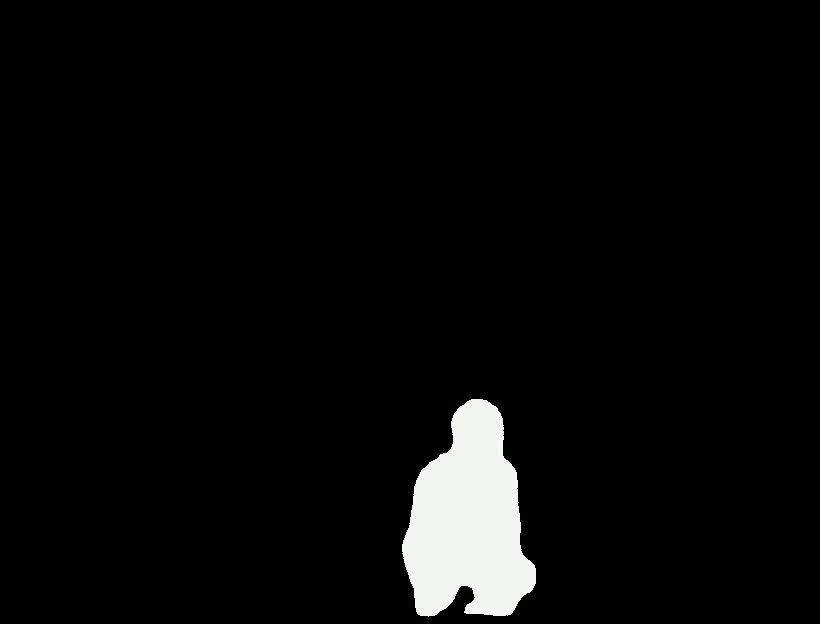} \\
            \vspace{-0.3cm}
        \end{minipage}
        \label{groundtruth}
    }
    \subfloat[]{
        \begin{minipage}[b]{0.22\linewidth}
            \includegraphics[width=1\textwidth]{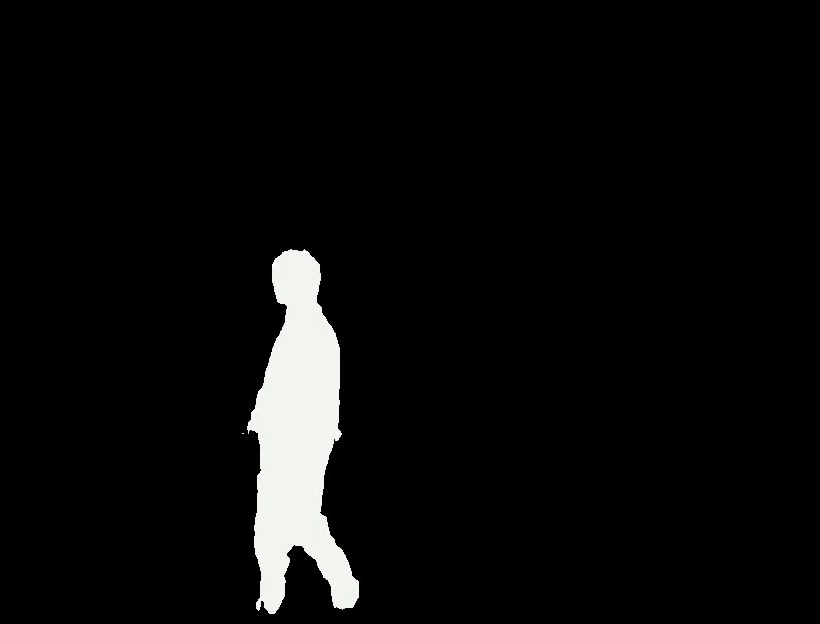} \\
            \vspace{-0.4cm}
            \includegraphics[width=1\textwidth]{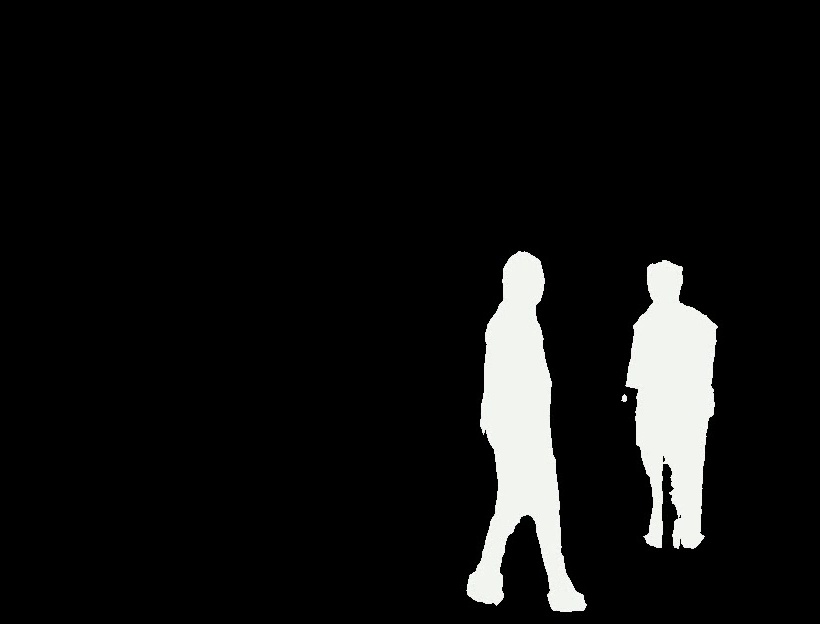} \\
            \vspace{-0.4cm}
            \includegraphics[width=1\textwidth]{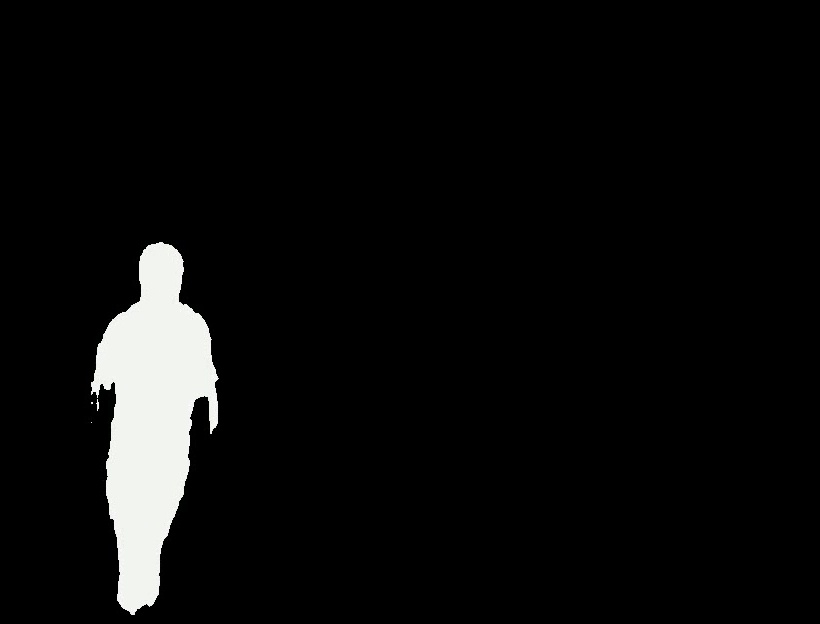} \\
            \vspace{-0.4cm}
            \includegraphics[width=1\textwidth]{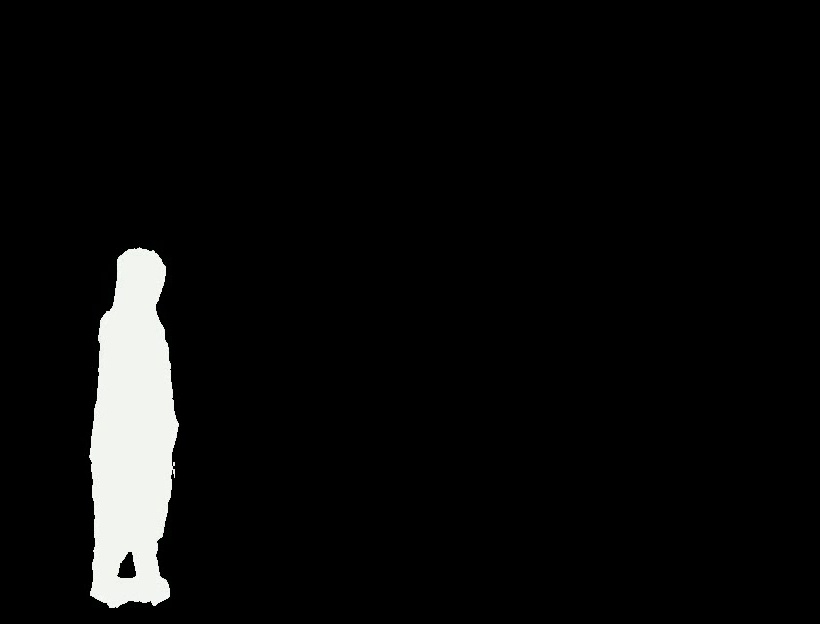} \\
            \vspace{-0.4cm}
            \includegraphics[width=1\textwidth]{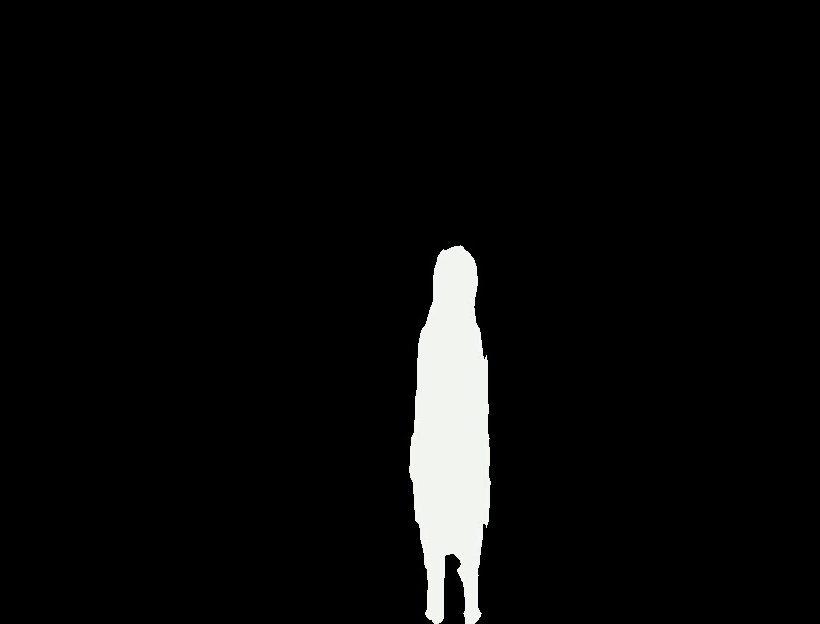} \\
            \vspace{-0.3cm}
        \end{minipage}
        \label{two_branch}
        \vspace{-0.3cm}    
    }
    \hspace{-0.25cm}
    \subfloat[]{
        \begin{minipage}[b]{0.22\linewidth}
            \includegraphics[width=1\textwidth]{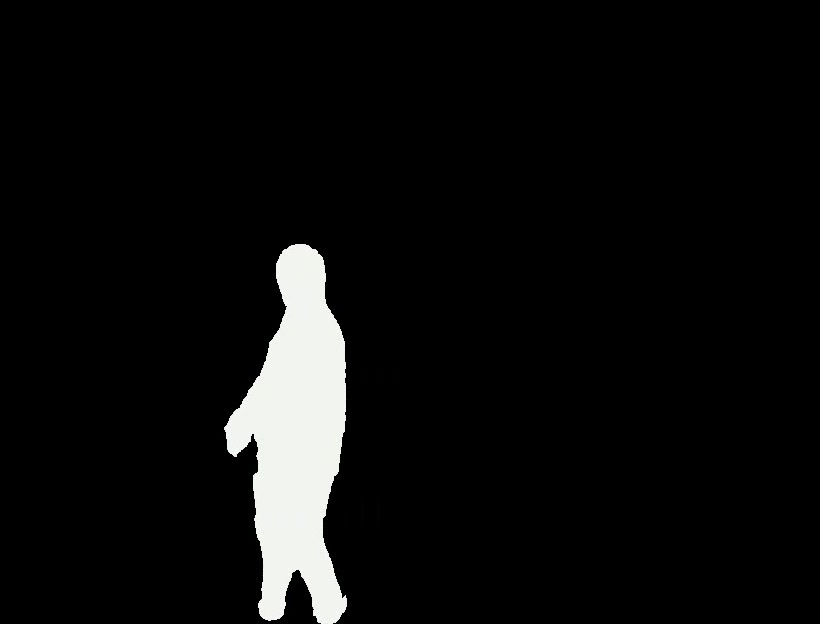} \\
            \vspace{-0.4cm}
            \includegraphics[width=1\textwidth]{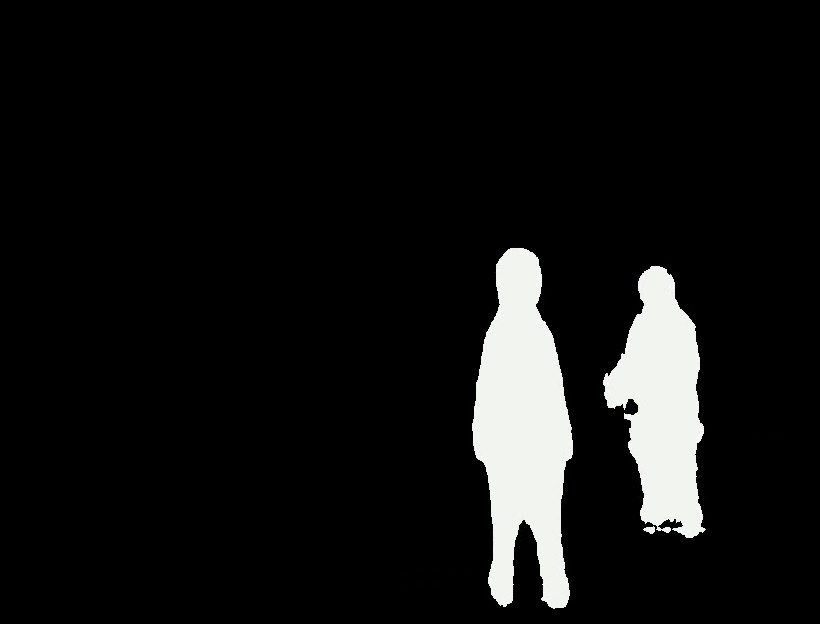} \\
            \vspace{-0.4cm}
            \includegraphics[width=1\textwidth]{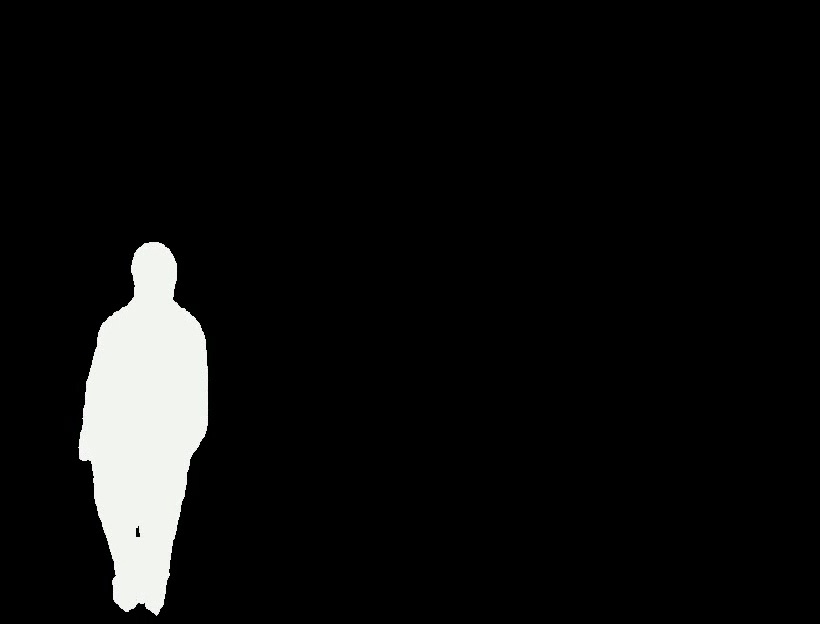} \\
            \vspace{-0.4cm}
            \includegraphics[width=1\textwidth]{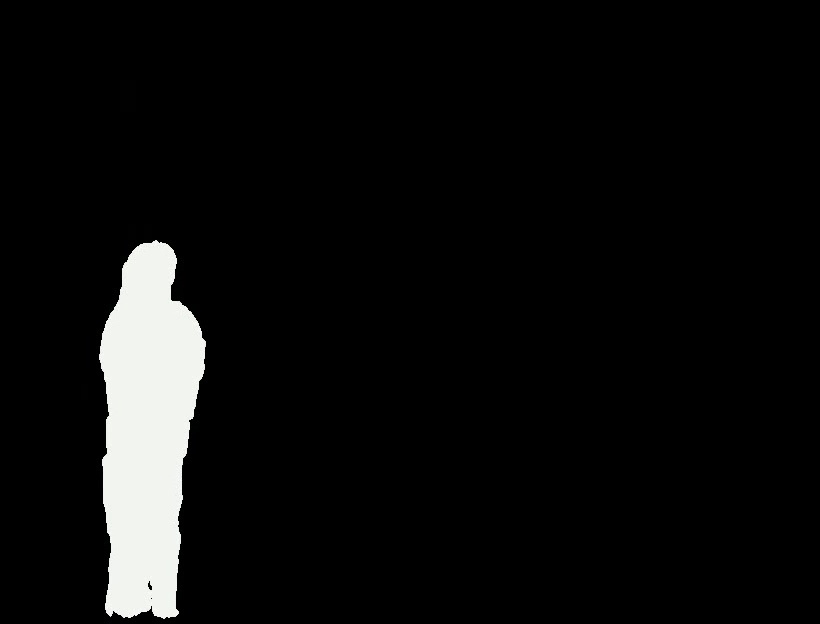} \\
            \vspace{-0.4cm}
            \includegraphics[width=1\textwidth]{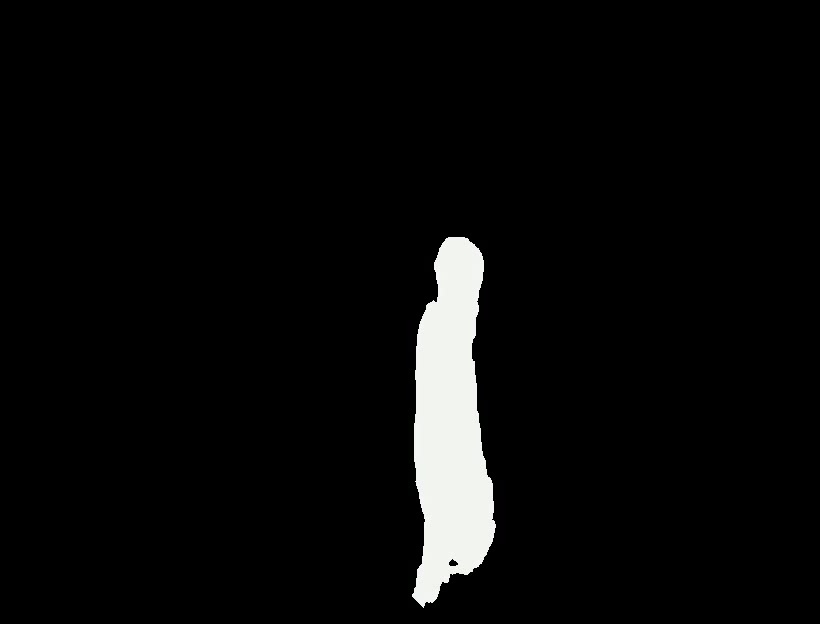} \\
            \vspace{-0.3cm}
        \end{minipage}
        \label{single_branch}
    }
    \subfloat[]{
        \begin{minipage}[b]{0.22\linewidth}
            \includegraphics[width=1\textwidth]{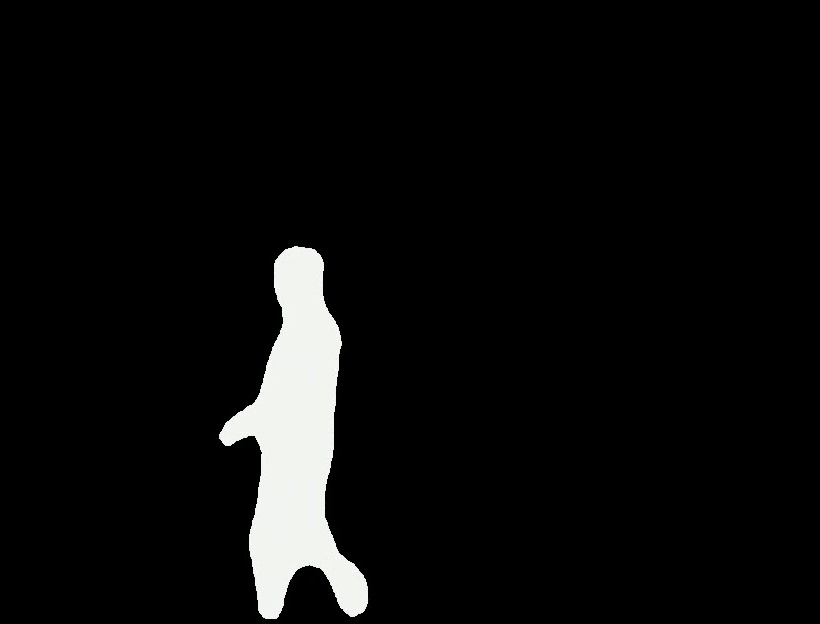} \\
            \vspace{-0.4cm}
            \includegraphics[width=1\textwidth]{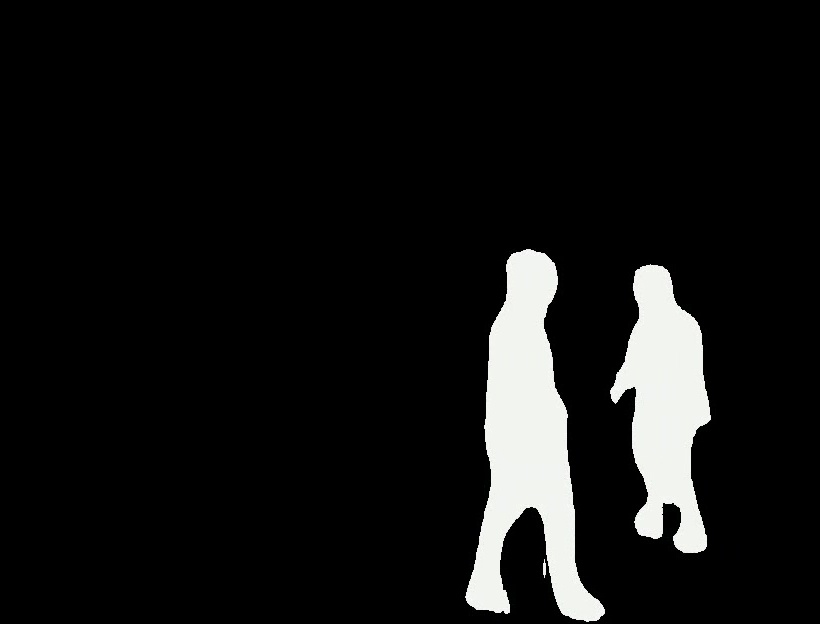} \\
            \vspace{-0.4cm}
            \includegraphics[width=1\textwidth]{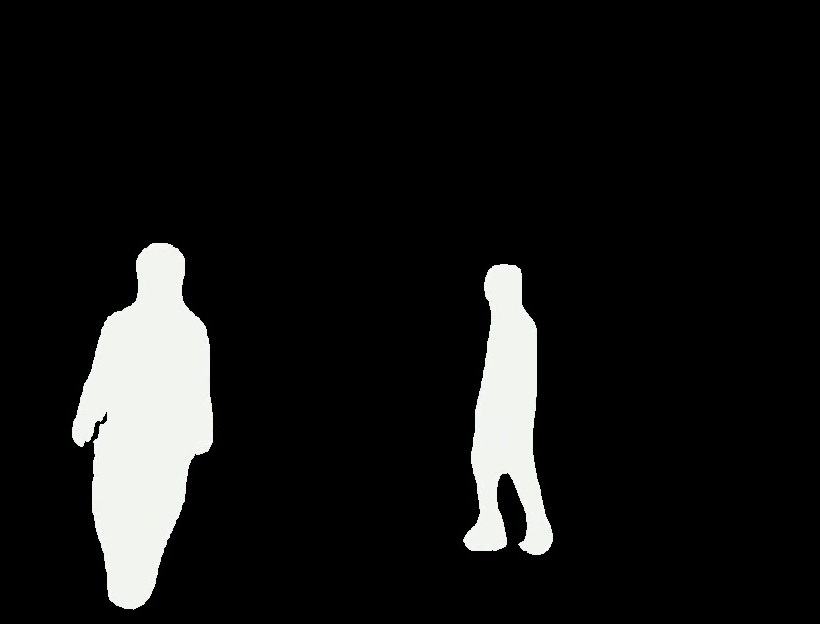} \\
            \vspace{-0.4cm}
            \includegraphics[width=1\textwidth]{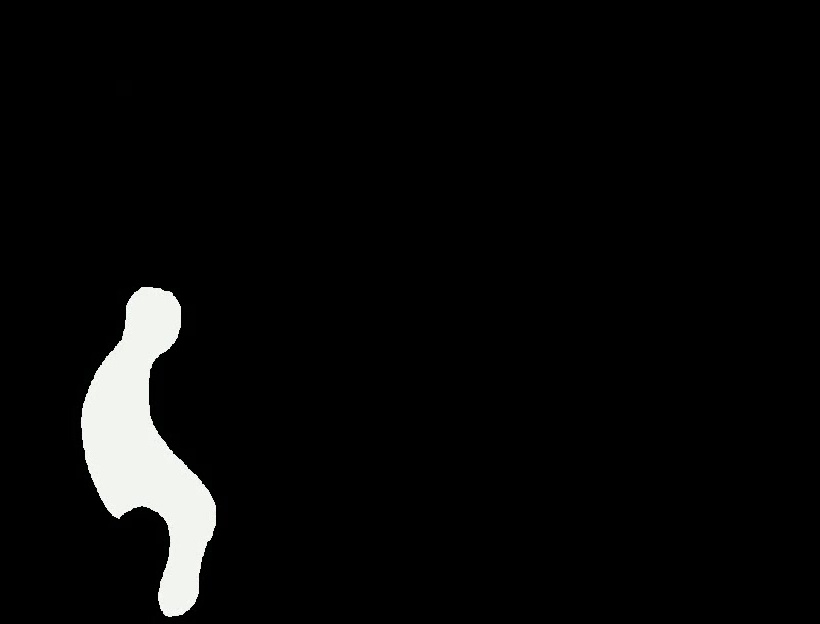} \\
            \vspace{-0.4cm}
            \includegraphics[width=1\textwidth]{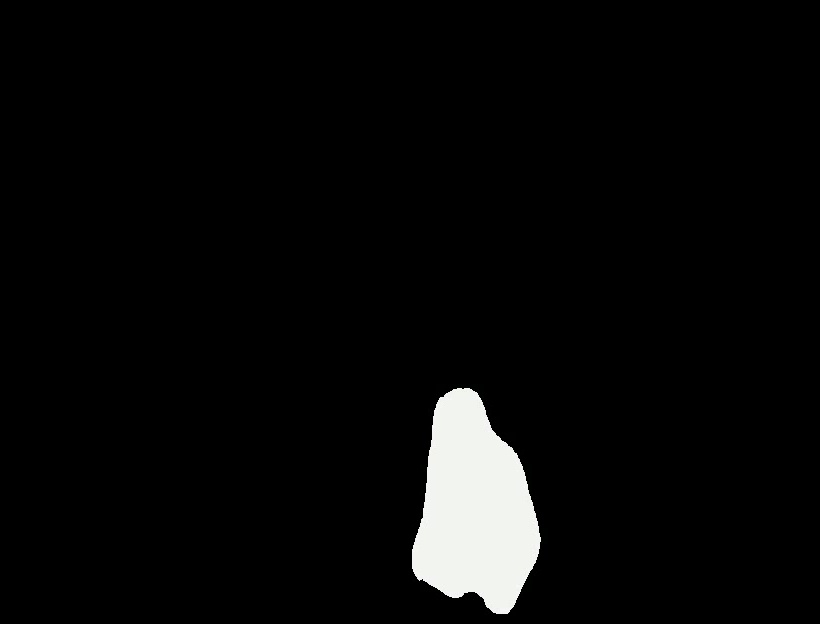} \\
            \vspace{-0.3cm}
        \end{minipage}
        \label{trans_fusion}
    }
	
    \caption{Qualitative comparison: (a) the ground-truth; (b) the results of RFMask with the horizontal information only; (c) the results of RFMask with both horizontal and vertical information, but using simple concatenation to fuse feature; (d) the results of RFMask with both horizontal and vertical information, and adopt our proposed Multi-Head Fusion module.}
    \label{ablation_results}
    \vspace{-3mm}
\end{figure}

We also illustrate some qualitative comparisons of our single-branch and dual-branch version of RFMask in Figure \ref{ablation_results}. 

\begin{table}[ht]
    \vspace{-4mm}
    \small
    \caption{Ablation Study}
    \label{ablation}
    \centering
    \setlength{\tabcolsep}{4.7mm}{
        \begin{tabular}{cccc}
            \toprule
            & H & H \& V & H \& V \\
            \midrule
            Dual-Branch & & \checkmark & \checkmark \\
            Multi-Head Fusion & & & \checkmark \\
            \midrule
            Single-Person(4) & 0.634 & 0.644 & \textbf{0.681} \\
            Multi-Person(4) & 0.587 & 0.604 & \textbf{0.682} \\
            Action(4) & 0.582 & 0.585 & \textbf{0.681} \\
            Single-Person(12) & 0.655 & 0.670 & \textbf{0.706} \\
            Multi-Person(12) & 0.638 & 0.642 & \textbf{0.711} \\
            Action(12) & 0.603 & 0.603 & \textbf{0.705} \\
            \bottomrule
        \end{tabular}
    }
    \vspace{-5mm}
\end{table}

\section{Conclusion}
\label{conclusion}
In this paper, we propose a human silhouette segmentation framework, RFMask, to segment human silhouette from the millimeter wave RF signals. To the best of our knowledge, this is the first work to segment human silhouette from the millimeter wave RF signals. We also create a multi-modal dataset that contains tens of thousands of RF frames and the corresponding optical camera images of human activity. We hope that our RFMask can serve as a baseline and together with the dataset can inspire more research that perform vision tasks with RF signals. Our dataset and codes will be released in public.
\bibliographystyle{IEEEtran}
\bibliography{aaai22}

\vfill

\end{document}